%% file: main.tex
\documentclass[10pt,journal,compsoc]{IEEEtran}

\usepackage{graphicx}
\usepackage{amsmath}
\usepackage{amssymb}
\usepackage{booktabs}

\usepackage[breaklinks,colorlinks]{hyperref}
\usepackage{cite}
\usepackage{subcaption}

\usepackage{xcolor}
\usepackage{siunitx}
\usepackage{adjustbox} %
\usepackage{array}
\usepackage{multirow}
\usepackage{rotating}
\usepackage{etoolbox}
\robustify\bfseries

\usepackage{makecell}
\usepackage[capitalize]{cleveref}
\crefname{section}{Sec.}{Secs.}
\Crefname{section}{Section}{Sections}
\Crefname{table}{Table}{Tables}
\crefname{table}{Tab.}{Tabs.}

\usepackage{etoolbox} %

\newif\ifarxiv
\arxivtrue %

\input{macros}
\def\MYTITLE{Event-based Background-Oriented Schlieren}

\ifarxiv
\usepackage{orcidlink}
\usepackage{balance}
\usepackage[absolute]{textpos}
\hypersetup{
  pdftitle={\MYTITLE},
  pdfsubject={Computer Vision, Robotics, Deep Learning},
  pdfauthor={Shintaro Shiba, Friedhelm Hamann, Yoshimitsu Aoki, Guillermo Gallego},
  pdfkeywords={Event camera, Schlieren imaging, Background-oriented schlieren, Optical flow}
}
\makeatletter
\long\def\@IEEEtitleabstractindextextbox#1{\parbox{0.922\textwidth}{#1}}
\makeatother
\fi

\begin{document}

\title{\MYTITLE}

\ifarxiv
\definecolor{somegray}{gray}{0.6}
\newcommand{\darkgrayed}[1]{\textcolor{somegray}{#1}}
\begin{textblock}{11}(2.5, 0.4)
\begin{center}
\darkgrayed{This paper has been accepted for publication at the\\
IEEE Transactions on Pattern Analysis and Machine Intelligence, 2023.
\copyright IEEE}
\end{center}
\end{textblock}
\fi 

\ifarxiv
\author{Shintaro~Shiba$^{1,2}$\orcidlink{0000-0001-6053-2285},
Friedhelm~Hamann$^{1,3}$\orcidlink{0009-0004-8828-6919},
Yoshimitsu~Aoki$^2$\orcidlink{0000-0001-7361-0027},
and~Guillermo~Gallego$^{1,3,4}$\orcidlink{0000-0002-2672-9241}%
\IEEEcompsocitemizethanks{
\IEEEcompsocthanksitem $^1$ Technische Universit\"at Berlin, Berlin, Germany.
\IEEEcompsocthanksitem $^2$ Department of Electronics and Electrical Engineering, Faculty of Science and Technology, Keio University, Kanagawa, Japan.
\protect\\
E-mail: sshiba@keio.jp
\IEEEcompsocthanksitem $^3$ Science of Intelligence Excellence Cluster, Berlin, Germany.
\IEEEcompsocthanksitem $^4$ Einstein Center Digital Future, Berlin, Germany.
\IEEEcompsocthanksitem Preprint of IEEE T-PAMI paper. DOI: 10.1109/TPAMI.2023.3328188}%
}
\fi

\IEEEtitleabstractindextext{%
\begin{abstract}
Schlieren imaging is an optical technique to observe the flow of transparent media, such as air or water, without any particle seeding.
However, conventional frame-based techniques require both high spatial and temporal resolution cameras, which impose bright illumination and expensive computation limitations.
Event cameras offer potential advantages (high dynamic range, high temporal resolution, and data efficiency) to overcome such limitations due to their bio-inspired sensing principle.
This paper presents a novel technique for perceiving air convection using events and frames by providing the first theoretical analysis that connects event data and schlieren.
We formulate the problem as a variational optimization one combining the linearized event generation model with a physically-motivated parameterization that estimates the temporal derivative of the air density.
The experiments with accurately aligned frame- and event camera data reveal that the proposed method enables event cameras to obtain on par results with existing frame-based optical flow techniques.
Moreover, the proposed method works under dark conditions where frame-based schlieren fails, and also enables slow-motion analysis by leveraging the event camera's advantages.
Our work pioneers and opens a new stack of event camera applications, as we publish the source code as well as the first schlieren dataset with high-quality frame and event data. \url{https://github.com/tub-rip/event_based_bos}.
\end{abstract}

}

\maketitle
\IEEEdisplaynontitleabstractindextext

\section{Introduction}
\label{sec:intro}

\IEEEPARstart{S}{ensing} the flow of transparent media, such as air or water, is important for various applications from aerodynamics to gas leakage detection.
Optical imaging is a useful tool to examine such transparent media
because it can capture the media with high detail in space-time remotely.
Among existing methods, schlieren imaging is a simple but efficient optical tool for seeing the ``invisible'' \cite{settles2001schlieren,settles2017review}: 
inhomogeneities in transparent media that are not necessarily perceived by the naked eyes.
It requires simple recording settings: lenses, cameras, and mirrors or background patterns 
to image how light rays deviate due to refractive index variations in the media.
While it was initially conceived as a visualization technique, recent developments in schlieren and shadowgraphy have extended the usage to velocimetry \cite{settles2017review,settles2022schlieren}.
However, it requires a high-speed camera with a large spatial resolution to analyze the velocity of the flow, such as convection.
This is not only a constraint for real-world applications but also a limitation of the methodology 
because: 
($i$) achieving high shutter speeds requires unnaturally bright illumination, which is not always practical,
($ii$) transmitting and processing the large amount of redundant data acquired involves high bandwidth, storage, and power-hungry components, %
and ($iii$) regardless of the large power consumption, the trade-off between speed and spatial resolution limits accuracy in estimating the flow velocity.
\input{floats/fig_eyecatch}

Event cameras \cite{Lichtsteiner08ssc,Finateu20isscc} are novel bio-inspired sensors that respond to pixel-wise intensity changes, which are not always visible to conventional frame-based cameras.
They offer advantages such as high speed, high dynamic range (HDR), low power consumption, and data efficiency (temporal redundancy suppression) \cite{Gallego20pami},
which makes them potential candidates to overcome the limitations of traditional (i.e., frame-based) schlieren techniques.
However, despite these potential capabilities, the application of event cameras to imaging applications is yet to be explored and developed. 

This paper presents a novel technique, event-based background-oriented schlieren (BOS), for sensing air convection with event cameras and proposes a novel method to estimate the temporal derivative of air density from events and frames (\cref{fig:eyecatch}).
Throughout the paper, we tackle the following challenges of event-based BOS:
($i$) Theory. There is no established mathematical theory for event-based schlieren techniques.
($ii$) Data. Event cameras sense only increments of schlieren as opposed to the larger differences with respect to a reference in frame-based BOS.
($iii$) Methodology. 
The origin of events in BOS (flickering because they happen only at the edges of the background pattern) 
and large amounts of noise are novel and difficult for previous work in event-based vision.
($iv$) Evaluation. The real-world ground truth of the air density is not easy to obtain, hence we need some proxy ground truth and baselines.

First, we develop a theoretical connection between the schlieren and events, showing that event cameras can sense the inhomogeneities of transparent media in a more direct way (as flickering events) compared to frame-based cameras.
Such direct sensing of schlieren through event data enables us to observe air convection at high speed more precisely and under challenging lighting conditions.
Second, we propose a novel method that extends the linearized event generation model with physically-inspired parameterization to estimate the temporal density fluctuation
due to the schlieren.
Third, in order to evaluate the estimated density change, whose real-world ground truth is not easy to obtain,
we establish the evaluation method using optical flow, by revealing the theoretical connection between the temporal density change and optical flow (i.e., pixel displacement).
Using a co-located frame-based camera enables us to benchmark different methods of estimating temporal density change as a computer vision problem.
The experimental results show that:
($i$) our proposed method recovers the flow that corresponds to the temporal change of density gradient by comparing with the standard frame-based methods and other baseline methods,
($ii$) flickering-like events are a more direct measurement of such schlieren,
($iii$) event cameras record the density inhomogeneities even in poor lighting conditions, which state-of-the-art frame-based algorithms cannot provide,
and ($iv$) the high temporal resolution of event cameras enables slow-motion schlieren analysis.

The main technical contributions of this work are:
\begin{itemize}
    \item A novel method for computation of schlieren combining events and frames (\cref{sec:method,sec:proposedMethod}). 
    The proposed method is rigorously obtained and well connected with the physical model of the sensors involved via the linearized event generation model.
    \item The first schlieren event-frames dataset (\cref{sec:implementation}). 
    We publicly provide recordings of several schlieren scenes by means of events and frames, at high resolution (1~Mpixel), accurately synchronized and calibrated using an in-house acquisition system.
    \item A thorough comparison with baseline methods despite the lack of truly ground truth data in this type of turbulent fluid dynamics phenomena (\cref{sec:experim}). 
    \item \sblue{A simulation experiment based on the high-speed frame data from \cite{settles2022schlieren} 
    to confirm the results of event-based BOS using those from frame-based BOS
    (\cref{sec:simulation}).} 
\end{itemize}

To the best of our knowledge, this is the first work showing the potential advantages of event cameras for schlieren imaging applications. %

\section{Related Work}
\label{sec:related}

\subsection{Background-Oriented Schlieren}
Schlieren photography was invented in 1864 to study the flow of air around objects moving at supersonic speed \cite{settles2001schlieren}.
In contrast to other imaging and velocimetry techniques such as particle image velocimetry \cite{Raffel1998particle,Ding2023pami}, it does not require any particle seeding in the media of interest.
Among different schlieren-imaging techniques (see \cref{tab:schlierenComparison}), BOS is a relatively recent technique since it utilizes digital image processing \cite{richard2001principle}.
In BOS (\cref{fig:method:bos}), an object of interest with density variations (e.g., the hot air stream from a burning candle) is placed between the camera and a constant (non-moving) background pattern.
The schlieren generates complex deformation to the background pattern,
which is observed by cameras as the apparent motion of the background pattern with respect to a reference image (without density variations) \cite{richard2001principle}.
Different methods have been proposed to compute the displacement vector field of the apparent motion, such as using 
cross-correlation \cite{goldhahn2007background},
optical flow \cite{atcheson2009evaluation},
or wavelet-based analysis \cite{schmidt2021wavelet}.
As equally important as the data processing method is the data acquisition setup.
Best practices for parameter settings, such as the distance from the camera to the background and the media, are provided in \cite{settles2017review}.

\input{floats/tab_schlieren_comparison.tex}

\input{floats/fig_bos_setting}

BOS has been used to image various transparent media, such as
shock waves from explosions \cite{sommersel2008application},
turbulent flows \cite{settles2022schlieren},
and shock waves underwater \cite{hayasaka2016optical}.
Also, the background pattern of BOS can be extended to natural images \cite{hargather2010natural},
which allows us to image the flow with a large field-of-view (FOV).
In \cite{Heineck21aiaa}, BOS is utilized to visualize supersonic jets in flight,
by leveraging the natural vegetation of the terrain seen from above as the constant background pattern.
The large FOV is one of the unique characteristics of BOS, unlike other schlieren techniques, which enables measuring natural outdoor scenes \cite{settles2017review}.
Notwithstanding, BOS can be used as input to other analysis tools, such as Dynamic Mode Decomposition (DMD) to reveal the main frequency modes of variation of the signal in space and time \cite{Tu14jcd}, which ultimately inform about the physical parameters of the turbulent flow.
Recently, some works have extended BOS from an imaging technique to a quantitative method, e.g., \cite{venkatakrishnan2004density} measures density of axisymmetric supersonic flow.
In \cite{settles2022schlieren}, a method is proposed to extract velocity data from flows.
For this application, Kymography works better than classical image correlation, and the self-similarity of round turbulent jet velocity appears in the schlieren results.

\subsection{Event Cameras}

Event cameras are a relatively new technology compared to BOS imaging with standard frame-based cameras.
Since the 2008 seminal work \cite{Lichtsteiner08ssc}, they have been slowly commercialized and explored in computer vision and robotics for various applications. 
Event cameras naturally respond to motion in the scene at high speed and HDR in a data-efficient manner, 
hence large progress has been made in motion-related tasks, such as optical flow estimation \cite{Benosman14tnnls,Zhu18rss,Paredes21neurips,Shiba22eccv,Shiba23spl},
ego-motion estimation \cite{Gallego18cvpr,Nunes21pami,Peng21pami,Shiba22aisy},
SLAM \cite{Kim16eccv,Rebecq17ral,Zhu17cvpr,Hidalgo22cvpr}, 
or video deblurring and frame interpolation \cite{ZhangX22cvpr,Tulyakov22cvpr,Gao22pami}.

Only recently, the larger spatial resolution of event cameras and higher fill factor of their pixels \cite{Finateu20isscc,Suh20iscas} has enabled fine-detail applications that were not possible with older models.
Some works have explored event cameras for detecting small changes in the scene. 
These include vibration monitoring \cite{Pfrommer22arxiv},
particle-image velocimetry \cite{Willert22expfl},
and time-resolved 3D fluid flow reconstruction via collimated illumination \cite{Wang20eccv}.
These works open another stack of event camera applications in the field of fluid dynamics. 
Event-based BOS aims at pushing the limits, by imaging and quantifying flow fields without any particle seeding.
\sblue{Event cameras are available from several manufacturers, costing in the range of 2000 to 6000 USD (as of 2023).
We refer to \cite{Gallego20pami,Gallego17resources} for more details on the different camera types and manufacturers.}

\sblue{\textbf{Physics-based Methods}. 
The method developed in this work is related to a body of literature that leverages the physics of the camera (event generation model) in an optimization framework 
to either estimate some motion parameters (e.g., feature tracks \cite{Gehrig19ijcv}, camera ego-motion \cite{Gallego15arxiv,Bryner19icra,Hidalgo22cvpr}, optical flow with additionally-provided frame information \cite{Pan20cvpr}), and/or a grayscale image given the motion \cite{Paredes21cvpr,Zhang22pami,Pan20cvpr}.}
\sblue{Our work builds on top of the event generation model of the camera \cite{Gallego20pami}, extending it to the considered BOS problem.
Specifically, we extend \cite{Gehrig19ijcv}, which was designed for sparse patches around keypoints undergoing Euclidean in-plane motion, to the case of dense flow fields caused by complex (i.e., non-rigid) schlieren.}

\section{Event-based Schlieren}
\label{sec:event}
\label{sec:method}

\subsection{Principles of Frame-based BOS}
\label{sec:frame:principle}

In frame-based BOS the schlieren object $S$ (e.g., a gas with varying density) produces an apparent displacement of the background pattern, 
which is measured with respect to the initial state (i.e., image acquired in the absence of density gradient).
The displacement $\Delta \bx \doteq (\Delta x, \Delta y)^{\top}$ is directly related to the small deflection angle $\beps \doteq (\epsilon_x, \epsilon_y)^{\top}$ (\cref{fig:method:bos}) via the distance from lens to $S$ ($\dobj$), the distance from $S$ to the background ($Z_D$), 
and the focal length of the lens $f$ \cite{raffel2015background}:
\begin{equation}
  \label{eq:triangulation}
  \Delta \bx 
  \approx f\left( \frac{Z_D}{Z_D + \dobj -f} \right) \beps.
\end{equation}

On the other hand, for the refractive index  $n$, the angle $\beps$ is the result of aggregating the spatial gradient $\partial{n}/\partial{\bx}$ along the length $Z$ of the schlieren object $S$ on the optical axis:
\begin{equation}
\label{eq:refractiveAngle}
  \beps = \frac{1}{n} \int \prtl{n}{\bx} dz = \frac{Z}{n_{\infty}} \prtl{n}{\bx},
\end{equation}
where the ambient-air refractive index is given as $n_{\infty}$.
Finally, $n$ is related to the density $\rho$ of the gas (schlieren object) via the Gladstone-Dale relation,
$n = G \rho + 1$, 
with constant $G = 2.23 \times 10^{-4} \text{m}^3/\text{kg}$ \cite{raffel2015background}.

In short, the spatial gradient of the density $\partial{\rho} / \partial{\bx}$ within a gas causing schlieren can be directly quantified by measuring the pixel displacement $\Delta \bx$:
\begin{equation}
\label{eq:DisplacementDensityGrad}
    \Delta \bx \propto \prtl{\rho}{\bx},
\end{equation}
as summarized in \cref{tab:schlierenComparison}.
Here, the displacement is measured against the initial state (the background pattern), hence the corresponding density-gradient field is the change with respect to the initial (also called ``reference'') state.

\subsection{Principles of Event-based BOS}
\label{sec:event:principle}

\input{floats/fig_bos_principle_comparison.tex}
One of the main differences between frame-based BOS and event-based BOS is that
event cameras only sense temporal changes of the scene,
while the former measures the displacement between a reference frame and the current frame (\cref{fig:method:bosPrincipleComparison}).
Hence, the key challenge is how we can relate events (the asynchronous intensity changes between two timestamps $t_1$ and $t_2$) to the density $\rho$.
Since events are very noisy \cite{Graca21iisw,Gallego20pami},
accumulating the differences between far away timestamps to estimate the same displacement as frame-based BOS \eqref{eq:DisplacementDensityGrad} leads to high noise levels \cite{Scheerlinck18accv,Brandli14iscas}, which makes it difficult to estimate this displacement with events.

In order to establish the theoretical connection between schlieren and events, let us first extend the previous frame-based BOS theory to compute the displacement between two \emph{nearby} timestamps.
Given frames at timestamps $t_1, t_2$, their displacements from a reference frame at $\tref$ \eqref{eq:triangulation} are $\Delta \bx (\tref, t_1)$ and $\Delta \bx (\tref, t_2)$. 
The optical flow $\velflow(\bx) = \partial{\bx} / \partial{t}$ between consecutive frames for small $\Delta t = t_2 - t_1$ is
\begin{equation}
\label{eq:flow}
    \velflow(\bx) %
    = \frac{\Delta \bx (\tref, t_2) - \Delta \bx (\tref, t_1)}{\Delta t}.
\end{equation}

From the frame-based BOS theory, the displacement at each timestamp can be related to the density gradient as follows \eqref{eq:DisplacementDensityGrad}:
\begin{equation}
\label{eq:spatialTwoTimestamps}
\begin{split}
    \Delta \bx (\tref, t_1) \propto \prtl{\rho_{t1}}{\bx}, \\
    \Delta \bx (\tref, t_2) \propto \prtl{\rho_{t2}}{\bx}.
\end{split}
\end{equation}

Plugging \eqref{eq:spatialTwoTimestamps} into \eqref{eq:flow}, using finite-difference approximations and Schwarz's theorem, gives:
\begin{equation}
\label{eq:flowToTemporalDeliv}
\begin{split}
    \velflow(\bx) & \propto \frac{1}{\Delta t} \bigl( \prtl{\rho_{t2}}{\bx} -  \prtl{\rho_{t1}}{\bx} \bigr) \\
    & = \frac{1}{\Delta t} \prtl{}{\bx} \bigl( \rho_{t2} -  \rho_{t1} \bigr) \\
    & \approx \prtl{}{\bx} \prtl{}{t} \rho, \\
    & = \prtl{}{t} \prtl{}{\bx} \rho. \quad \text{(Schwarz's thm)}
\end{split}
\end{equation}

That is, the optical flow between two nearby timestamps is related to the \emph{temporal derivative of the density gradient} (see the last row of \cref{tab:schlierenComparison}).
Since events are the measurements between such nearby timestamps, the key question is how can optical flow (i.e., spatio-temporal derivative of the density) be estimated from event data.

\input{floats/fig_method_cocapture}

\section{Estimation Method}
\label{sec:proposedMethod}

One of the main challenges of event-based BOS is its data modality: 
events generated by schlieren objects are sparse, happening only at the edges of the background pattern (\cref{fig:method:overlay}) and in a flickering form.
Previous event-based optical flow estimation methods \cite{Shiba22eccv,Zhu18rss,Paredes21neurips} often assume a continuous, non-flickering apparent motion of the visual patterns on the image plane.
Also, events triggered during the short time interval needed to capture fine details of the complex %
motion patterns 
are few compared to those in scenes from typical optical flow benchmarks \cite{Zhu18ral,Gehrig21ral}. 
Consequently prior methods fail to produce accurate flow since they are not tailored to the schlieren scenario, as we show in \cref{sec:experim:flowAccuracy}.
Due to these challenges, we propose a method that combines events and knowledge of the background pattern (e.g., frames) to estimate the flow. 
The proposed method extends the linearized event generation model (LEGM) \cite{Gehrig19ijcv,Bryner19icra,Pan20cvpr,Hidalgo22cvpr} to the characteristics of our problem.
The overall pipeline is described in \cref{fig:blockdiagram}.

\input{floats/fig_block_diagram}
\subsection{Event Generation Model} 
An event $e_k\doteq (\bx_k,t_k,\pol_k)$ conveys that the logarithmic brightness $\Lum$ at pixel 
$\bx_k$ changes by a specified contrast sensitivity $C$~\cite{Lichtsteiner08ssc,Gallego20pami}:
\begin{equation}
\label{eq:generativeEventCondition}
\Delta \Lum(\bx_k,t_k) \doteq \Lum(\bx_k,t_k) - \Lum(\bx_k, t_k-\Delta t_k) = \pol_k \, C,
\end{equation} 
where polarity $\pol_k \in \{+1,-1\}$ is the sign of the brightness change,
and $\Delta t_k$ is the time since the last event at pixel~$\bx_k$.
Given a set of events $\cE \doteq \{e_k\}_{k=1}^{\numEvents}$,
summing their polarities pixelwise produces a brightness increment image:
\begin{equation}
\label{eq:brightnessIncrementEvents}
\Delta \Lum(\bx) = \sum_{k} \pol_k C\, \delta(\bx-\bx_k),
\end{equation}
where the Kronecker $\delta$ selects the pixel $\bx_k$. 
The LEGM states that, assuming brightness constancy during a small $\Delta t = t_{\numEvents} - t_1$, 
the increment~\eqref{eq:generativeEventCondition} is caused by brightness gradients $\nabla\Lum$
moving with image velocity $\velflow$ \cite{Gallego15arxiv}:
\begin{equation}
\label{eq:brightnessIncrementGrad}
\Delta \Lum(\bx) \approx - \nabla \Lum(\bx) \cdot \Delta \bx = - \nabla \Lum(\bx) \cdot \velflow(\bx) \Delta t.
\end{equation}

\subsection{Optimization Objective}
We cast the problem of estimating the displacement \eqref{eq:flowToTemporalDeliv} as an optimization one,
where we minimize the mismatch between the event data (in the form of \eqref{eq:brightnessIncrementEvents}) 
and its prediction $\Delta \hat{\Lum}$ via \eqref{eq:brightnessIncrementGrad} exploiting the knowledge of the background pattern from a frame $\hat{\Lum}$.
This idea is summarized in \cref{fig:blockdiagram}.

To allow for the fact that $\hat{\Lum}$ may not be perfectly aligned with the corresponding events, we augment the model \eqref{eq:brightnessIncrementGrad} with a translation warp $\hat{\Lum}(\Warp(\bx;\bp))$, where $\Warp(\bx;\bp) = \bx + \bp$, and $\bp$ denotes a small per-pixel translation.

Our composite objective (i.e., loss) function implies a joint optimization over the flow and alignment parameters: 
\begin{equation}
\label{eq:Ecomposite}
E(\velflow, \bp) \doteq E_\text{data}(\velflow, \bp; \cE) + E_\text{reg}(\velflow, \bp; \cE).
\end{equation}

The data-fidelity term measures the goodness of fit between the event data $\cE$ and its prediction with our model:
\begin{equation}
\label{eq:Edata}
E_\text{data} \doteq \left\| \frac{\Delta \hat{\Lum}}{\|\Delta \hat{\Lum}\|_2}(\bx) - \frac{\Delta \Lum}{\|\Delta \Lum\|_2}(\bx) \right\|_\gamma,
\end{equation}
where $\gamma$ is the $L^1$ norm (robust norm).
Since $C$ in \eqref{eq:brightnessIncrementEvents} is unknown, we compute the difference between normalized brightness increments (norms are over the pixel domain $\Omega$).

The regularizer penalizes the non-smoothness of the flow $\velflow$ and the magnitude of the per-pixel translation $\bp$:
\begin{equation}
\label{eq:Ereg}
E_\text{reg} \doteq \lambda_1 \|w(\bx)\, \nabla \velflow(\velparams(\bx)) \|_1 + \lambda_2 \|\bp(\bx)\|_1.
\end{equation}

The flow regularizer (first term in \eqref{eq:Ereg}) is explained in \cref{sec:flowregularizer}, after the flow parameterization is introduced. 
For the second term, the magnitude of $\bp$ is given by its $L^1$ norm over the pixel domain. %
In the experiments, we set $\lambda_1 = 0.5$ and $\lambda_2 = 0.1$.

\subsection{Physically-motivated Parameterization}
Swapping the mixed derivatives (Schwarz's theorem) in \eqref{eq:flowToTemporalDeliv}, the flow $\velflow \sim \prtl{}{\bx} \prtl{\rho}{t}$ is interpreted as the spatial gradient of $\prtl{\rho}{t}$. 
Thus \eqref{eq:flowToTemporalDeliv} admits two interpretations. 
($i$) from left to right: once estimated, the flow may be Poisson-integrated \cite{Barkas05poisson} to obtain $\prtl{\rho}{t}$, 
(as the best $L^2$ fit to the estimated flow \cite{Kim14bmvc,Zhang22pami}). 
($ii$) from right to left: the flow may be obtained as the spatial (e.g., Sobel) gradient of a scalar field $\prtl{\rho}{t}$. 
In contrast to most optical flow estimation methods, which parametrize $\velflow(\bx)$ directly in terms of its $x$ and $y$ components, we go one step further and exploit the above second interpretation of \eqref{eq:flowToTemporalDeliv} to parametrize the flow by means of $\velparams \equiv \prtl{\rho}{t}$, which we call the Poisson parameters of the flow.
This not only reduces the complexity of the problem (number of variables being optimized), thus conferring robustness, but also provides a strong link with the physical meaning of the variables: 
according to \eqref{eq:flowToTemporalDeliv}, the resulting flow actually represents the schlieren objects.
\Cref{fig:suppl:poisson} shows examples of the Poisson parameters $\velparams$.

\Cref{fig:blockdiagram} summarizes the visual quantities involved in the calculation of \eqref{eq:Edata}.
The candidate scalar parameter field $\velparams$ is converted (via Sobel operator) into the vector flow field $\velflow$. 
The flow $\velflow$ and translation field $\bp$ are used in the augmented model of \eqref{eq:brightnessIncrementGrad} to generate a predicted (i.e., modeled) brightness increment image. 
On the other hand, events $\cE$ are summed in \eqref{eq:brightnessIncrementEvents} and Gaussian-smoothed to produce a measured brightness increment image. 
The difference between the measured and predicted brightness increments provides an error signal that is used to drive the iterative refinement of the unknown variables $\bp$ and $\velparams$.

\subsection{Flow Regularizer}
\label{sec:flowregularizer}
We penalize the non-smoothness of the flow using a weighted Total Variation (TV) (see \eqref{eq:Ereg}).
As illustrated in \cref{fig:method:overlay}, it is difficult to estimate accurate flow in regions with very few events, 
which correspond to constant (e.g., zero) flow, hence we impose this prior knowledge as a regularizer to encourage zero flow therein.
Specifically, from the events $\cE$ we compute a Gaussian-smoothed histogram $h(\bx; \cE)=\sum_k \cN(\bx; \bx_k, \sigma^2)$ (with $\sigma = 5$ px) 
and normalize it to the range $[0,1]$. 
Then, we define weight function $w(\bx) \doteq 1 - \alpha / h(\bx; \cE)(\bx)$ (large in ill-posed regions with very few events), with $\alpha=0.95$ in the experiments.

\subsection{Optimization}
\textbf{Multi-scale}. For improved convergence, a coarse-to-fine patch-based approach is used for $\velflow, \bp$ and the loss function \eqref{eq:Ecomposite}.
The coarsest patch size is $64\times 64$ px and we use four resolution levels in a pyramidal fashion, resulting in finest patches of $8\times 8$ px.
To reach pixel density from the finest patches, we use bilinear interpolation.

\textbf{Implementation}.
\sblue{We use events in the fixed time interval (i.e., $120$ fps) for optimization across all sequences.}
As an optimizer, we use Adam \cite{Kingma15iclr} with 600 iterations.
The learning rate is set to 0.05, with the decay of 0.1.
The initialization of the first frame at the coarsest scale is: zero for $\bp$ and $\velflow$ (when applicable) and random in $[-1,1]$ for the Poisson parameters $\velparams$.
We found the latter to be better than also setting $\velparams$ to zero. 
Then, the initialization of the next levels uses the optimization results from the previous scales (i.e., coarse-to-fine approach).
\input{floats/supplfig_poisson_img}

\section{Physical Setup and Data}
\label{sec:implementation}

\subsection{Recording Setup}
\label{sec:recordingSetup}

\textbf{Co-capture System.}
To achieve high-quality recordings of frames and events, we build our own acquisition system.
Although some devices exist that record colocated events and frames (such as DAVIS \cite{Brandli14ssc,Taverni18tcsii}), their data quality (resolution, dynamic range, etc.) is limited and not suitable for BOS applications.
Our custom-built co-capture system consists of a frame camera (Basler acA1300-200um, 1280$\times$1024 px) and the latest generation event camera (Prophesee EVK3 Gen4, 1280$\times$720 px \cite{Finateu20isscc}), sharing the same optical axis by using a beamsplitter (Plate Bs C-Mount VIS50R/50T).
Both cameras are hardware-triggered for accurate synchronization and are calibrated to achieve accurate pixel alignment, following \cite{Muglikar21cvprw}.
\Cref{fig:cocapture} shows the camera system and an example of acquired data.
Further details about the used recording system can be found in \cite{Hamann22icprvaib}.

\textbf{Optical Setup.}
The field of view (FOV) of our cameras is limited by the beamsplitter ($\approx 15^\circ$), hence we set the distance between the cameras and the background to 3.3~m. 
We use randomly-generated background patterns that cover the whole FOV, where black dots (covering approximately 2 to 3 px in the image plane) are printed on a white paper.

The data quality also depends on the distance between the camera and the schlieren object.
The schlieren are more visible (larger pixel displacement $\Delta \bx$) by keeping $\dobj$ small (object closer to the camera).
At the same time, the camera system has to be focused both on the background pattern and the schlieren object, thus $\dobj$ cannot be too small.
We experimentally found distance $Z_D = 1.6$ m to be a good compromise between both opposing effects.
To control the scene brightness and achieve uniform illumination in the background, we use LED panels (four Eurolite LED PLL-360).
This illumination allows us to lower the aperture to an f-number of 10, leading to a higher depth of field.
Note that our beamsplitter setup leads to a 50\% split of the light reaching each camera of the acquisition system.

\subsection{Data Acquired}
\label{sec:dataset}

We record multiple sequences with natural and non-natural (forced) air convection, which are summarized in \cref{tab:dataset}.
For natural convection, we use heat sources, such as a hot plate, a hair dryer (switched off), and ice.
To demonstrate the HDR capabilities of event cameras, we record the data in 
($i$) bright conditions ($\approx$ 4000~\si{\lux}) and ($ii$) low-light conditions ($\approx$ 225~\si{\lux}).
The low-light condition is set to be darker than normal office lighting, which is a more natural condition for real-world applications.

\input{floats/tab_dataset}

Each sequence is approximately 10 to 20 seconds long and consists of events, frames and a calibration parameter file.
The recording starts with the scene in the absence of the schlieren object, which is useful for frame-based BOS methods (reference frame).
All sequences are recorded at normal room temperature ($\approx 24^\circ$C).
For the forced convection sequence of the running hair dryer, we set the event camera's refractory period to its minimum possible value to capture the fast dynamics of the airflow.
In total, we record nine sequences, each of which has up to 200M events.
\sblue{Regarding storage, events take about 1/10 of the data size required to store frames (e.g., 800 MB vs. 7.8 GB for a hot plate sequence).}
\input{floats/supplfig_dataset_gt}

Frames of sample sequences are shown in \cref{fig:suppl:dataset}.
Each frame is mapped from its original resolution (1280$\times$1024 px) to the event-camera resolution (1280$\times$720 px) (see \cref{sec:recordingSetup}).

Since we cannot obtain real ground truth (GT), we use frame-based estimated flow as GT flow (\cref{fig:suppl:dataset}).
The calculation of the flow is based on the classical Farneb\"ack algorithm \cite{Farnebaeck03scia} with four pyramidal scales at the frame rate (120~fps).
We test different parameters and find no significant difference on the quality of the results.
Before settling for Farneb\"ack's algorithm, we tested recent DNN-based state-of-the-art methods, such as \cite{Huang2022eccv1,Huang2022eccv2}, and found that they do not produce reasonable flow.
\Cref{fig:suppl:otherFlow} shows the comparison of several frame-based optical flow estimation methods:
two state-of-the-art optical flow and video-frame interpolation works \cite{Huang2022eccv1,Huang2022eccv2} and Farneb\"ack's method.
Due to the large gap between the training datasets of \cite{Huang2022eccv1,Huang2022eccv2} and our dataset, these recent DNN-based methods fail to estimate reasonable flow.
Farneb\"ack's algorithm works robustly and better, because
($i$) the background pattern is parallel to the image plane,
($ii$) the scene has no occlusions,
($iii$) the background pattern has clear and random edges that are useful to calculate the deformation between two frames.
Since we cannot determine the real GT, we do not explore a further analysis of frame-based estimation methods, which we leave for future research, such as simulation.
That is, to establish the first event-based BOS problem settings we leverage the knowledge of established frame-based BOS techniques.
Note that the quantitative evaluation is only based on the well-illuminated sequences, since the frame-based flow degrades in dark scenes (see \cref{sec:ablation:illumination}).
We publish the dataset and the code to compute the GT.

\input{floats/supplfig_other_flow}

\section{Experimental Evaluation}
\label{sec:experim}

This section reports the performance of the proposed estimation method and its properties.
First, we explain the baseline methods and evaluation metrics (\cref{sec:experim:metrics}).
Second, we benchmark the accuracy of all methods considered (\cref{sec:experim:flowAccuracy}).
Third, we show the capabilities of our method in low-light conditions (\cref{sec:experim:hdr}) and how it achieves high temporal resolution (1200 Hz ``slow motion'') in \cref{sec:experim:slowmo}, including a velocimetry application (\cref{sec:experim:velo}).
Finally, we analyze the proposed method further, especially regarding the dependency on frames (\cref{sec:ablation:illumination,section:ablation:frameFree}),
its sensitivity to hyper-parameters (\cref{sec:experim:sensitivity}),
\sblue{and the effect of event warping (\cref{sec:experim:deblur})}.

\subsection{Evaluation Metrics and Baseline Methods}
\label{sec:experim:metrics}

\textbf{Evaluation Metrics.}
We evaluate the proposed method in terms of optical flow $\velflow$ accuracy.
Two variants of the method are assessed:
($i$) using $\velparams$ as parameterization, from which we obtain $\velflow$ afterwards via \eqref{eq:flowToTemporalDeliv}, 
and ($ii$) using $\velflow$ directly.

The optical flow evaluation metrics are the average endpoint error (AEE),
the percentage of pixels with AEE $> 1$~px (denoted by ``\% Out''), 
and the angular error (AE).
We select the time interval (from 1 to 4~s) and region of interest (ROI) to remove objects, such as a hair dryer and a face from the scene.
All metrics are computed over pixels with at least one event inside the ROI.

\Cref{tab:suppl:benchmark} reports the detailed duration, ROI, and total number of events used for the benchmark.
The duration is selected such that the quality of schlieren is the best and stable.
For the ``Hair dryer (ON)'' sequence, we limit the height of the ROI due to extremely large number of events observed: otherwise, we set the ROI to have approximately 720$\times$512 px.
\input{floats/suppltab_benchmark_parameters}

\textbf{Baselines.}
As baseline flow estimators we use the two self-implemented methods from events because, to the best of our knowledge, there are no methods that estimate schlieren flow from event camera data.
\begin{itemize}
    \item The Multi-reference Contrast Maximization (MCM) \cite{Shiba22eccv} is a state-of-the-art optical flow estimation algorithm from events alone.
    It is a model-based method, hence there is no mismatch in the training dataset (due to our specific background pattern).
    We use the events between two consecutive frames (i.e., in a time span of 8.3 ms).
    \item Flow estimation from reconstructed intensity images: 
    we use E2VID \cite{Rebecq19pami} (a learning-based approach) to compute grayscale images from events and then apply the same (frame-based) optical estimator as the one for the GT.
    Images are reconstructed at 120 fps, i.e., the same frequency as the frames.
\end{itemize}

To the best of our knowledge, we found no methods with publicly-available implementation combining events and frames to estimate the optical flow,
we therefore believe this is a best-effort comparison. %
Also, notice that we do not train a Deep Neural Network (DNN) model with the supervisory GT flow, as the purpose of the paper is not a purely data-driven approach, but to develop an interpretable model-based method, by deriving a connection between the physical parameters and the data.

\subsection{Optical Flow Evaluation}
\label{sec:experim:flowAccuracy}

\input{floats/tab_full_result.tex}
\input{floats/fig_result_comparison}

Flow accuracy is reported in \cref{tab:fullResult}.
We evaluate on illuminated sequences for valid GT flows from frames (please see \cref{sec:experim:hdr} for the dark sequences).
Consistently for almost all sequences, the proposed method (``Ours (Poisson)'') provides the best accuracy compared with the baseline methods.
Due to the nature of schlieren, the GT flow magnitude has normally subpixel values.
Hence, we find that the angular error (AE) is a more reliable metric for the purpose of this benchmark.
The largest magnitude of the displacement ($\approx$ 3 px) is observed in the hotplate sequences. 
Still, it is remarkable that the proposed method achieves AEE $< 1$ pixel.
We acknowledge that the proposed method utilizes both event and frame data, while the baselines use only event data as input.
This is further discussed in \cref{sec:ablation:illumination}.

Also, it is noticeable that the Poisson-parameterized estimation (``Ours (Poisson)'') results in better accuracy than the flow-parameterized estimation (``Ours (Flow)'').
This clearly states the effectiveness of our physically-motivated parameterization.
It provides not only a smaller number of parameters, as discussed in \cref{sec:proposedMethod}, but also contributes with better accuracy.

Additionally, we observe that the forced  convection usually has a smaller displacement magnitude than natural convection.
This is because the optical flow $\velflow$, which we evaluate on, is the temporal derivative of the density gradient.
In the forced convection case (e.g., hair dryer (ON)), the spatio-temporal changes of the air density at a pixel might be smaller than in the natural, heat-induced schlieren, since the advection of the flow is dominant, which can be seen as nearly constant. %

\Cref{fig:main:compare} shows qualitative results.
Although the GT flow is based on a classical, general-purpose estimation method, it provides remarkably reasonable flow.
The baseline methods (MCM and E2VID) fail to estimate reasonable flow from events.
Especially, we find the alignment-based method \cite{Shiba22eccv} fails to estimate schlieren flow.
This is because most events are generated at the edges of the background pattern, resulting in an uneven spatial distribution despite air being actually moving, and consequently, triggering more flickering events.
The E2VID-based method surprisingly reconstructs edge structures of the background pattern (see also \cref{section:ablation:frameFree}) in spite of this specific (flickering) event input, and estimates comparable flow.
However, it fails to recover the fine structure of the flow.
Finally, the flow estimated by our method resembles the GT flow the most, and it even seems to capture more fine-scale (high-frequency) structures.

\subsection{HDR Experiment}
\label{sec:experim:hdr}

So far we have established that the proposed method is able to recover the fine flow structure of the schlieren object.
However, schlieren based on events has another interesting aspect:   %
as shown on the left column of \cref{fig:main:compare}, the existence of schlieren is already visible in the event data histogram.
By contrast, the schlieren structure is not visible to the naked eye on the raw frame data but only as the result of optical flow processing.
The fact, that schlieren is observable in a more direct way using events, allows us to leverage the advantages of the event camera itself, such as HDR and high temporal resolution.

\input{floats/fig_hdr}

\Cref{fig:hdr} shows qualitative results of the frame-based and event-based schlieren imaging under poor illumination.
The frame-based schlieren method fails to estimate realistic flow under such conditions, as it needs intense lighting sources, especially if high-speed cameras are used.
Due to the insufficient brightness, the quality of the frames collapses even after normalization (i.e., using the entire grayscale range).
On the other hand, the event data captures the schlieren structure (\cref{fig:hdr}, top right).
Furthermore, the proposed algorithm combining events and frames is surprisingly robust against such low-quality image inputs.
Using natural light (225 \si{\lux}) the result (\cref{fig:hdr}, bottom right) shows the potential of event cameras to push the limits for future BOS applications.
We further discuss the effect of the amount of illumination in \cref{sec:ablation:illumination}.

\subsection{Super-Slow Motion}
\label{sec:experim:slowmo}

Event-based BOS also enables us to see the schlieren at markedly higher temporal resolution (i.e., slow motion) than conventional frames.
To this end, we conduct a streak-schlieren analysis \cite{settles2022schlieren}.
The streak analysis focuses on a single column of the schlieren image to see how it evolves in time, by showing an $x-t$ diagram (kymogram) of the air convection.
The frame-based schlieren method uses for example Poisson images as schlieren images.  %
For event-based methods, schlieren images can be either Poisson images or simply event histograms.
\Cref{fig:kymogram} shows a comparison of kymograms obtained from frames at 120 Hz (the frame rate) and obtained from events (10$\times$ higher rate, i.e., at 1200~Hz).
Event-based BOS can provide high temporal resolution kymograms due to the asynchronous nature of event data.
Compared with the frame-based analysis (\cref{fig:kymogram:frames}), the event-based one (\cref{fig:kymogram:events}) shows thinner lines of schlieren in space-time.
The slow motion schlieren visualization is best viewed in the supplementary video.

\input{floats/fig_kymogram}

\subsection{Velocimetry}
\label{sec:experim:velo}

One can perform velocimetry by fitting curves to the kymograms \cite{settles2022schlieren}.
Let us analyse the speed of propagation of schlieren ($\partial{\rho} / \partial{t}$ in the case of Poisson image) along one direction (e.g., vertical).
\Cref{fig:kymogram:events} shows an example on the hot plate sequence.
By fitting a curve (line), the flow propagates 166 px during approximately 68.8 milliseconds.
The geometry of the BOS setup (focal length $f = 25\si{\milli\meter}$, distance to object $Z_A = 1.7\si{\meter}$, pixel size $4.86\si{\micro\meter}$) leads to an approximate velocity of \SI{0.8052193}{\meter/\second}.

\subsection{Dependency on Frames}
\label{sec:ablation:illumination}

\input{floats/supplfig_declining_light}
The proposed method uses events and frames.
Naturally, the question arises to which extent the algorithm relies on which signal.
To this end, we present the ablation study with different brightness levels (see also \cref{sec:experim:hdr}).
\Cref{fig:suppl:lowLight} shows the qualitative results for both: frame-based method and our method (frame plus events), for different illumination levels (measured with a Voltcraft MS-1300 light meter).
As clearly shown, the frame-based flow (column (b)) starts to deteriorate when the illumination is 1000~\si{\lux} or smaller.
For a better performance, we even normalize the range of the frames used (the exposure time is fixed to maintain the frame rate of 120 fps).
However, this does not provide significantly better results that can compete with those of our method.
By contrast, the following two points are remarkable about our method:
($i$) schlieren is still visible at 110~\si{\lux} in the event histograms, indicating the HDR capabilities of the noisy input data (column (c)), and
($ii$) the estimated flow (column (d)) still looks reasonable when the illumination is as low as 225~\si{\lux}, despite our method using the naturally darker frame as an input (column (a)).
Note that our method does not work when the frame is completely black (less than 50~\si{\lux}).
All the above indicates that the proposed method requires frames, but it can overcome the limited dynamic range of the frames due to the HDR advantages of event cameras.

\subsection{Towards a Frame-Free Method}
\label{section:ablation:frameFree}

The proposed method utilizes the information from events and a frame, however the quality of the frame data does not need to be the best, as shown in the previous section.
Hence, an interesting challenge is to replace frame data with intensity reconstruction from events, such that the proposed method could be extended to be \emph{frame-free}.
To this end, \Cref{fig:frameFree} shows the comparison of the different input frames.
Instead of using an acquired frame as an input to the proposed method,
we reconstruct intensity images using E2VID \cite{Rebecq19pami} and feed them as input.
Despite the large visual difference between the two different inputs, the output flow and Poisson images seem to have similar structures.
Although we do not further investigate the quality of the intensity reconstruction, the results show future possible extension towards frame-free event-based BOS methods.

\input{floats/fig_e2vid_framefree.tex}

\subsection{Effect of the Regularizers and the Translation Field}
\label{sec:experim:sensitivity}

\textbf{Ablation.}
To assess the importance of the regularization and the translation field parameters $\bp$, we conduct an ablation study.
The top half of \cref{tab:suppl:ablation} reports optical flow accuracy of the proposed method, the one without regularization, and the one without the translation.
There is a significant improvement due to the regularizers:
without regularizers, the estimated $\velparams$ and $\bp$ become not smooth anymore, which leads to irregular flow estimation.
The effect of $\bp$ is relatively minor, but still noticeable.

\input{floats/suppltab_ablation}

\textbf{Sensitivity Analysis.}
We test different weights for each regularizer $\lambda_1, \lambda_2$ in \eqref{eq:Ereg}.
The weights are set as follows:
we fix one parameter ($\lambda_1 = 0.5$), and vary $\lambda_2$ between $0.01$ and $1.0$; 
then we fix the other parameter ($\lambda_2 = 0.1$) and vary $\lambda_1$ between $0.05$ and $1.0$.
The flow accuracy is reported in the bottom half of \cref{tab:suppl:ablation}.
We observe on-par accuracy when $\lambda_1 = 1.0$ with respect to the base condition (the top row).

\sblue{
\subsection{Effect of Event Warping}
\label{sec:experim:deblur}
In some of the previous event-based flow estimation methods \cite{Paredes21cvpr,Zhang22pami},
warping the events using the estimated optical flow produces sharp intermediate images that improve convergence (e.g., of the image reconstruction task).
We test the possible efficacy of such an event warping in the BOS setting.
As shown in \cref{fig:deblur}, warping events with the estimated flow does not have a sharpening effect in schlieren.
This is because:
($i$) event warping leads to sharpening \emph{if} events are generated by moving edges (e.g., \cite{Shiba22eccv}), which does not hold true in BOS (see also the large errors of the ``MCM'' method in \cref{fig:main:compare,tab:fullResult}),
and ($ii$) the time window between consecutive frames (e.g., at 120 fps) is small enough to already produce sharp brightness increment images.
Hence, we do not examine further sharpness/deblur-based approaches.
}
\input{floats/fig_abl_deblur}

\sblue{
\section{Studies on a Helium Jet Experiment}
\label{sec:simulation}
The theoretical connection between events and schlieren that we establish in \eqref{eq:flowToTemporalDeliv} highlights an important difference in the experimental settings between frame-based and event-based BOS:
event-based BOS focuses on the temporal derivative of the air density, which requires a time-dependent flow.
Steady flows that are often used for frame-based BOS (e.g., supersonic flow) do not satisfy the time dependency.
More precisely, ``steady'' flow in frame-based BOS means that it can be averaged over multiple frames (i.e., ``$\partial{\rho} / \partial{\bx}$'' in \cref{tab:schlierenComparison}) to improve the signal-to-noise ratio.
This difference makes it difficult to apply the experimental knowledge of common frame-based BOS data to the event-based one.
Nonetheless, we conduct a simulation experiment to make a stronger connection between these complementary schlieren methods.
To this end, we use a frame-based BOS dataset that is well documented through in experimental evaluation \cite{settles2022schlieren}.
The dataset consists of high-speed frames ($6000$~fps) recording a round turbulent helium jet in air, whose flow has self-similarity (axisymmetric) properties.
Each sequence has 3000 images, of 1024 $\times$ 512 px and with 167\si{\micro\second} exposure time, i.e., 0.5\si{\second} of data.
}

\input{floats/fig_helium_esim_single_col}

\sblue{
We analyze the velocimetry in accordance with the self-similarity property using simulated events.
The analyzed sequence is the jet with Reynolds number $Re_d = 5,980$.
First, we run ESIM \cite{Rebecq18corl} to simulate events from the high-speed frames.
We use a contrast sensitivity of $0.05$ for both positive and negative events to have a reasonable signal-to-noise ratio.
Events are shown in \cref{fig:heliumSingleCol:events}.
Then, velocimetry (as in \cref{sec:experim:velo}) provides the velocity at different pixel locations on the image plane.
Instead of manually fitting lines to the kymogram using a drawing software as in \cite{settles2022schlieren},
we detect lines automatically, as follows:
first, we extract patches along the $x/d$ axis of the image plane (e.g., 100$\times$100 px);
then, we smooth the patches with a Gaussian filter and estimate the slope of the dominant direction within each patch iteratively, by rotating the patch and finding the angle that maximizes the magnitude of the rotated patch gradient in a predefined direction.
}

\sblue{\Cref{fig:heliumSingleCol:similarity} shows the self-similarity by analyzing the simulated events.
Similarly to \cite[Fig.~7]{settles2022schlieren}, the estimated velocity is symmetric and consistent with the theoretical values (dashed line) along the relative distance using the jet spreading rate.
We observe larger error accumulation than the purely frame-based method in \cite{settles2022schlieren}, which can be attributed, among other factors, to simulation inaccuracies.
\Cref{fig:heliumSingleCol:velocity} shows the comparison of the velocity at the center of the jet nozzle along with the distance.
Note that the observed jet centerline velocity is the convective velocity of large scale turbulent structures (viewed along the line of sight), which is not necessarily the mean fluid velocity.
Nonetheless, the velocity values are similar and reasonably close to the results in \cite[Fig.~8]{settles2022schlieren}.
Moreover, they seem to follow the $1/x$-type decay, which agrees with the theoretical model of turbulent flow:
The frame-based results in \cite[Fig.~8]{settles2022schlieren} degrade approximately for $x/d \leq 80$, while the simulated event-based results
provide a better fit here.
}

\sblue{Kymograms are shown in \cref{fig:esimVelo:kymogram}.
The curve patterns in the event-based kymogram look consistent with the frame-based kymogram in \cite[Fig.~9]{settles2022schlieren}, although they show some artifacts due to the simulation.
As mentioned in \cite{settles2022schlieren}, even $6000$~fps is not enough for the pixels close to the nozzle (e.g., $x/d < 20$, where the flow is very fast.
Hence, the event data that is simulated using the frames inherit such a limitation: the kymogram becomes noisier for smaller $x/d$ (closer to the nozzle).
The high temporal resolution of event cameras could help overcome such a rate limit of frame-based BOS.
}

\input{floats/fig_esim_kymogram}

\section{Discussion}
\label{sec:discussion}

\sblue{
Let us summarize some findings of the first event-based BOS technique.
\begin{itemize}
    \item The megapixel race has enabled new applications for event cameras in the field of fluid dynamics.
    \item Event-based BOS can capture the spatio-temporal derivative of the media density without any particle seeding.
    \item As event cameras only sense the temporal changes of the scene, they capture incremental changes of schlieren, as opposed to changes with respect to a reference frame. 
    \item Schlieren can be observed directly in the form of flickering events, %
    while it is more indirect to perceive in frames.
    \item Event-based BOS can overcome the limitations of dynamic range and temporal resolution of traditional frame-based BOS.
    \item Due to the data sparsity, event-based BOS is approximately ten times more data efficient (smaller storage size). 
    However, estimating the density fluctuations is more challenging than using frames.
    \item The knowledge of the background pattern (frames) can be used to handle the problems of sparsity and noise.
    \item The proposed method casts the problem into an optical flow estimation one (i.e., in the realm of computer vision).
    \item State-of-the-art optical flow methods %
    (e.g., DNN) do not work well for schlieren because of the very different data properties (e.g., edge motions vs.~complex density fluctuations).
    \item Directly parameterizing the temporal derivative of the density reduces the complexity of the problem, provides a link with the physical meaning of the variables and yields the most accurate results.
    \item Event-based BOS enables high-temporal resolution velocimetry analysis using kymograms.
\end{itemize}
}

\textbf{Limitations}.
The proposed BOS technique using events shows advantages over frame-based BOS in terms of HDR capabilities and temporal resolution, lowering the demand for bright illumination and high-speed cameras.
However, in other aspects, it inherits the limitations of frame-based BOS. 
Optically, the estimated brightness gradient is a mean value integrated along the optical axis, and the technique inherently has a trade-off between the observed displacement and the obtained sharpness of the gradient under investigation.

Additionally BOS is sensitive to vibrations, due to the underlying assumption that the small perceived changes are only caused by refractive index variations.
Specific to event cameras is that the signal is noisy, and careful tuning of the camera's biases is necessary.
The proposed method furthermore relies on a combination of events and frames, thus an accurate spatio-temporal alignment of both data sources is required.
The flow estimation method does not run in real time. However, raw events visualized as histograms can be computed online and resemble schlieren images.
While the proposed multi-scale approach improves convergence of the optimization, it limits the spatial resolution of the flow, which is a similar limitation as in frame-based BOS.

\sblue{
Since event-based BOS relies on temporal changes, flows are observed well if they are time-dependent. 
This property makes it difficult to use steady flows to validate event-based BOS, and hence to evaluate frame- and event-based BOS in strictly the same settings.
Also, event-based BOS cannot improve the signal-to-noise ratio like the frame-based one does by averaging multiple frames.
Frame-based BOS and event-based BOS have their own strengths, which may be complementary (e.g., frame-based BOS is good for steady flows, whereas event-based BOS is good for time-dependent flows). 
This could be further investigated in the future.
}

\section{Conclusion}
\label{sec:conclusion}

We have presented the first event-based BOS imaging technique and an algorithm to estimate the temporal derivative of the air density gradient. 
The approach has been obtained in a mathematically rigorous way and has a physically-motivated parameterization. 
Using the frame-based method in well-lit conditions as ground truth, the experiments evidenced that our approach outperforms all other tested methods.
We furthermore illustrated how the advantages of event cameras could be leveraged for BOS applications, lowering the requirements for high illumination and visualizing the turbulent eddies at a significantly higher temporal resolution.
\sblue{Additional studies on a high-speed schlieren dataset of a helium jet provided further validation of the method.}
We release the code and dataset to the public and hope that this research opens up new possibilities for the computer vision and fluid dynamics communities.

\section*{Acknowledgment}

We would like to thank Prof. Dr. A. Liberzon at Tel Aviv University for useful discussions and advice.
We also thank the anonymous reviewers for valuable suggestions. 
This research was funded by
the German Academic Exchange Service (DAAD), Research Grant-Bi-nationally Supervised Doctoral Degrees/Cotutelle, 2021/22 (57552338)
and the Deutsche Forschungsgemeinschaft (DFG, German Research Foundation) under Germany’s Excellence Strategy – EXC 2002/1 ``Science of Intelligence'' – project number 390523135.

\ifarxiv
\balance
\fi

\bibliographystyle{IEEEtran}
\bibliography{all,refs}

\begin{IEEEbiography}[{\includegraphics[width=1in,height=1.25in,clip,keepaspectratio]{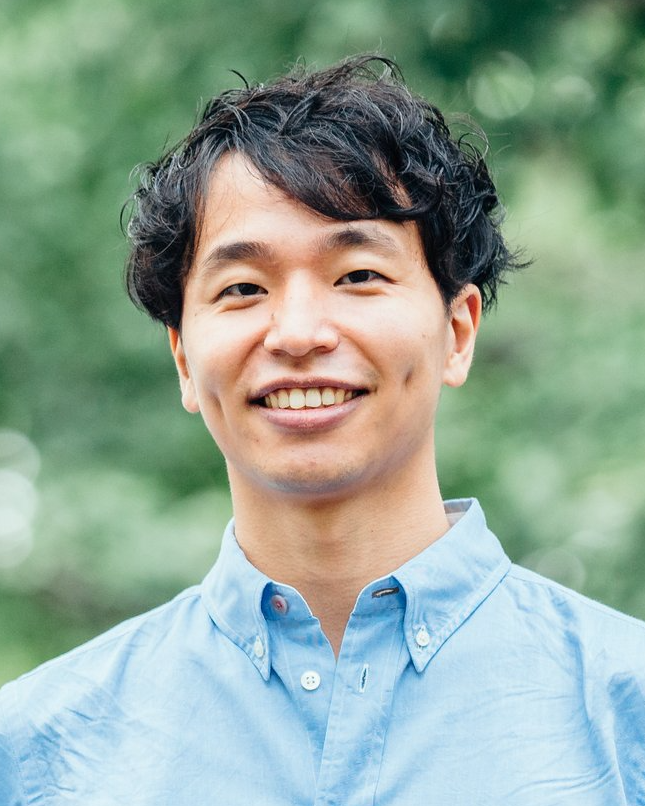}}]{Shintaro Shiba} 
received the Ph.D. degree in Engineering from Keio University, Japan, in 2023.
From 2021 to 2023, he was a visiting researcher at the Robotics Interactive Perception Laboratory, Technische Universität Berlin,
supported by a German Academic Exchange Service (DAAD) bi-national doctoral research fellowship.
He received his Master's in Cognitive Neuroscience from the University of Tokyo in 2017.
His research interests include computer vision, machine learning, robotics, and neuroscience.
\end{IEEEbiography}

\begin{IEEEbiography}[{\includegraphics[width=1in,height=1.25in,clip,keepaspectratio]{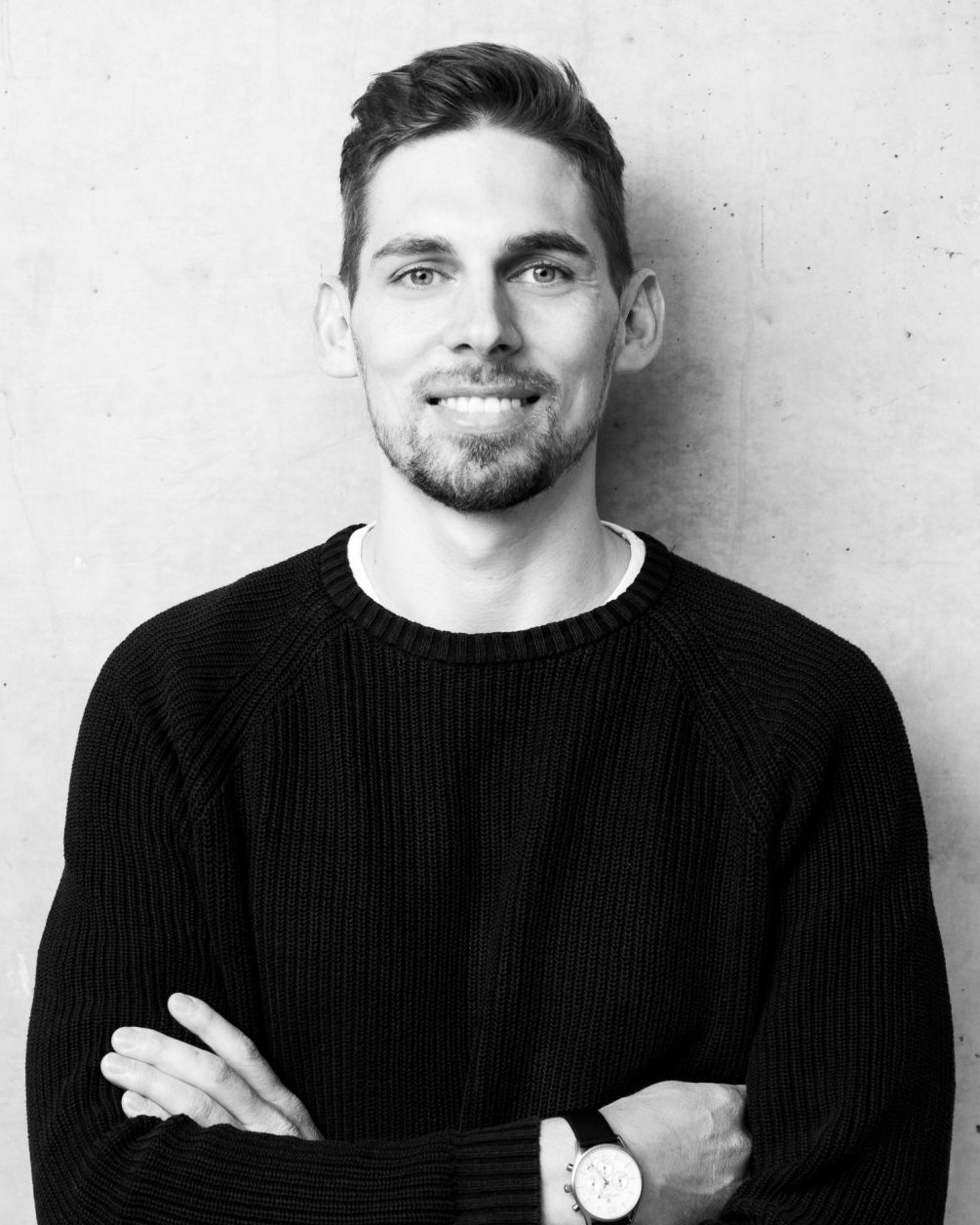}}]
{Friedhelm Hamann} is a Ph.D. student at the Robotic Interactive Perception Laboratory, Technical University Berlin, and a member of the Science of Intelligence Excellence Cluster, Berlin, Germany.
He received a master's degree in Electrical Engineering, Computer Engineering, and Information Technology from RWTH Aachen University in 2021. His research interests include computer vision, machine learning, and signal processing.
\end{IEEEbiography}

\begin{IEEEbiography}[{\includegraphics[width=1in,height=1.25in,clip,keepaspectratio]{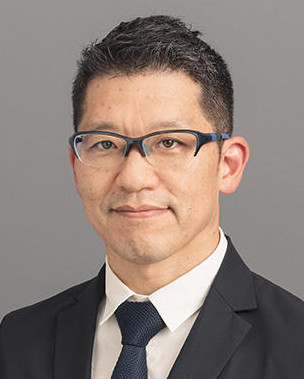}}]{Yoshimitsu Aoki} 
received the Ph.D. degree in engineering from Waseda University in 2001. From 2002 to 2008, he was an Associate Professor with the Department of Information Engineering, Shibaura Institute of Technology. He is currently a Professor with the Department of Electronics and Electrical Engineering, Keio University. He performs research in the areas of computer vision, pattern recognition, and media understanding.
\end{IEEEbiography}

\begin{IEEEbiography}[{\includegraphics[width=1in,height=1.25in,clip,keepaspectratio]{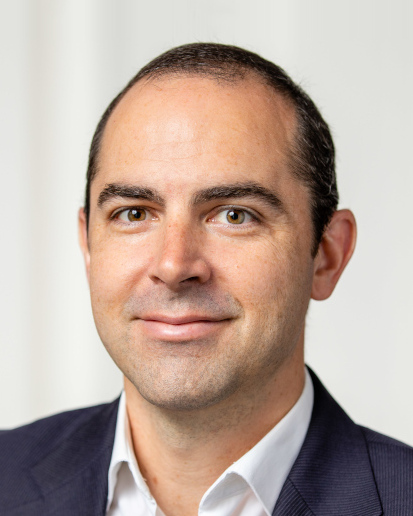}}]{Guillermo Gallego} (SM'19) 
is Associate Professor at Technische Universit\"at Berlin, in the Dept. of Electrical Engineering and Computer Science, and at the Einstein Center Digital Future, Berlin, where he leads the Robotic Interactive Perception Laboratory.
He is also a Principal Investigator at the Science of Intelligence Excellence Cluster, Berlin, Germany.
He received the PhD degree in Electrical and Computer Engineering from the Georgia Institute of Technology, USA, in 2011, supported by a Fulbright Scholarship.
From 2011 to 2014 he was a Marie Curie researcher with Universidad Politecnica de Madrid, Spain, and from 2014 to 2019 he was a postdoctoral researcher at the Robotics and Perception Group, University of Zurich and ETH Zurich, Switzerland.
Since 2022 he is also Co-Director of the HEIBRiDS interdisciplinary research school, Berlin.
His research interests include robotics, computer vision, signal processing, optimization and geometry. 
\end{IEEEbiography}

\end{document}

%% file: macros.tex
\def\Lum{L}
\def\tref{t_\text{ref}} %
\def\pol{p} %
\def\prtl#1#2{\frac{\partial#1}{\partial#2}}
\def\prtlsecond#1#2{\frac{\partial^2#1}{\partial#2^{2}}}

\def\cE{\mathcal{E}} %
\def\numEvents{N_e} %
\def\Warp{\mathbf{W}}

\def\bx{\mathbf{x}}

\def\pol{p}
\def\velflow{\mathbf{v}}
\def\velparams{\mathbf{q}}

\def\cN{\mathcal{N}} %

\def\bp{\mathbf{p}}

\def\beps{\boldsymbol{\epsilon}}
\def\dobj{Z_A}

\newcommand{\bnum}[1]{\bfseries #1}

\newcommand{\novalue}{{\textendash}}

\definecolor{light-gray}{gray}{0.6}
\newcommand\gframe[1]{{\color{light-gray}\frame{#1}}}

\newcommand{\sblue}[1]{#1}

%% file: floats/fig_eyecatch.tex
\def\figmethodwidth{.48\linewidth}
\begin{figure}[t]
 {\includegraphics[clip,trim={2cm 2cm 5cm 3cm},width=\linewidth]{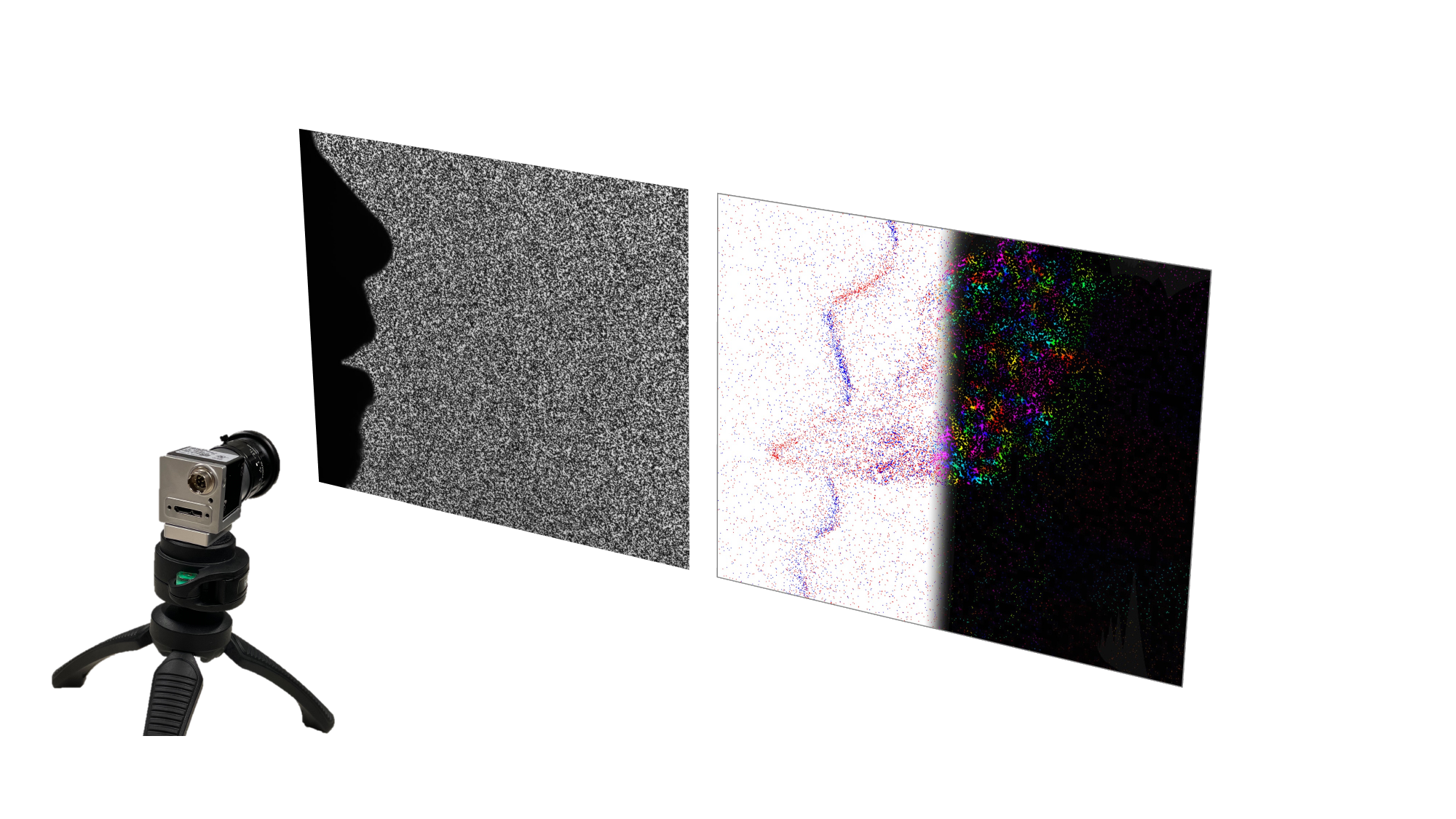}}
\caption{In background-oriented schlieren (BOS) imaging local density gradient variations between a camera and a background pattern lead to tiny perceived changes on the image plane. 
We present %
how to combine events and frames to observe schlieren in the scene
and how to leverage the advantages of event cameras to visualize gas streams such as the human breath.}
\label{fig:eyecatch}
\end{figure}

%% file: floats/tab_schlieren_comparison.tex
\begin{table}[t!]
\centering
\caption{Comparison of various schlieren imaging techniques and the physical quantities they measure.}
\adjustbox{max width=\columnwidth}{%
\begin{tabular}{lc}
\toprule
Method & Observation %
\\ 
\midrule
Shadowgraphy \cite{Hooke1665shadowgraphy,raffel2015background,Gardner2020bos} & $\displaystyle \prtlsecond{\rho}{\bx}$ \\[2ex]
Toepler's schlieren photography \cite{krehl1995august,raffel2015background} & $\displaystyle \prtl{\rho}{\bx}$ \\[2ex]
Laser speckle photography \cite{raffel2015background} & $\displaystyle \prtl{\rho}{\bx}$ \\[2ex]
Frame-based BOS \cite{raffel2015background} & $\displaystyle \prtl{\rho}{\bx}$ \\[2ex]
Event-based BOS (this work) & %
$\displaystyle \frac{\partial^2 \rho}{\partial t \partial\bx}$ \\[1.5ex]
\bottomrule
\end{tabular}
\label{tab:schlierenComparison}
}
\end{table}

%% file: floats/fig_bos_setting.tex
\begin{figure}[t]
  \centering
  {\includegraphics[clip,trim={1.5cm 3cm 1cm 3.5cm},width=\linewidth]{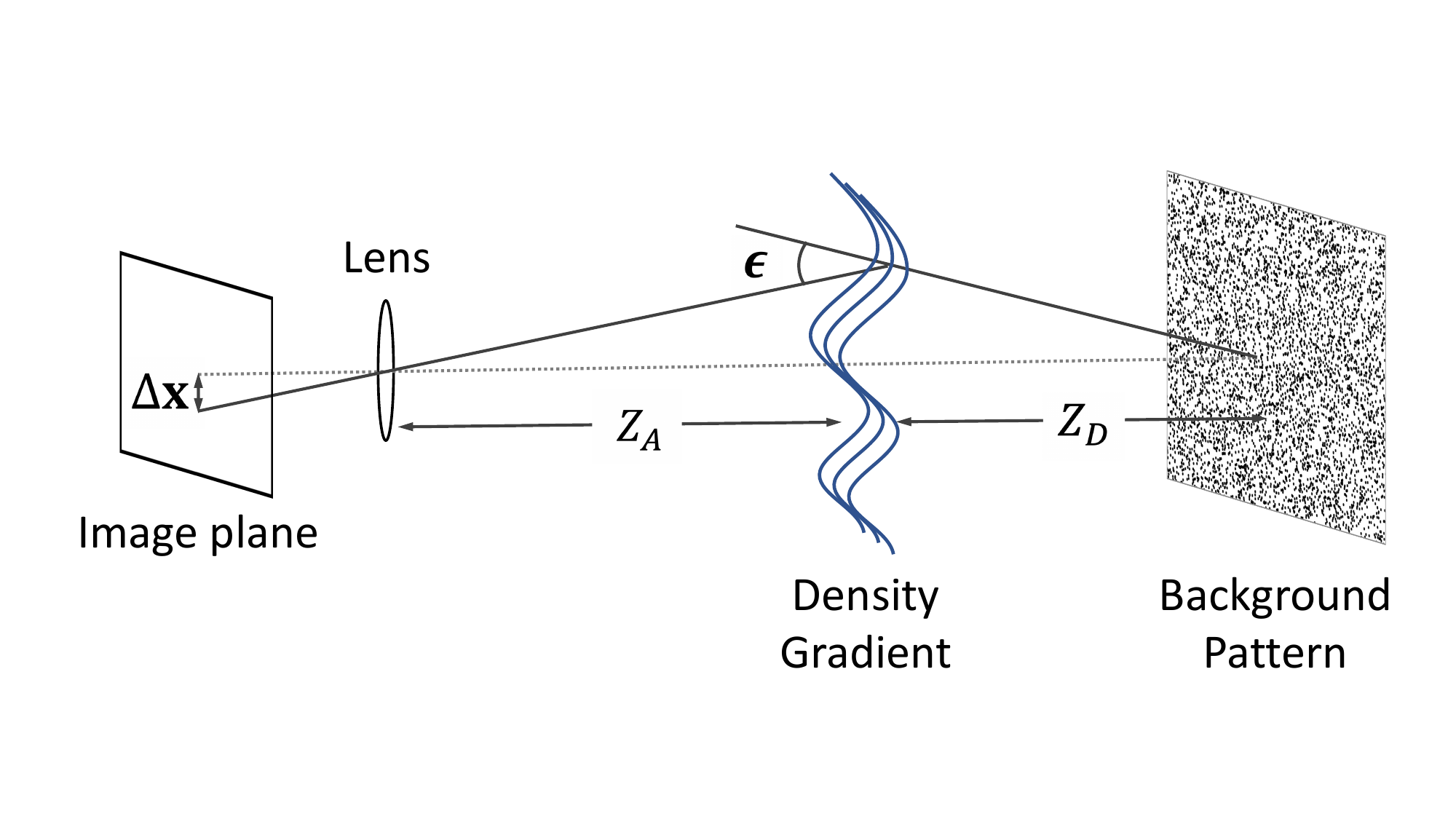}}
\caption{Background-Oriented Schlieren (BOS) setup.}
\label{fig:method:bos}
\end{figure}

%% file: floats/fig_bos_principle_comparison.tex
\begin{figure}[t]
  \centering
  {\includegraphics[clip,trim={0 5.5cm 11.5cm 0},width=\linewidth]{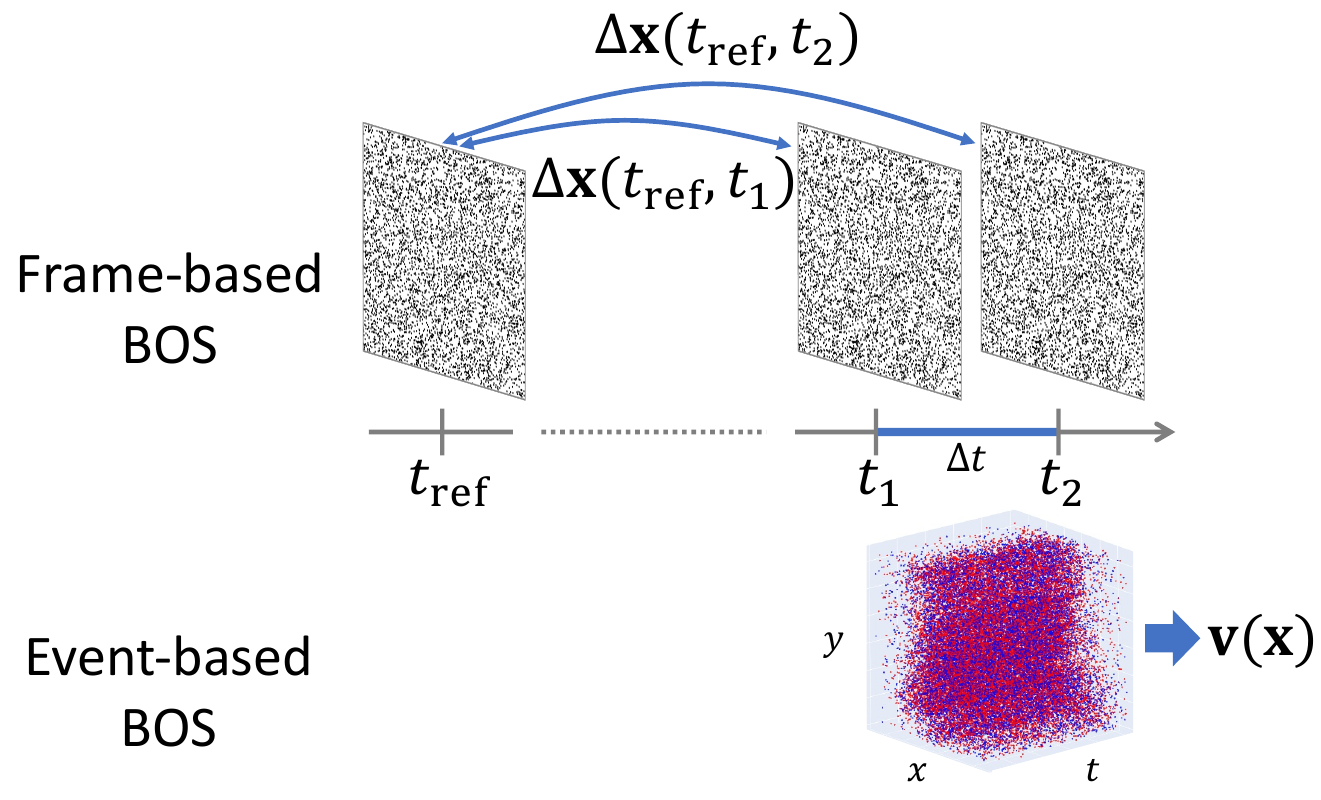}}
\caption{Frame-based BOS and event-based BOS.}
\label{fig:method:bosPrincipleComparison}
\end{figure}

%% file: floats/fig_method_cocapture.tex
\begin{figure}[t]
\centering
\begin{subfigure}{.48\linewidth}
  \centering
  {\includegraphics[trim={1.3cm 1.7cm 5.8cm 2.3cm},clip,height=0.72\linewidth]{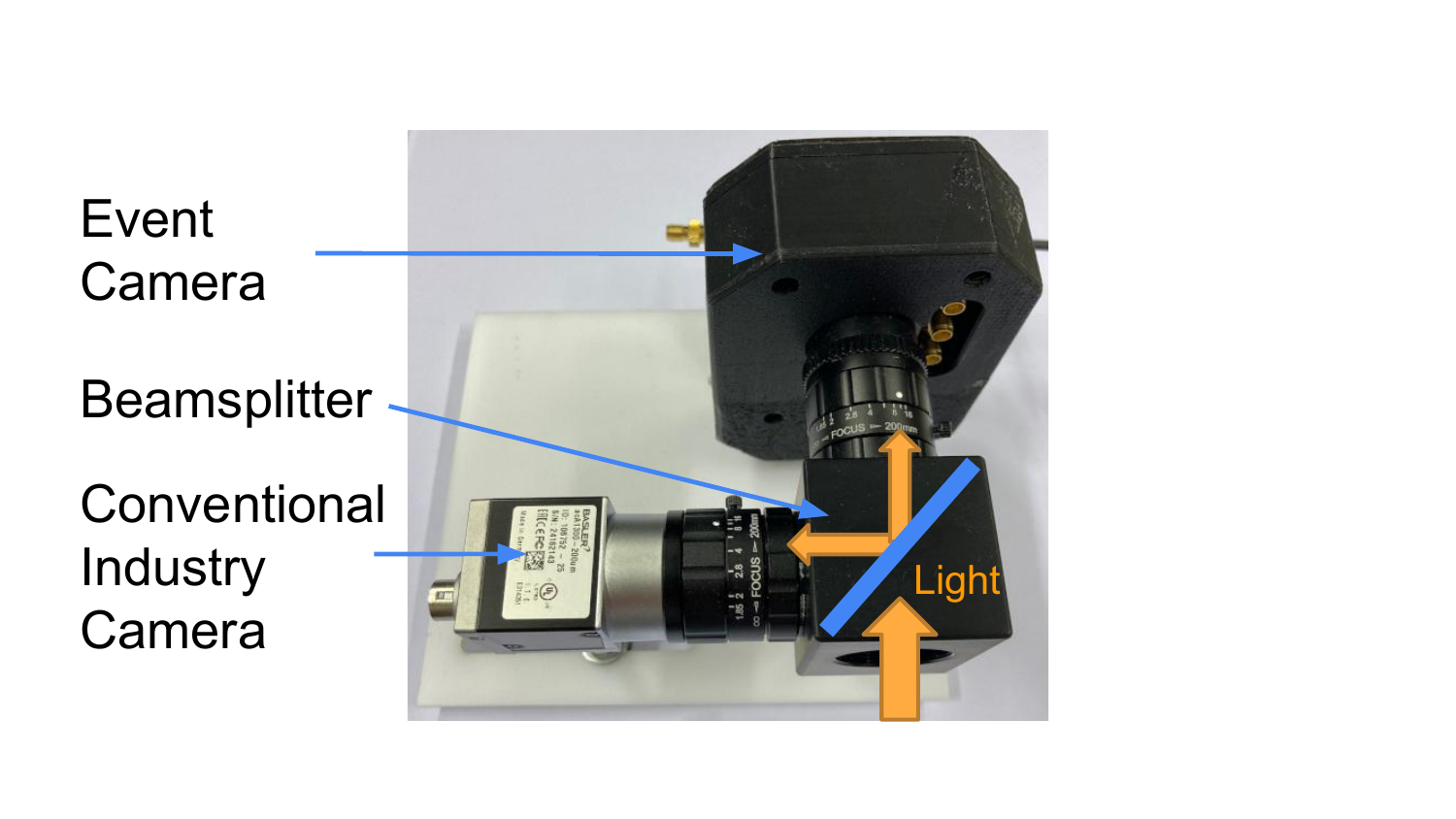}}
  \caption{\centering Recording system}
  \label{fig:method:cocapture}
\end{subfigure}\;\;
\begin{subfigure}{.48\linewidth}
  \centering
  {\includegraphics[height=0.72\linewidth]{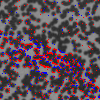}}
  \caption{\centering Events and frame}
  \label{fig:method:overlay}
\end{subfigure}
\caption{(a) Actual synchronized data recording system, combining an event camera and a frame-based camera via a beamsplitter (\cref{sec:recordingSetup}). (b) Data: events (red and blue, colored according to polarity) during a short time window overlaid on a grayscale frame (a $100\times 100$ pixel region for better visualization).}
\label{fig:cocapture}
\end{figure}

%% file: floats/fig_block_diagram.tex
\begin{figure*}
    \centering
    {\includegraphics[trim={0 1.65cm 9.5cm 3.89cm},clip,width=.85\linewidth]{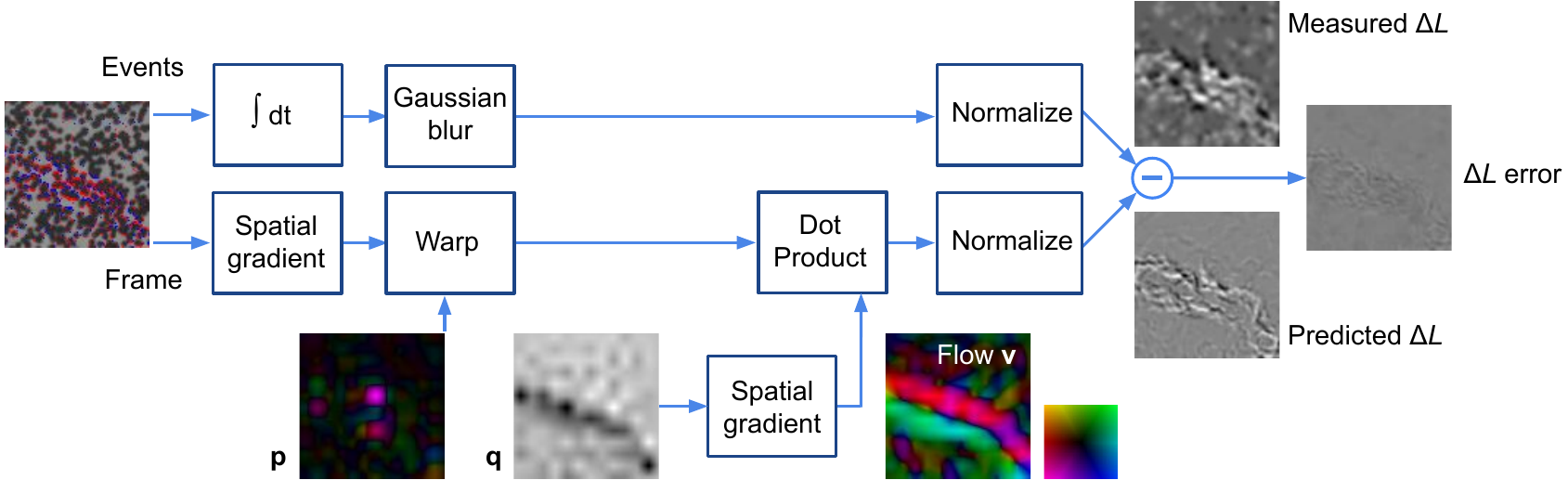}}
    \caption{Block diagram of the objective $E_\text{data}$ in \eqref{eq:Edata}.
    On the top branch, events are integrated in time using \eqref{eq:brightnessIncrementEvents} and smoothed with a Gaussian kernel ($\sigma=2$ px) to produce the measured brightness increment image $\Delta \Lum$.
    The bottom branch shows how to compute the predicted brightness increment $\Delta \hat{\Lum}$ from the frame and the problem unknowns:
    the translation field $\bp$ and the Poisson parameters of the flow, $\velparams$.
    The flow $\velflow$ and $\bp$ are pseudo-colored (see color wheel). 
    Same data as \cref{fig:cocapture}.
}
    \label{fig:blockdiagram}
\end{figure*}

%% file: floats/supplfig_poisson_img.tex
\def\figmethodwidth{.4\linewidth}
\begin{figure}[t]
\centering
\begin{subfigure}{\figmethodwidth}
  \centering
  {\includegraphics[clip,trim={12cm 3cm 12cm 1cm},width=\linewidth]{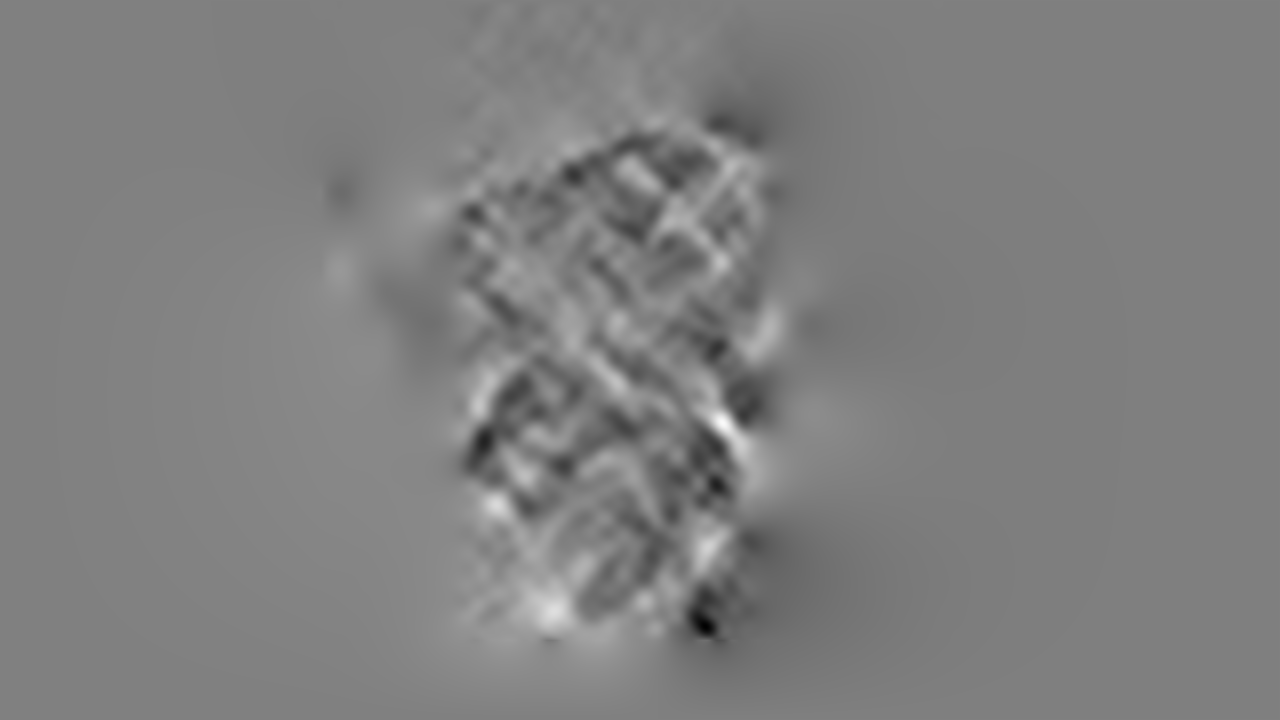}}
\end{subfigure}\;
\begin{subfigure}{\figmethodwidth}
  \centering
  {\includegraphics[clip,trim={12cm 3cm 12cm 1cm},width=\linewidth]{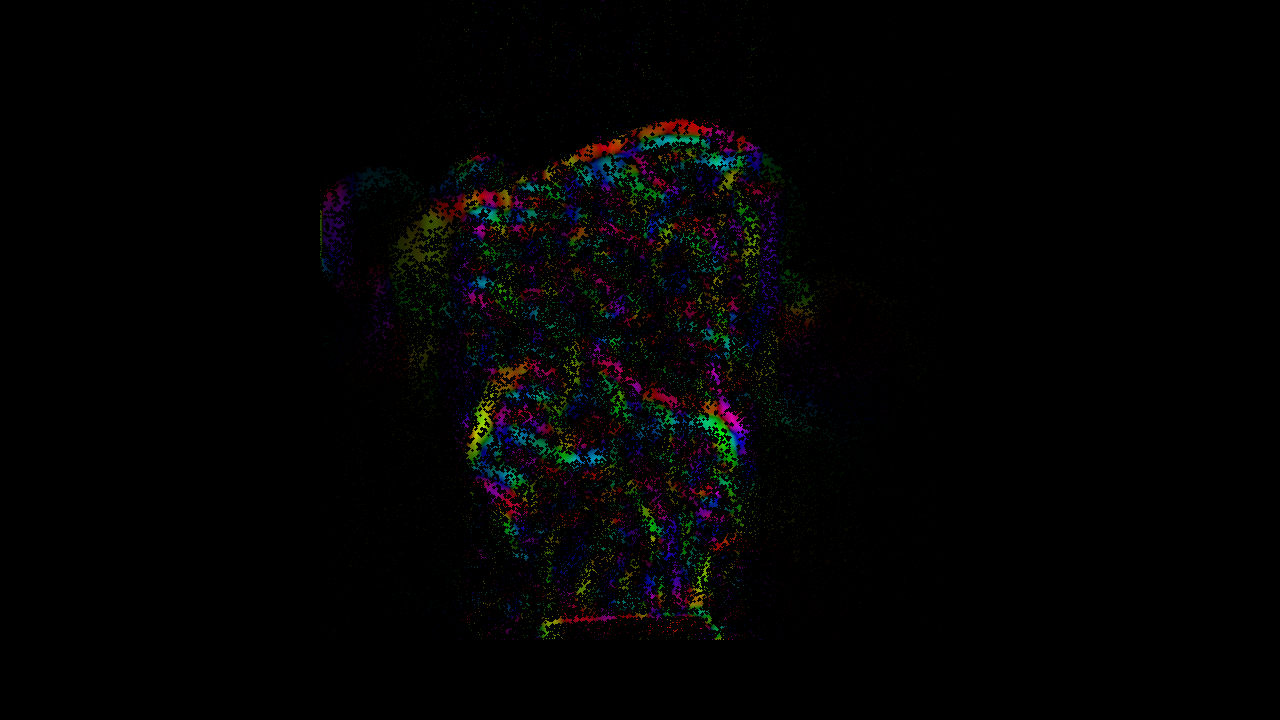}}
\end{subfigure}\\[0.5ex]
\begin{subfigure}{\figmethodwidth}
  \centering
  {\includegraphics[clip,trim={12cm 3cm 12cm 0cm},width=\linewidth]{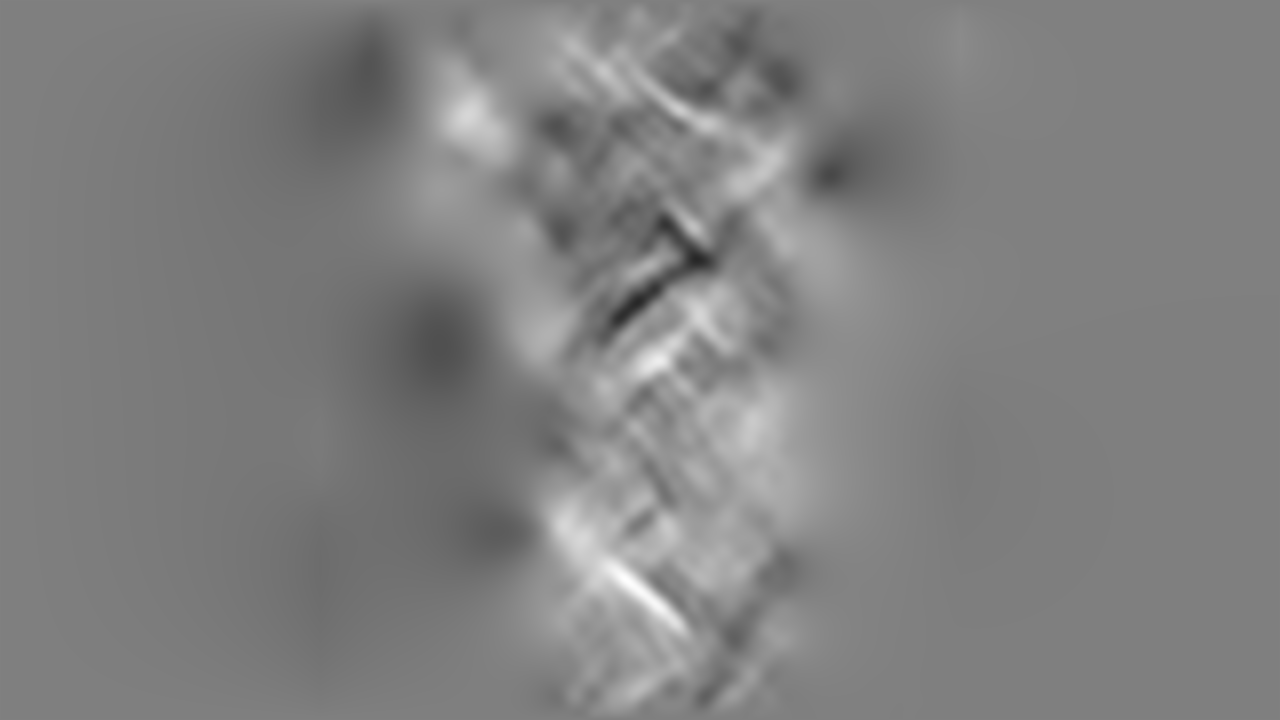}}
  \caption{\centering Poisson images ($\velparams$)}
\end{subfigure}\;
\begin{subfigure}{\figmethodwidth}
  \centering
  {\includegraphics[clip,trim={12cm 3cm 12cm 0cm},width=\linewidth]{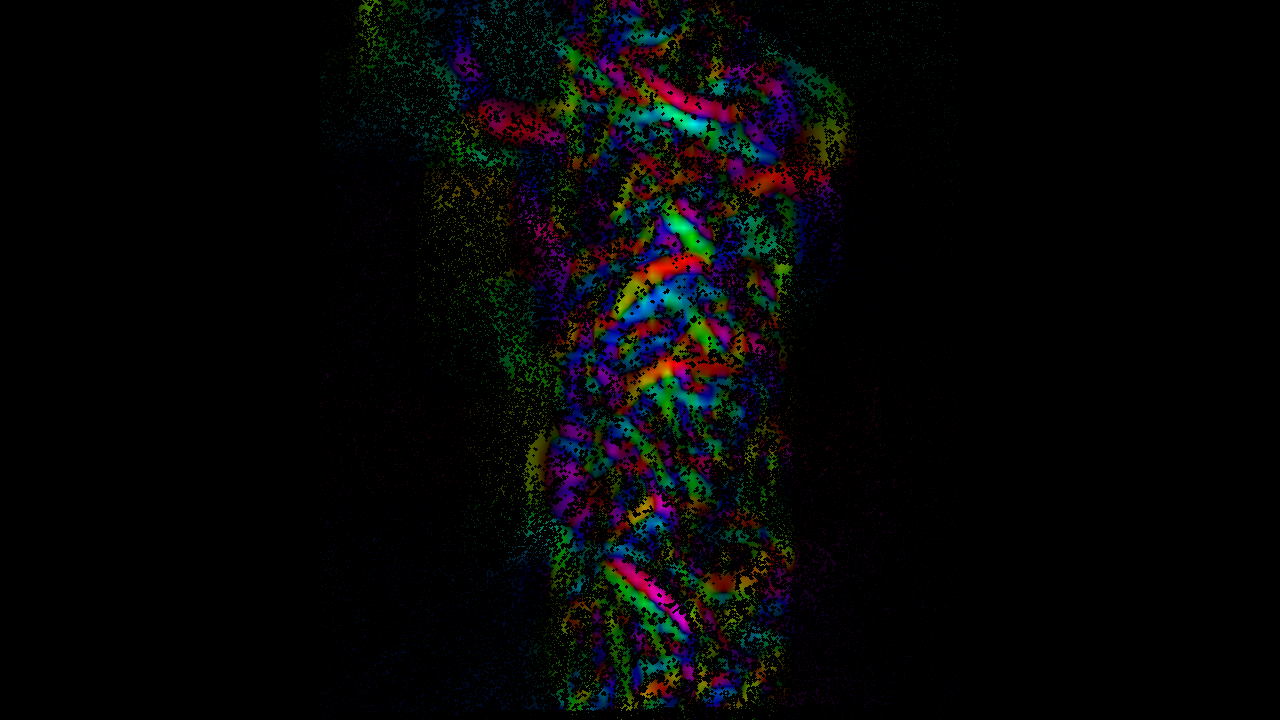}}
  \caption{\centering Optical flow ($\velflow$)}
\end{subfigure}
\caption{Example of Poisson parameters and optical flow.}
\label{fig:suppl:poisson}
\end{figure}

%% file: floats/tab_dataset.tex
\sisetup{round-mode=places,round-precision=1}
\begin{table}[t!]
\centering
\caption{Parameters of the recorded sequences. 
}
\adjustbox{max width=\columnwidth}{%
\setlength{\tabcolsep}{4pt}
\begin{tabular}{l*{6}{S}}
\toprule
Sequence & \text{Convection} & \text{Luminance} & \text{Duration} & \text{Event rate}\\ 
 &  & \text{[lx]} & \text{[s]} & \text{[Mev/s]}\\ 
\midrule
Hot plate 1 & \text{Natural} & \text{4000} & 19.390846 & 11.293378174423127 \\ 
Hot plate 2 & \text{Natural} & \text{225} & 19.774873 & 5.081686390602862 \\ 
Hair dryer (OFF) 1 & \text{Natural} & \text{4000} & 13.499232000000001 & 5.072274407907057 \\ 
Hair dryer (OFF) 2 & \text{Natural} & \text{4000} & 19.719239 & 5.296867794948882 \\ 
Hair dryer (OFF) 3 & \text{Natural}  & \text{225} & 14.683165 & 2.7536987427438158 \\ 
Crushed ice & \text{Natural} & \text{4000} & 17.375194 & 5.04016450118485 \\ 
Hair dryer (ON) & \text{Forced} & \text{4000} & 13.387983 & 14.968827119066404  \\ 
Breathing 1 & \text{Forced} &  \text{4000} & 12.779248 & 4.001311188263973 \\ 
Breathing 2 & \text{Forced} &  \text{4000} & 13.043173 & 3.6801959155184094 \\ 
Helium (synthetic) \cite{settles2022schlieren} & \text{Forced} & \novalue & 0.5 & 161.3 \\ 
\bottomrule
\end{tabular}
\label{tab:dataset}
}
\end{table}

%% file: floats/supplfig_dataset_gt.tex
\global\long\def\figWidth{0.42\linewidth}
\begin{figure}[ht!]
	\centering
    {\scriptsize
    \setlength{\tabcolsep}{2pt}
	\begin{tabular}{
	>{\centering\arraybackslash}m{0.3cm} 
	>{\centering\arraybackslash}m{\figWidth} 
	>{\centering\arraybackslash}m{\figWidth}}
 
        \rotatebox{90}{\makecell{Hot plate}}
		&{\includegraphics[width=\linewidth]{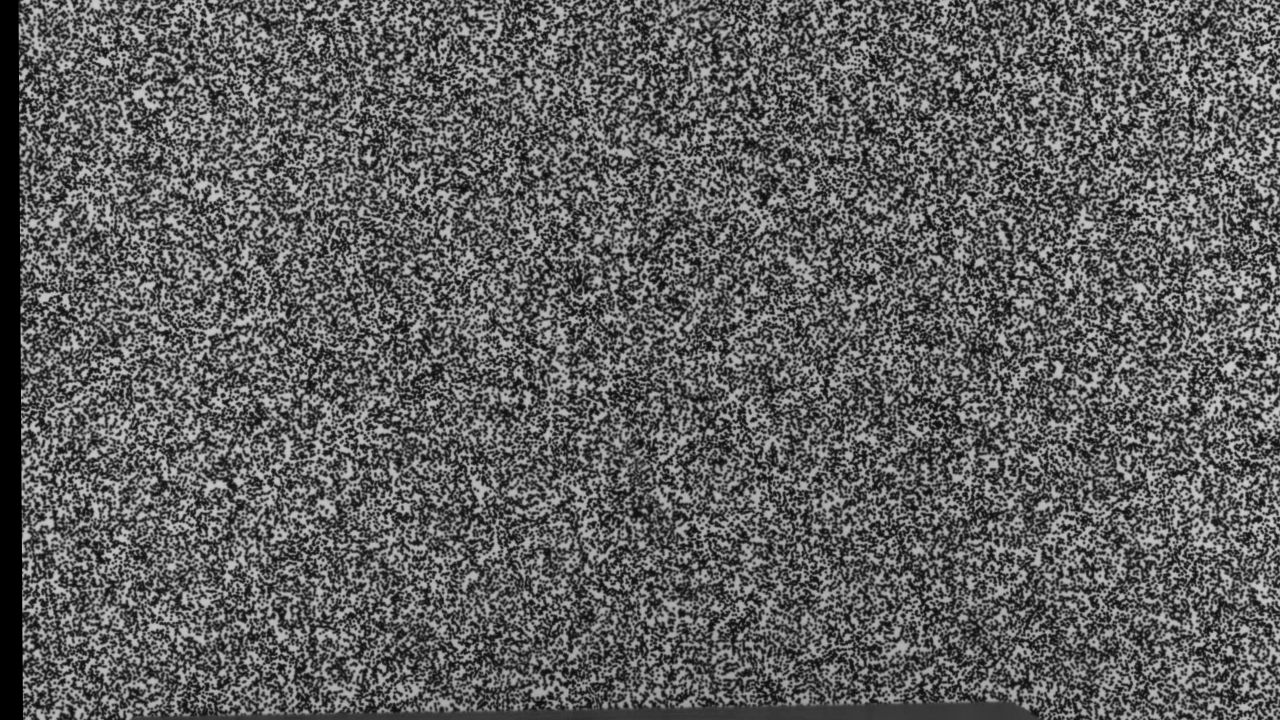}}
		&{\includegraphics[width=\linewidth]{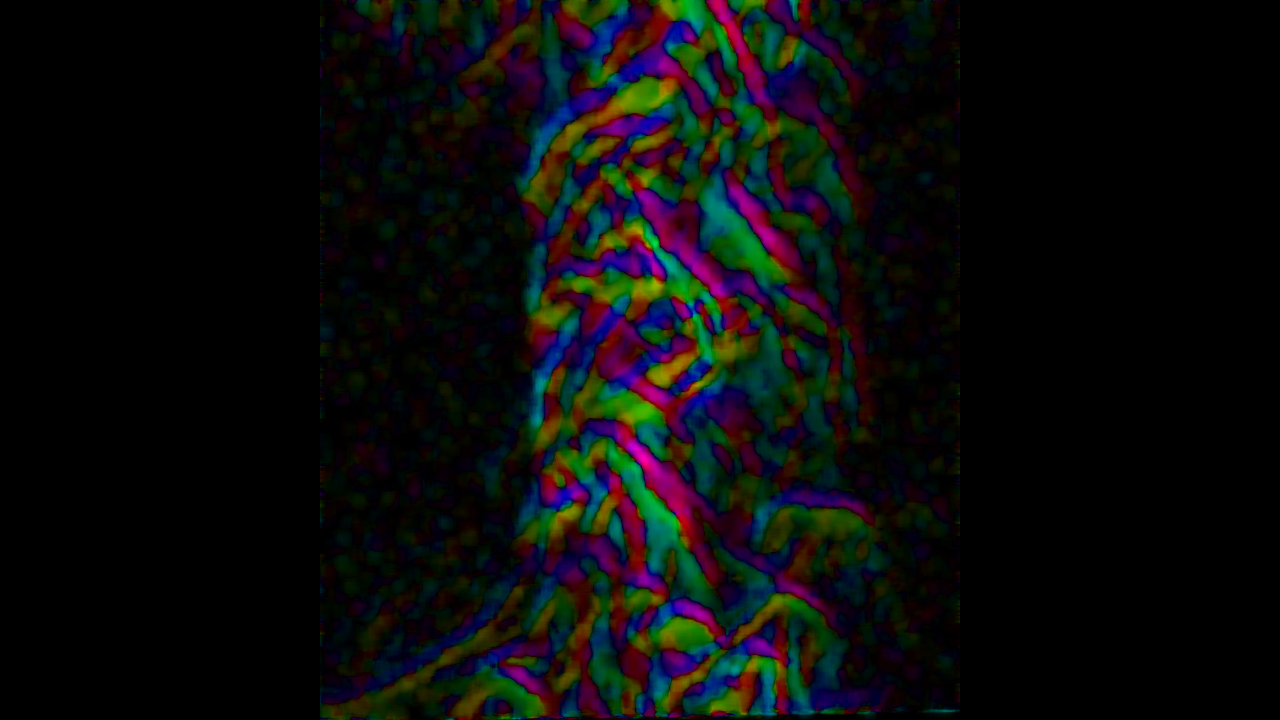}}
        \\
        \rotatebox{90}{\makecell{Hair dryer 1}}
		&{\includegraphics[width=\linewidth]{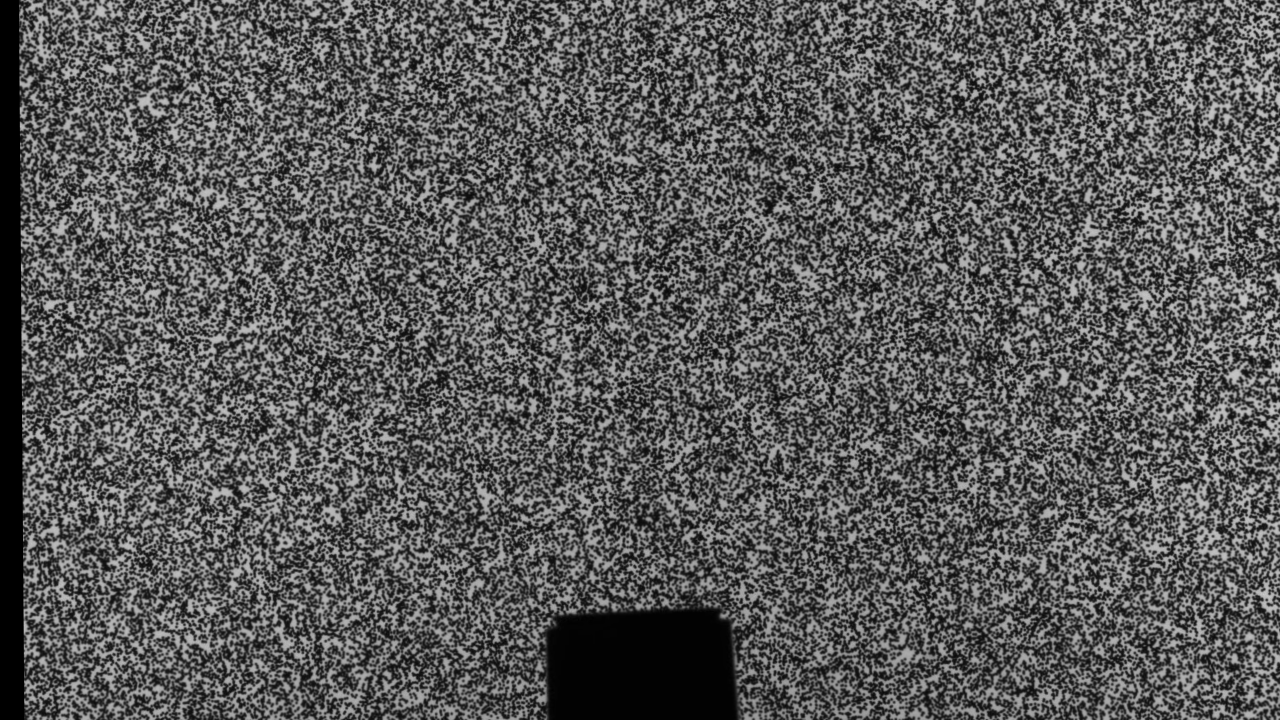}}
		&{\includegraphics[width=\linewidth]{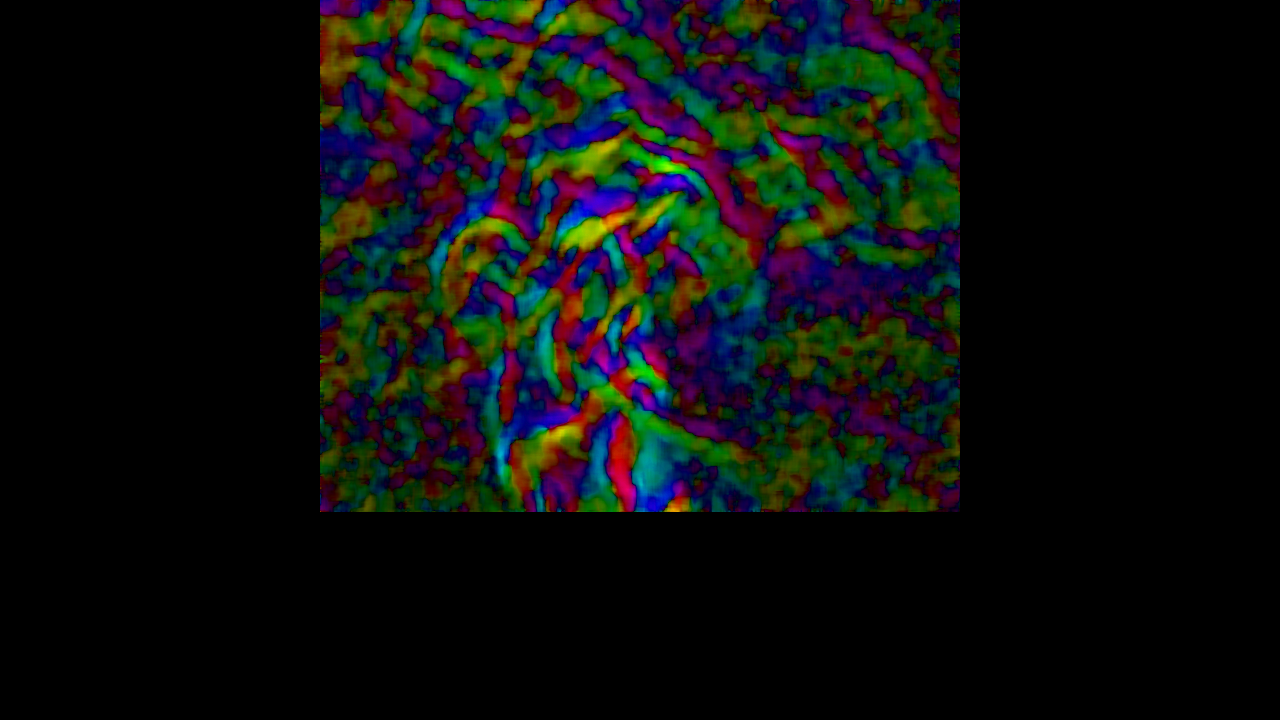}}
        \\

        \rotatebox{90}{\makecell{Hair dryer 2}}
		&{\includegraphics[width=\linewidth]{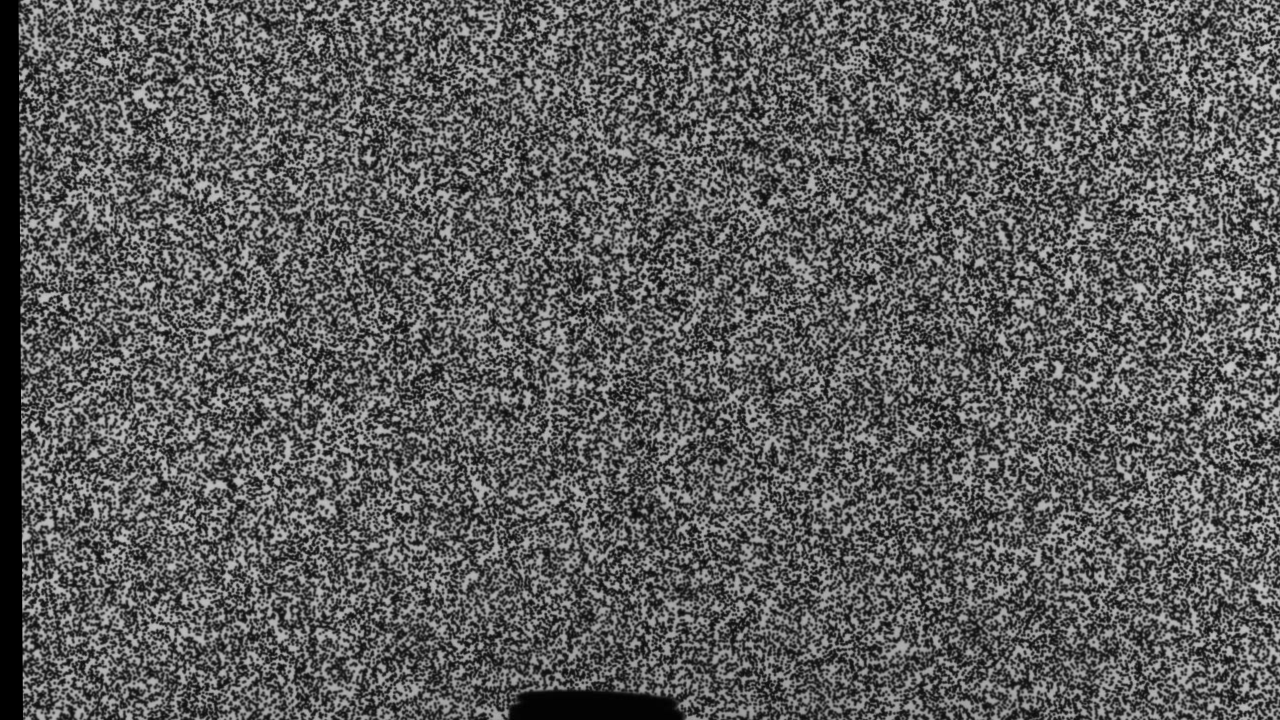}}
		&{\includegraphics[width=\linewidth]{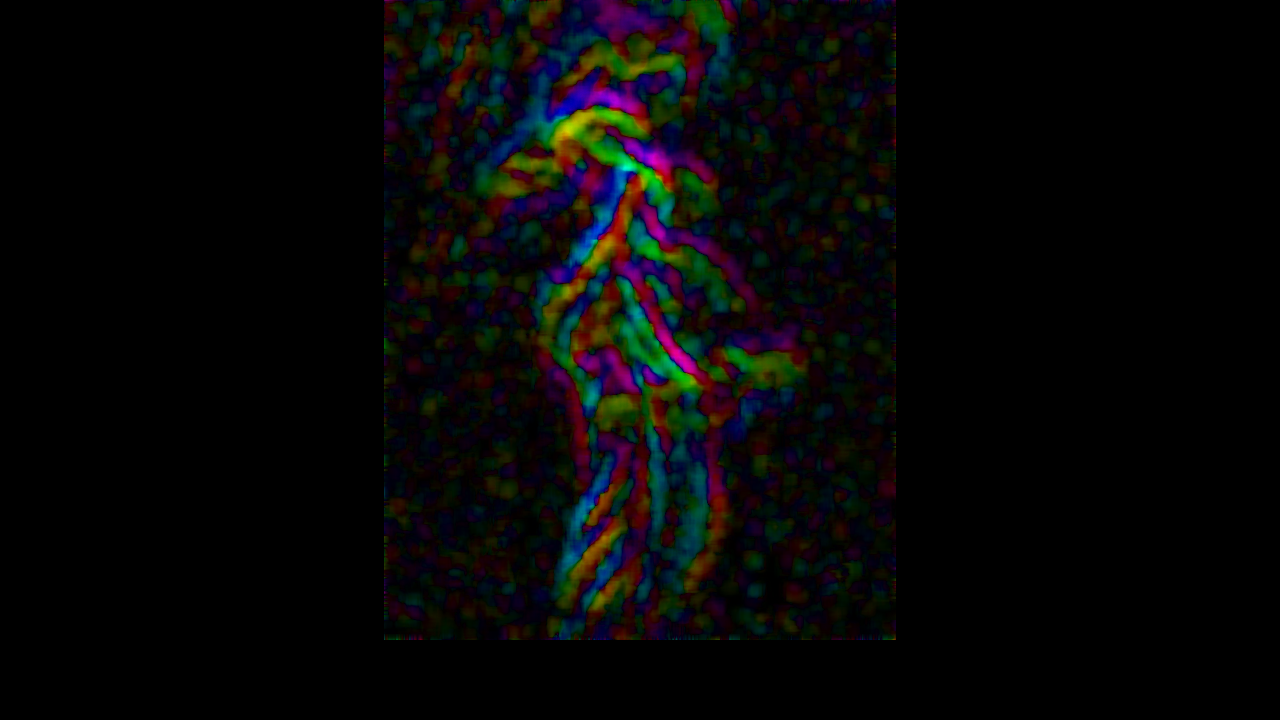}}
        \\

        \rotatebox{90}{\makecell{Breath 2}}
		&{\includegraphics[width=\linewidth]{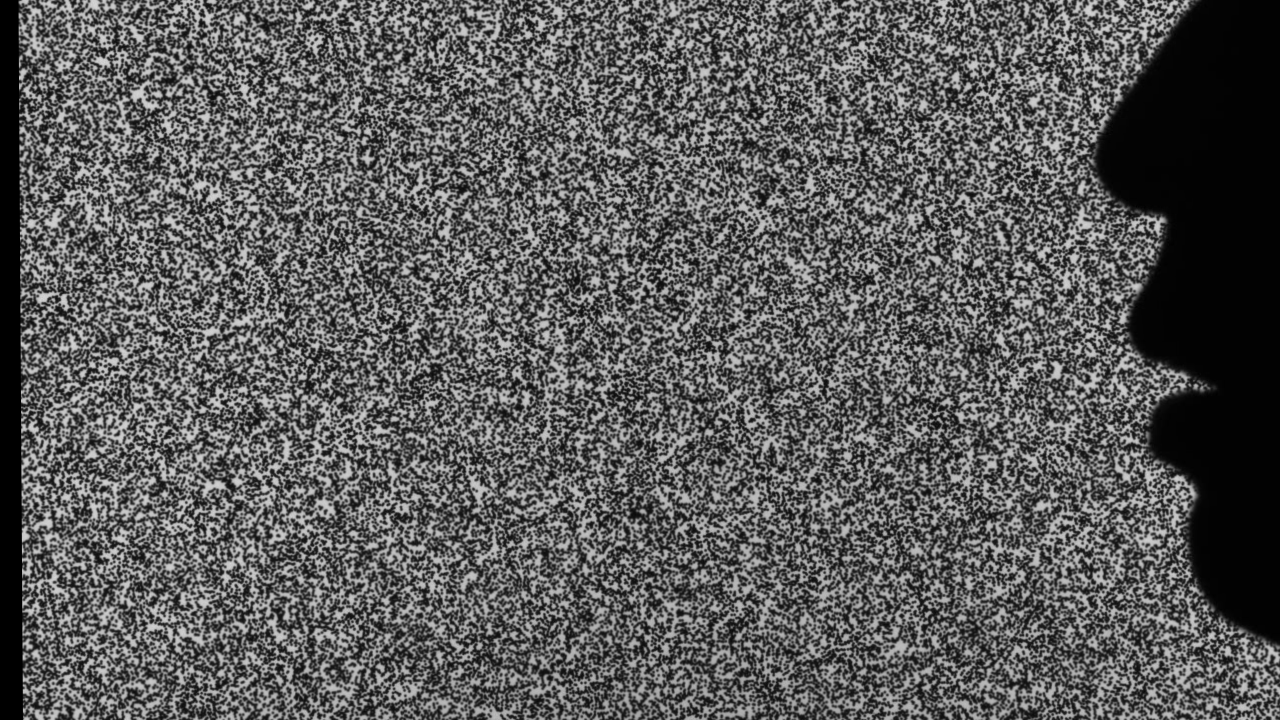}}
		&{\includegraphics[width=\linewidth]{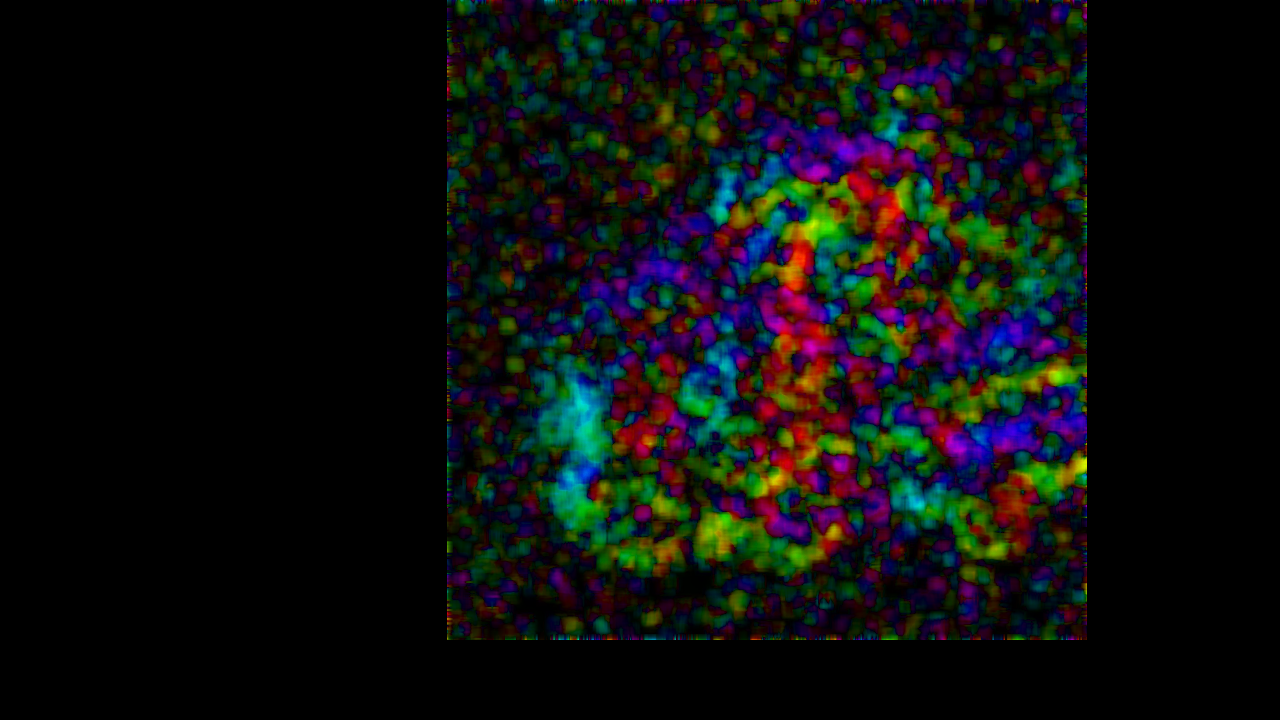}}
        \\

        \rotatebox{90}{\makecell{Hotplate (dark)}}
		&{\includegraphics[width=\linewidth]{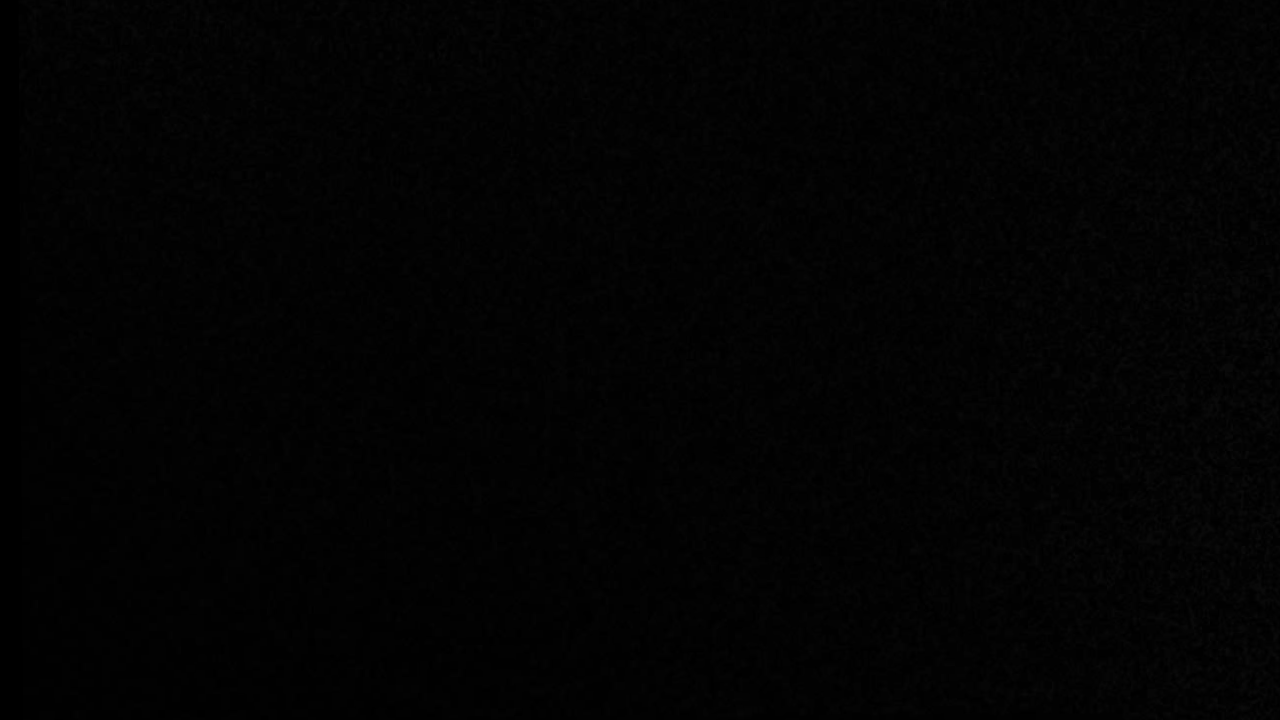}}
		&{\includegraphics[width=\linewidth]{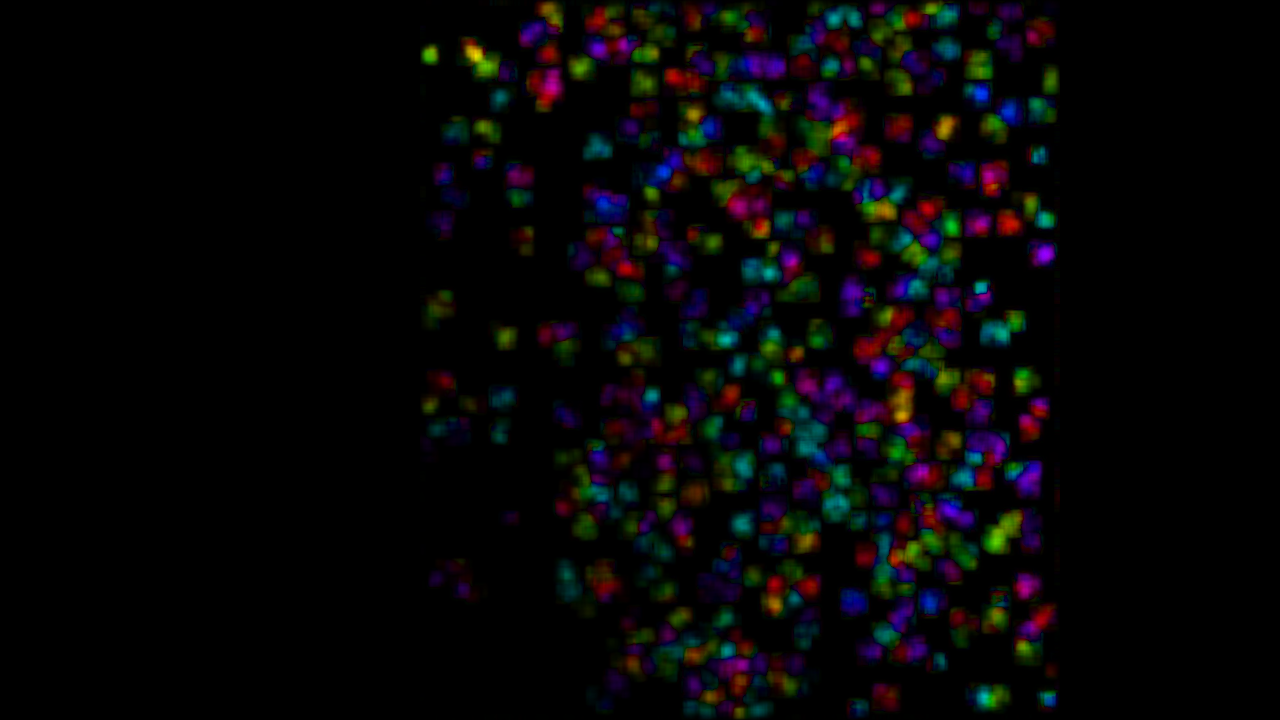}}
        \\

		& \textbf{(a)} Frame data %
		& \textbf{(b)} GT (Flow from frames)
	\end{tabular}
	}
	\caption{\emph{Dataset: samples of frame and frame-based flow.}
	Frames mapped into the event-camera image plane are shown on the left.
	The estimated optical flow (inside the ROI) is shown on the right.
	For the low-light sequences, the frame-based method fails to estimate reasonable flow. 
    Nevertheless, we show them for completeness.}
	\label{fig:suppl:dataset}
\end{figure}

%% file: floats/supplfig_other_flow.tex
\global\long\def\figWidth{0.31\linewidth}
\begin{figure}[t]
	\centering
    {\scriptsize
    \setlength{\tabcolsep}{1pt}
	\begin{tabular}{
	>{\centering\arraybackslash}m{0.3cm} 
	>{\centering\arraybackslash}m{\figWidth} 
	>{\centering\arraybackslash}m{\figWidth} 
	>{\centering\arraybackslash}m{\figWidth}}

        \rotatebox{90}{\makecell{Hot plate}}
        &\includegraphics[clip,trim={11cm 0.5cm 11cm 2cm},width=\linewidth]{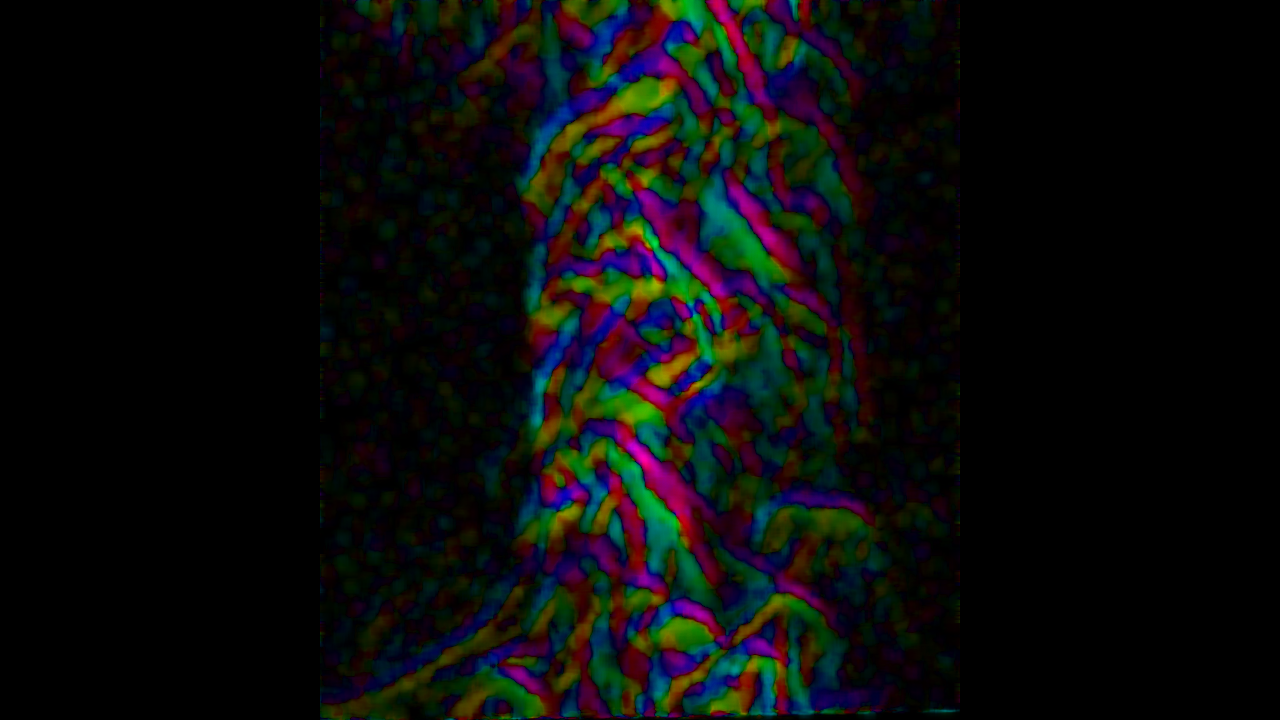}
		&\includegraphics[clip,trim={11cm 0.5cm 11cm 2cm},width=\linewidth]{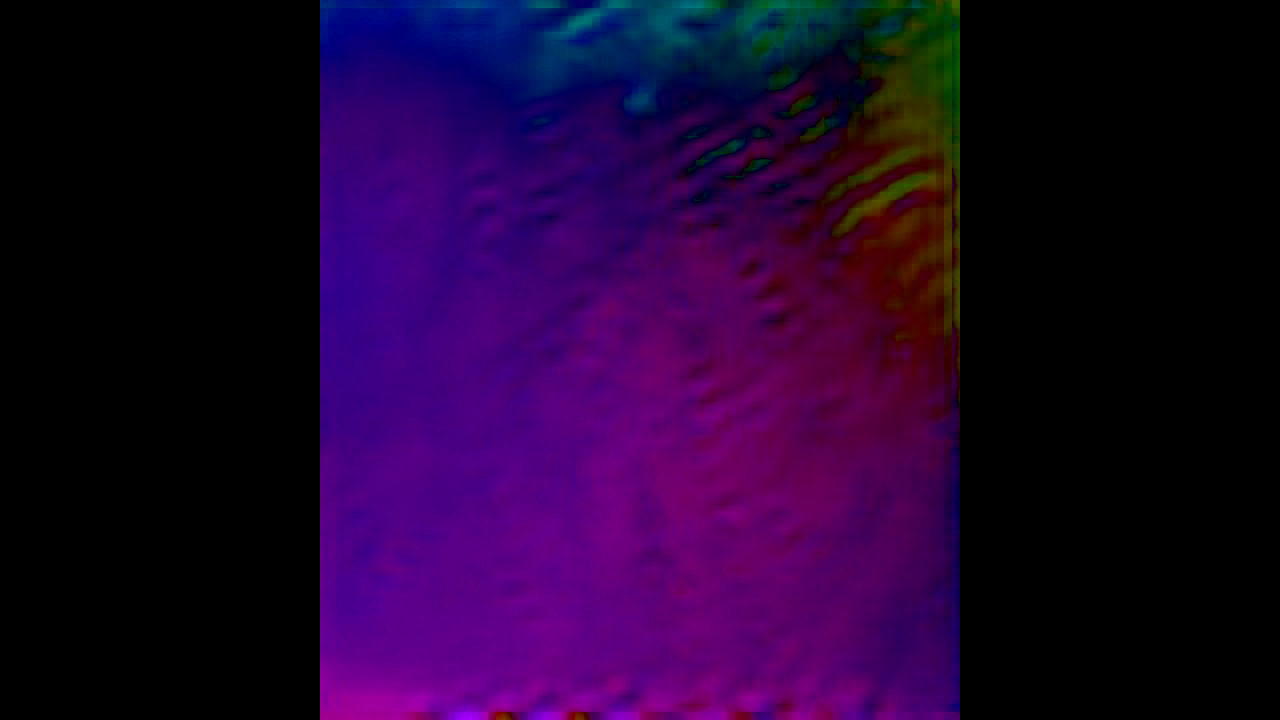}
		&\includegraphics[clip,trim={11cm 0.5cm 11cm 2cm},width=\linewidth]{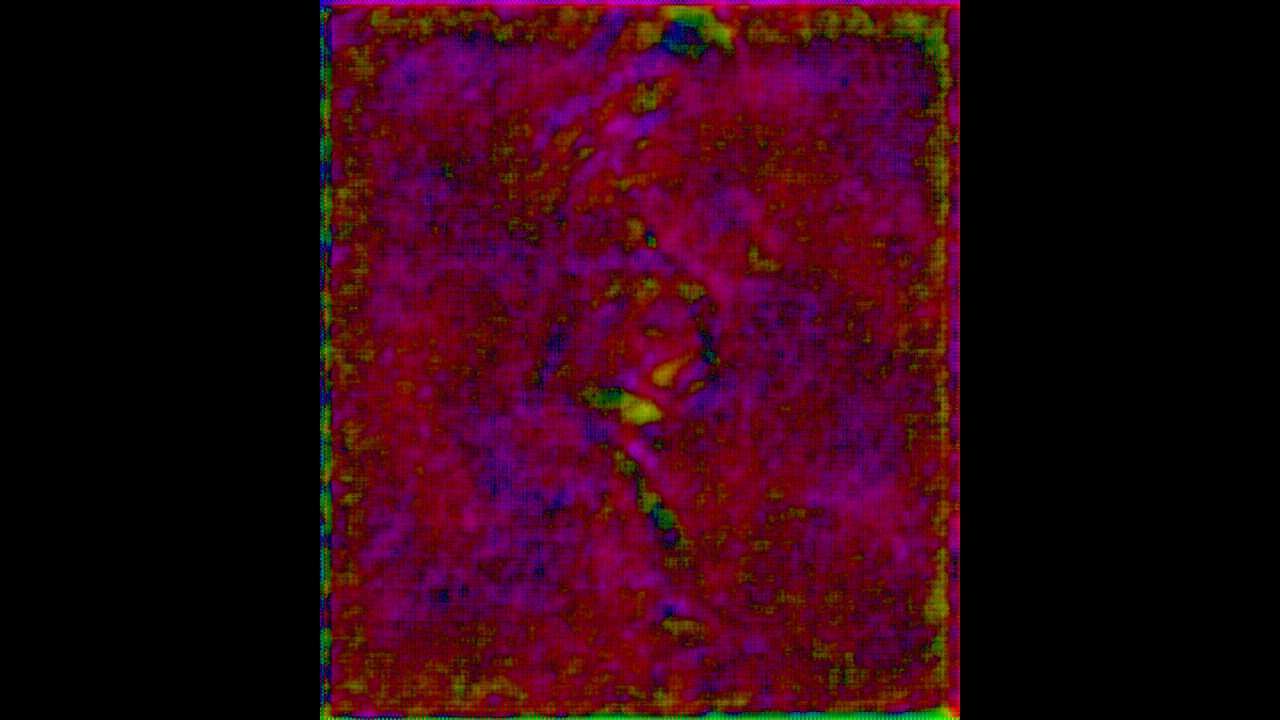}
		\\

        \rotatebox{90}{\makecell{Hair Dryer}}
        &\includegraphics[clip,trim={13.5cm 5cm 13.5cm 3cm},width=\linewidth]{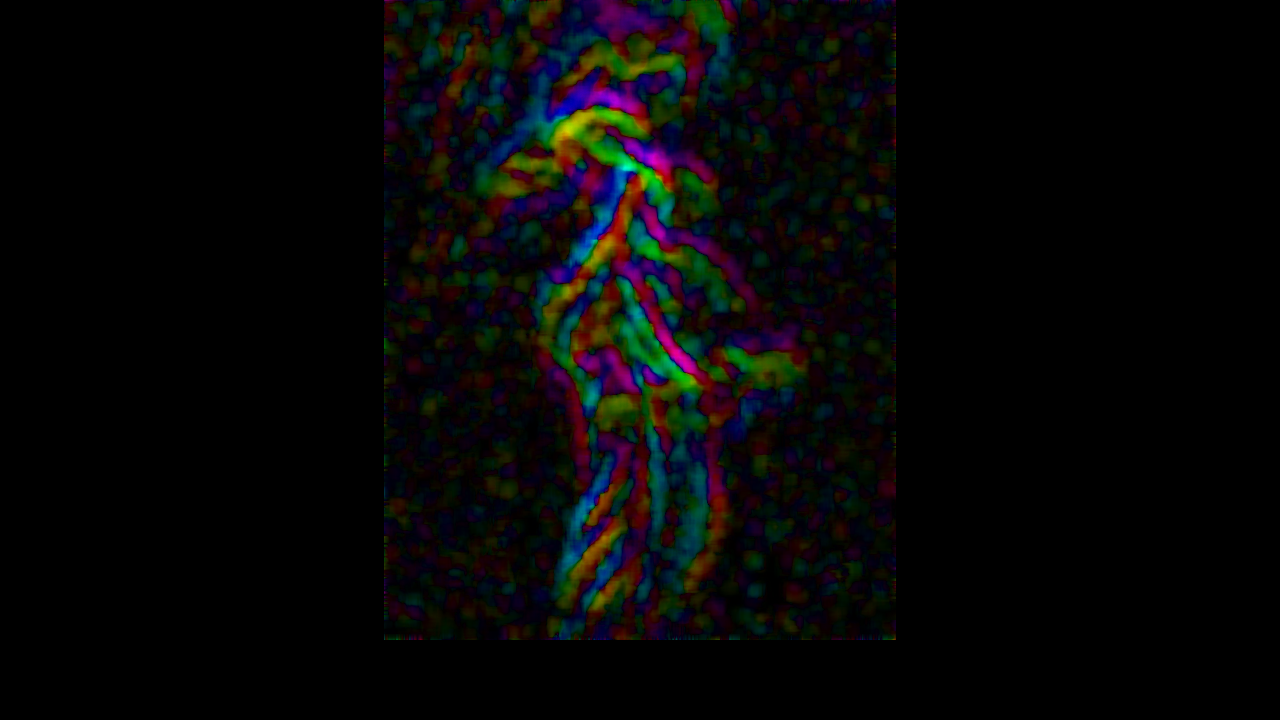}
        &\includegraphics[clip,trim={13.5cm 5cm 13.5cm 3cm},width=\linewidth]{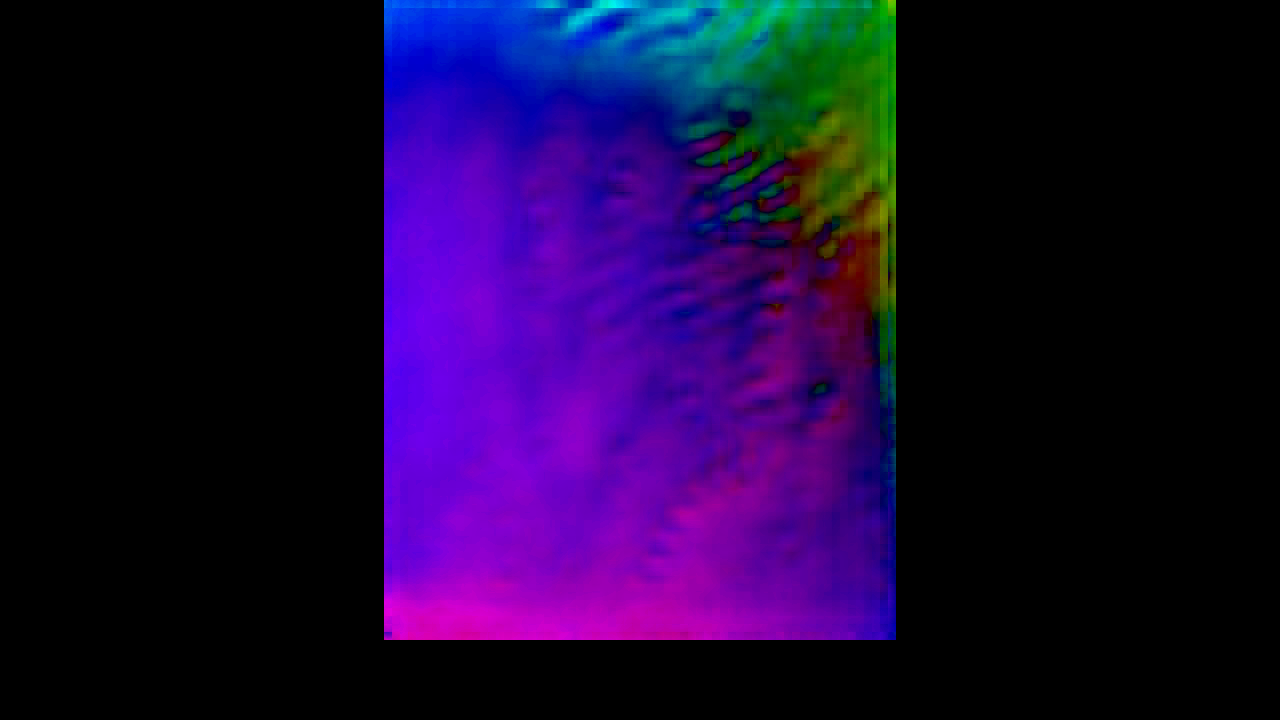}
        &\includegraphics[clip,trim={13.5cm 5cm 13.5cm 3cm},width=\linewidth]{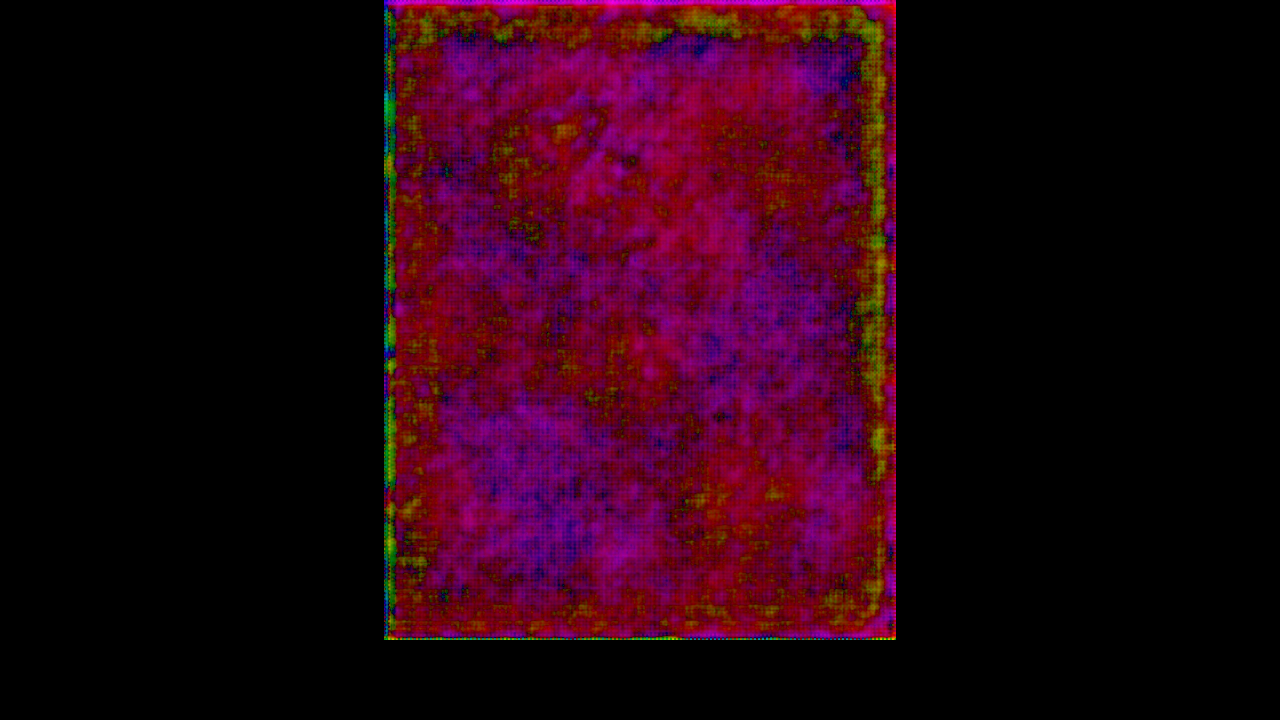}
        \\

        \rotatebox{90}{\makecell{Breath}}
        &\includegraphics[clip,trim={15.7cm 3cm 6.7cm 0cm},width=\linewidth]{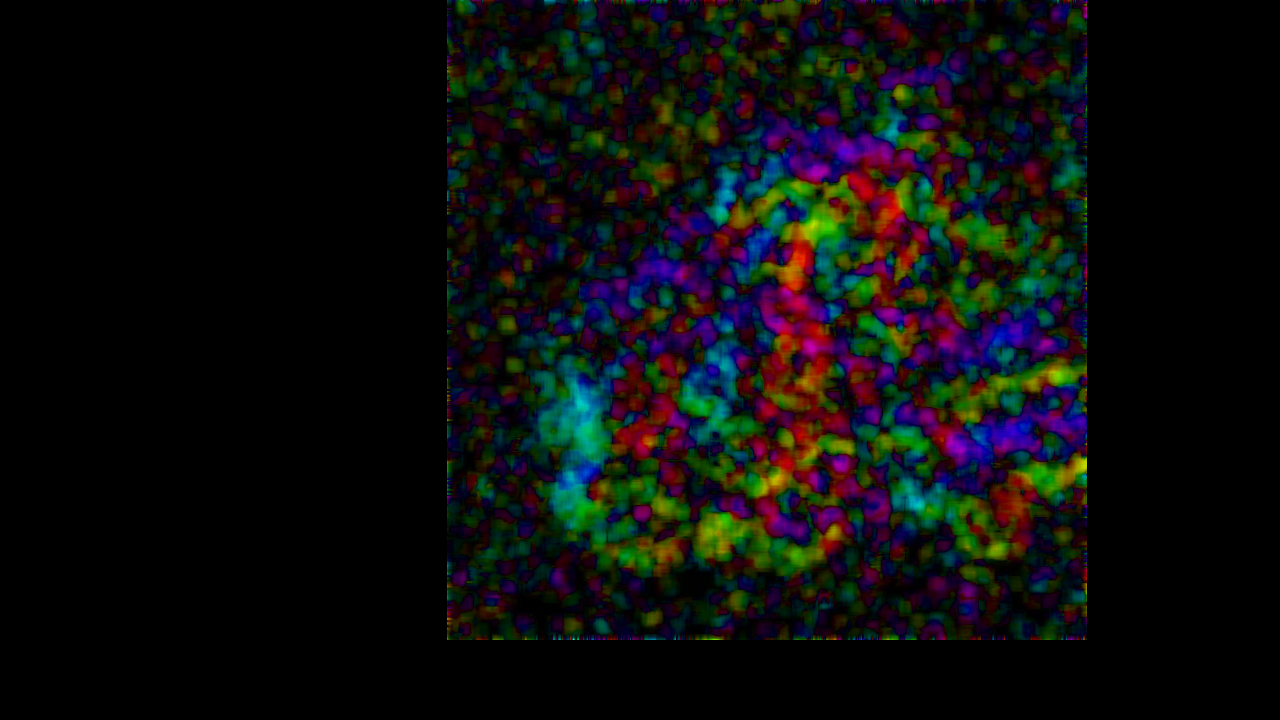}
		&\includegraphics[clip,trim={15.7cm 3cm 6.7cm 0cm},width=\linewidth]{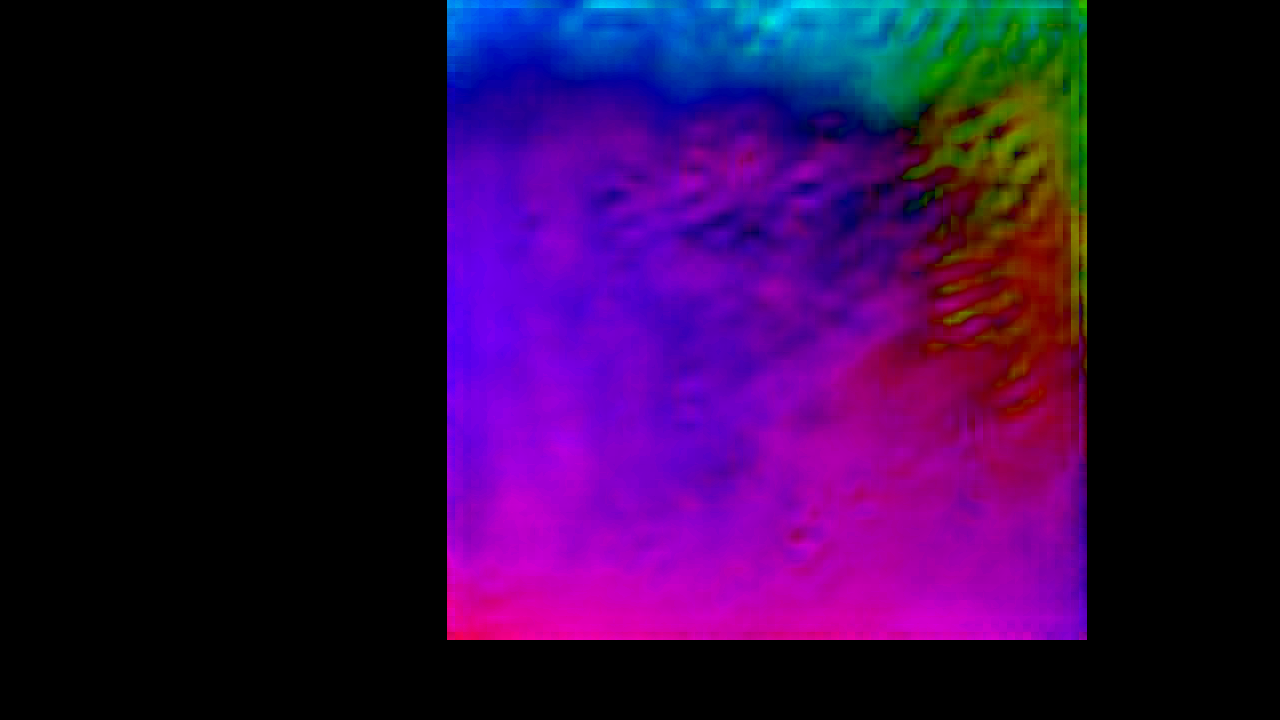}
		&\includegraphics[clip,trim={15.7cm 3cm 6.7cm 0cm},width=\linewidth]{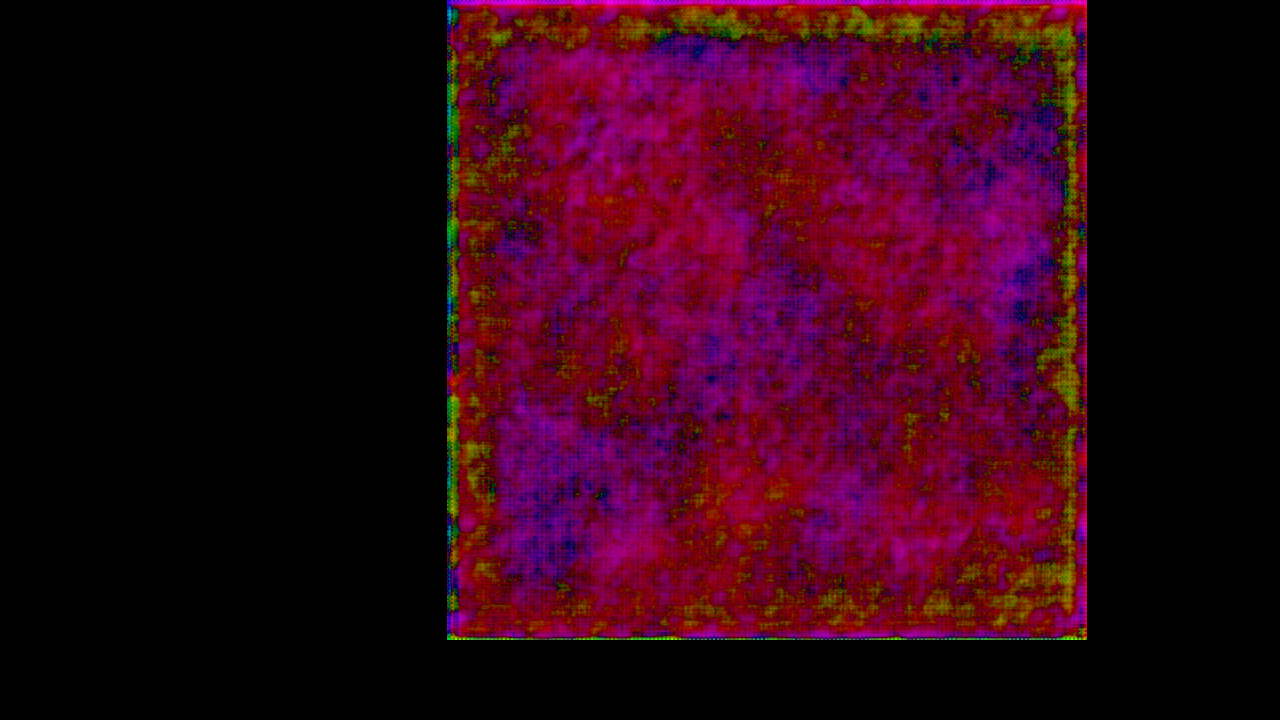}
		\\

		& \textbf{(a)} Farneb\"ack's \cite{Farnebaeck03scia}
		& \textbf{(b)} FlowFormer \cite{Huang2022eccv1}
		& \textbf{(c)} RIFE \cite{Huang2022eccv2}
	\end{tabular}
	}
	\caption{Different frame-based optical flow methods.
	}
	\label{fig:suppl:otherFlow}
\end{figure}

%% file: floats/suppltab_benchmark_parameters.tex
\sisetup{round-mode=places,round-precision=0}
\begin{table}[t]
\centering
\caption{Details of the benchmark. ``ROI position'' contains the coordinates of the top-left corner.
}
\adjustbox{max width=\columnwidth}{%
\setlength{\tabcolsep}{4pt}
\begin{tabular}{l*{5}{S}}
\toprule
Sequence & \text{ROI size [px]} & \text{ROI position [px]}& \text{Duration [s]} & \text{Total events} \\
\midrule
Hot plate 1 &  \text{640$\times$720} & \text{[320, 0]} &  \text{10 to 14}  & \num{51900802} \\ 
Hot plate 2 (dark) & \text{640$\times$720} & \text{[420, 0]} & \text{12 to 14} & \num{12912262} \\ 
Hair dryer (OFF) 1  &  \text{640$\times$640} & \text{[320, 0]} & \text{4 to 7} & \num{13498252}\\ 
Hair dryer (OFF) 2  & \text{512$\times$640} & \text{[384, 0]} &\text{6 to 7} & \num{4089883} \\ 
Hair dryer (dark) & \text{512$\times$640} & \text{[384, 0]}  & \text{5 to 7} & \num{3460579} \\ 
Crushed ice & \text{512$\times$512}  & \text{[384, 208]} & \text{8 to 11} & \num{5856190} \\ 
Hair dryer (ON)  & \text{1280$\times$200} & \text{[0, 260]} & \text{3.3 to 4.3} & \num{17860129} \\ 
Breathing 1  & \text{590$\times$600}  & \text{[400, 0]} & \text{4.36 to 5.5} & \num{2783122} \\ 
Breathing 2  & \text{640$\times$640}  & \text{[447, 0]} & \text{2.5 to 3.5} & \num{1811889} \\ 
\midrule
Total  & \novalue  & \novalue & \text{18.14} & \num{114173108} \\ 
\bottomrule
\end{tabular}
\label{tab:suppl:benchmark}
}
\end{table}

%% file: floats/tab_full_result.tex
\sisetup{round-mode=places,round-precision=3}
\begin{table*}[!ht]
\centering
\caption{
Results of optical flow estimation.
}
\adjustbox{max width=\textwidth}{%
\setlength{\tabcolsep}{2pt}
\begin{tabular}{l*{13}{S}}

\toprule
 & \multicolumn{3}{c}{Hair dryer (OFF) 1} %
 & \multicolumn{3}{c}{Hair dryer (OFF) 2} %
 & \multicolumn{3}{c}{Hot plate 1}  %
 & \multicolumn{3}{c}{Hair dryer (ON)}  %
 \\
 \cmidrule(l{1mm}r{1mm}){2-4}
 \cmidrule(l{1mm}r{1mm}){5-7}
 \cmidrule(l{1mm}r{1mm}){8-10}
 \cmidrule(l{1mm}r{1mm}){11-13}

&\text{AEE $\downarrow$} & \text{\%Out $\downarrow$} & \text{AE $\downarrow$}
&\text{AEE $\downarrow$} & \text{\%Out $\downarrow$} & \text{AE $\downarrow$}
&\text{AEE $\downarrow$} & \text{\%Out $\downarrow$} & \text{AE $\downarrow$}
&\text{AEE $\downarrow$} & \text{\%Out $\downarrow$} & \text{AE $\downarrow$}
\\

\midrule
MCM~\cite{Shiba22eccv} 
& 1.42500 & 35.63915 & 0.62126
& 0.4213942126 & 10.88615526 & 0.47629
& \bnum{0.39982915} & 21.788710999951608 & 0.42622
& 0.28747 & 5.93331 & 0.71178
\\
E2VID~\cite{Rebecq19pami}
& 1.05468070337411 & 39.06761 & 0.67731
&1.09089  & 37.73353 & 0.67048
& 1.09172 & 32.12134 & 0.61125
& 0.81097 & 25.99747 & 0.58748
\\ 
Ours (Flow)
& 0.674770877541533 & 22.10363 & 0.40357
& 0.68819 & 24.93027 & 0.44783
& 0.80993 & 30.28905 & 0.54433
& 0.30957 & 6.75617 & 0.25833
\\ 
Ours (Poisson)
& \bnum{0.383128899671618} & \bnum{9.31932} & \bnum{0.29945}
& \bnum{0.39536} & \bnum{10.17393} & \bnum{0.33696}
& 0.48688 & \bnum{12.21458} & \bnum{0.42067}
& \bnum{0.21521} & \bnum{0.92424} & \bnum{0.20159}
\\

\midrule\\[-1.5ex]
 & \multicolumn{3}{c}{Crushed ice} %
 & \multicolumn{3}{c}{Breathing 1} %
 & \multicolumn{3}{c}{Breathing 2} %
 \\
\midrule
MCM~\cite{Shiba22eccv} 
& 1.08983 & 96.96414 & 0.82328
& 1.7694736939213669 & 49.5523077933639 & 0.8531212079242585
& 2.05586 & 78.69012 & 0.97342
\\
E2VID~\cite{Rebecq19pami}
& 1.24944 & 55.02976 & 0.79073
& 1.01365 & 42.07163 & 0.69237
& 1.05568 & 43.34811 & 0.69908
\\ 
Ours (Flow)
& 0.58675 & 21.81487 & 0.45184
& 0.66475 & 11.87193 & 0.34102
& 0.55692 & 17.71566 & 0.43814
\\ 
Ours (Poisson)
& \bnum{0.32560} & \bnum{5.17659} & \bnum{0.30111}
& \bnum{0.34507} & \bnum{6.32194} & \bnum{0.20333}
& \bnum{0.47623} & \bnum{8.02823} & \bnum{0.41018}
\\ 

\bottomrule
\end{tabular}
\label{tab:fullResult}
}
\end{table*}

%% file: floats/fig_result_comparison.tex
\global\long\def\figWidth{0.185\linewidth}
\begin{figure*}[ht!]
	\centering
    {\scriptsize
    \setlength{\tabcolsep}{2pt}
	\begin{tabular}{
	>{\centering\arraybackslash}m{0.3cm} 
	>{\centering\arraybackslash}m{\figWidth} 
	>{\centering\arraybackslash}m{\figWidth} 
	>{\centering\arraybackslash}m{\figWidth} 
	>{\centering\arraybackslash}m{\figWidth} 
	>{\centering\arraybackslash}m{\figWidth}}

        \rotatebox{90}{\makecell{Hot plate}}
		&\gframe{\includegraphics[clip,trim={12cm 0cm 12cm 0cm},width=\linewidth]{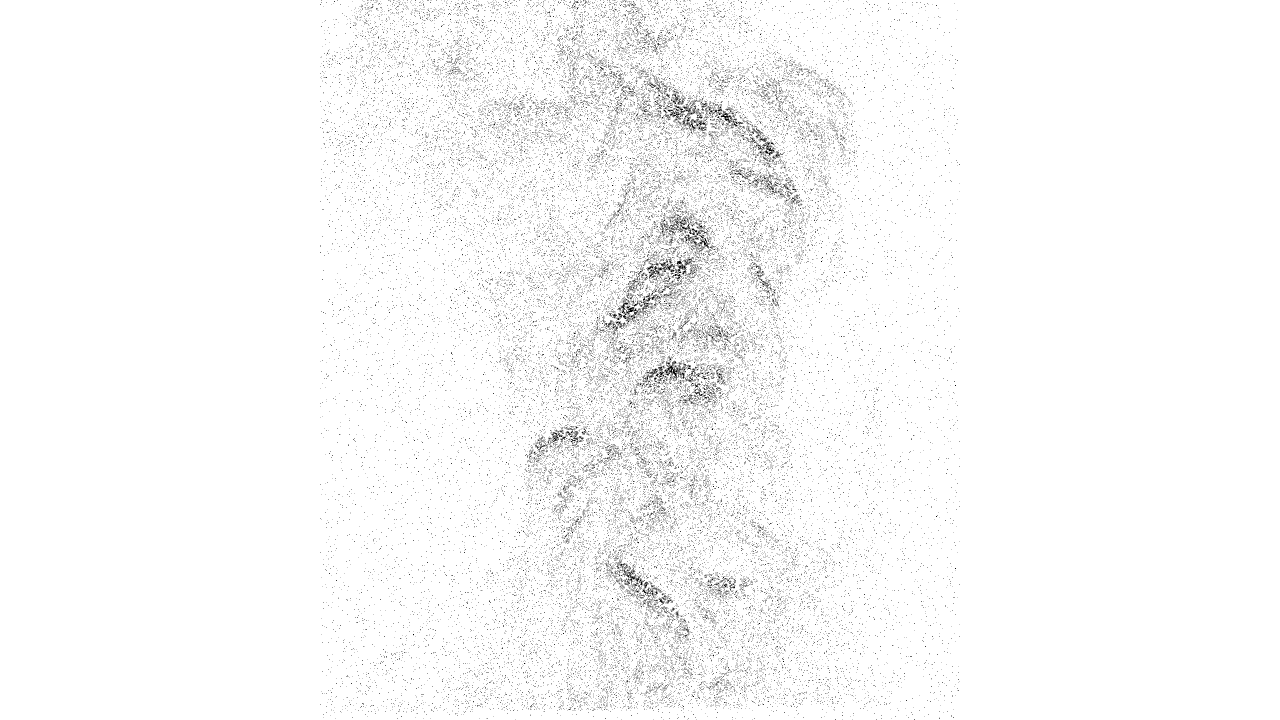}}
		&\includegraphics[clip,trim={12cm 0cm 12cm 0cm},width=\linewidth]{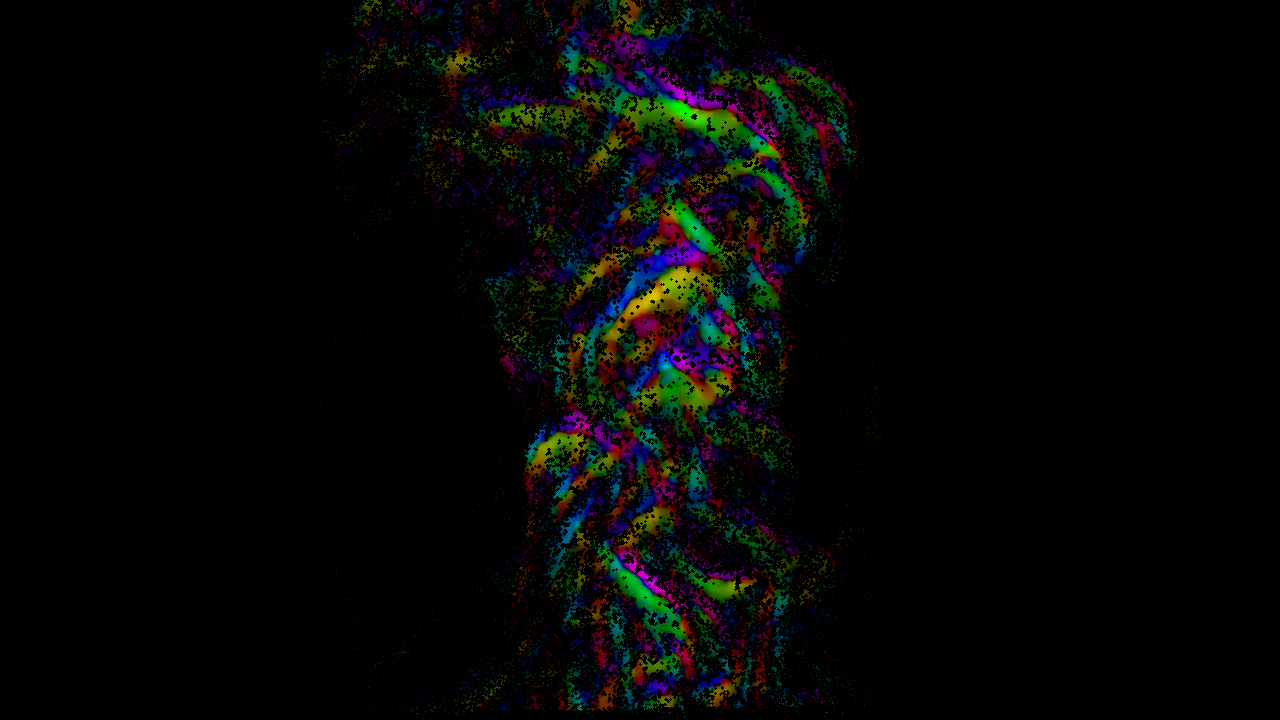}
		&\includegraphics[clip,trim={12cm 0cm 12cm 0cm},width=\linewidth]{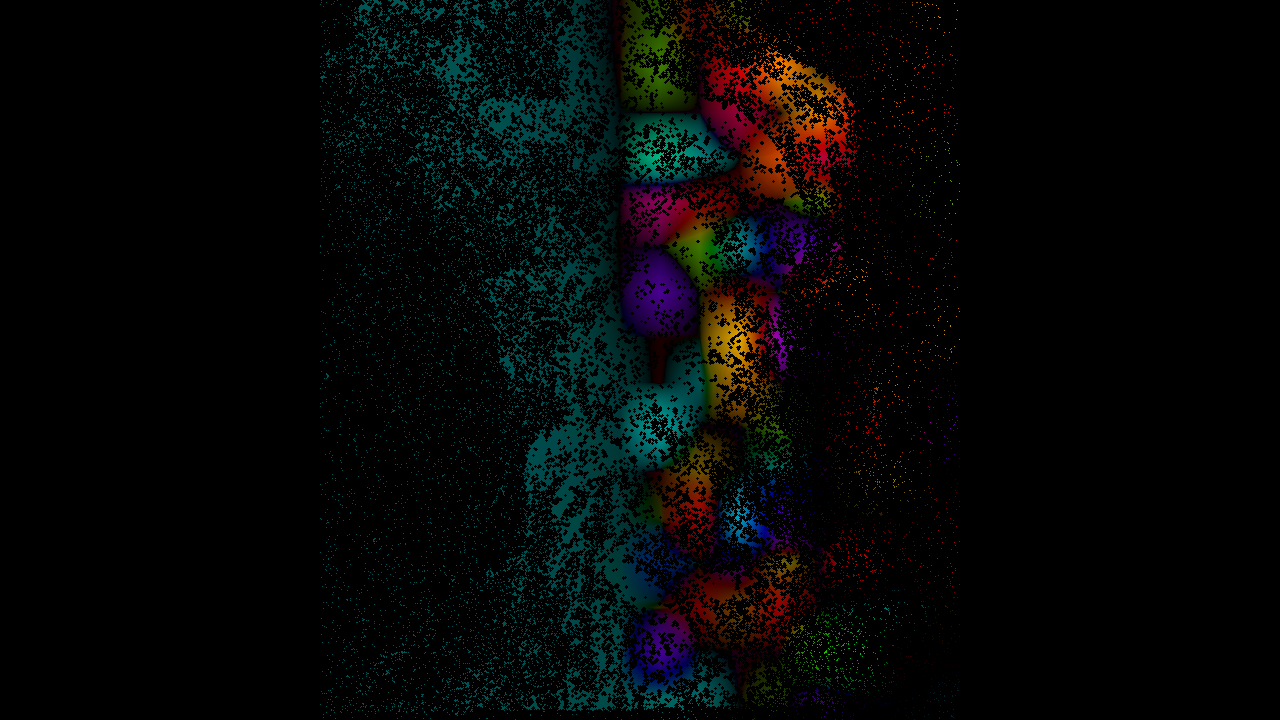}
		&\includegraphics[clip,trim={12cm 0cm 12cm 0cm},width=\linewidth]{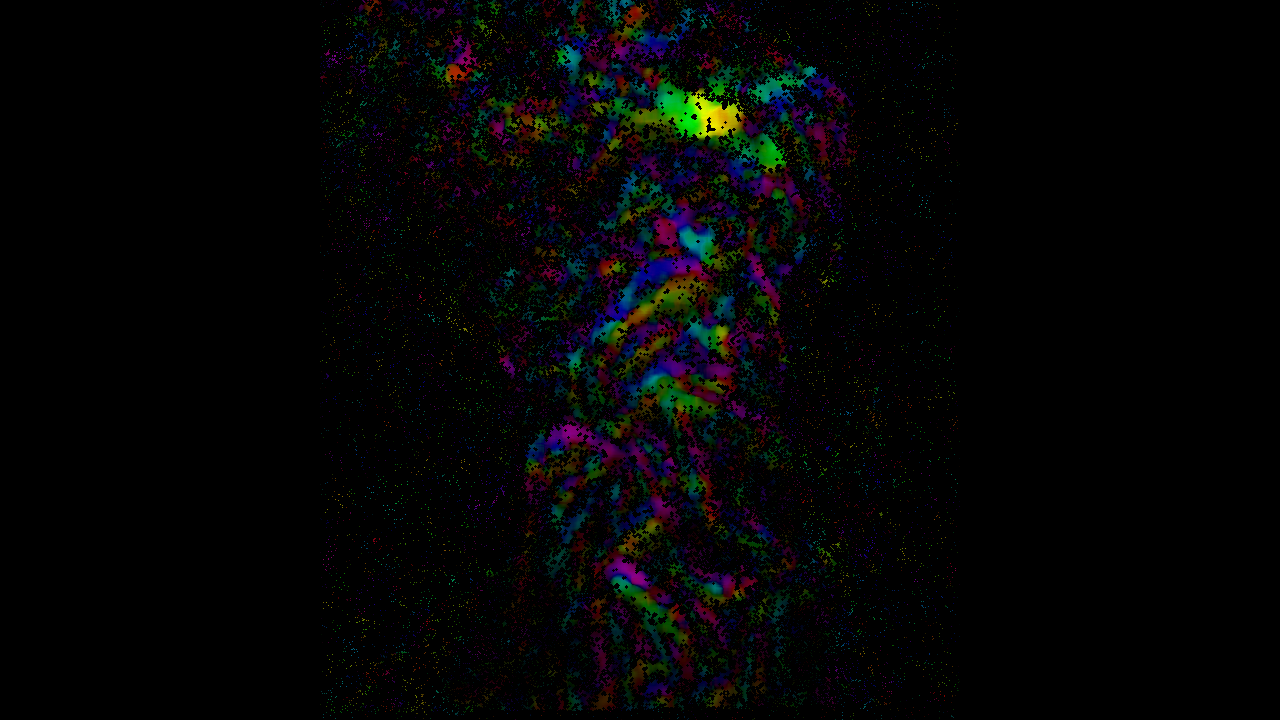}
		&\includegraphics[clip,trim={12cm 0cm 12cm 0cm},width=\linewidth]{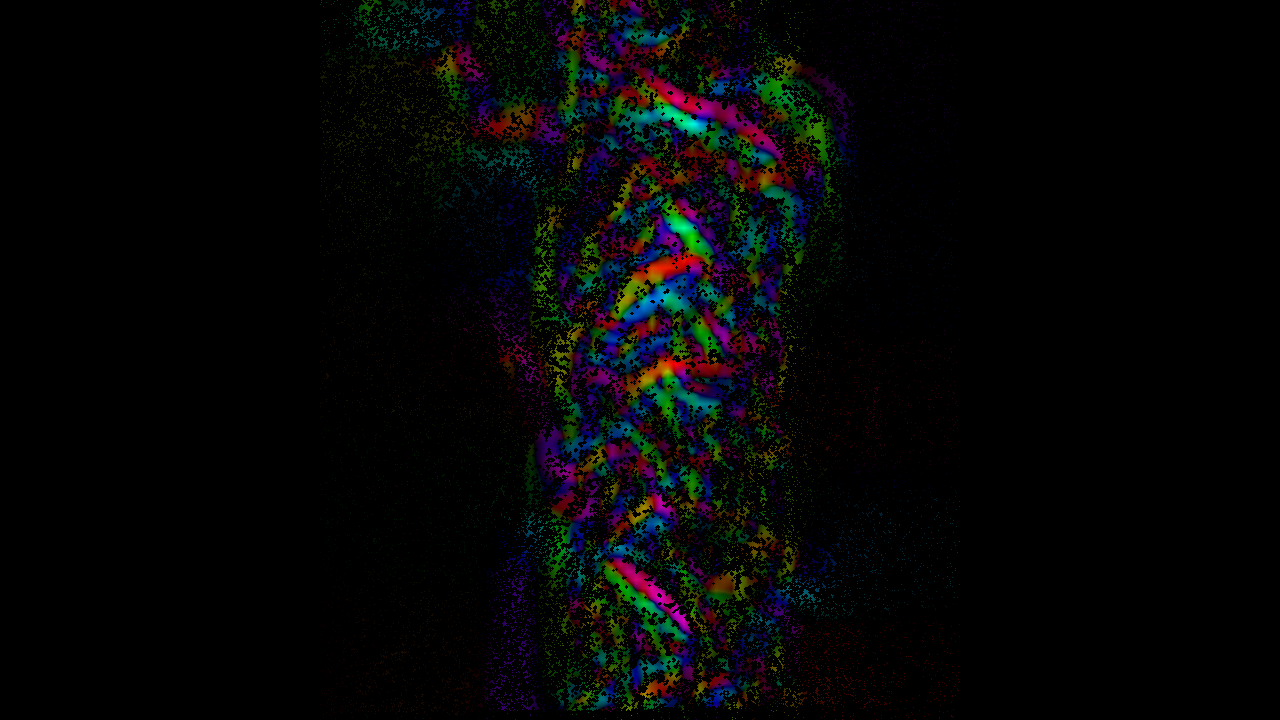}
        \\

        \rotatebox{90}{\makecell{Hot plate}}
		&\gframe{\includegraphics[clip,trim={12cm 0cm 12cm 0cm},width=\linewidth]{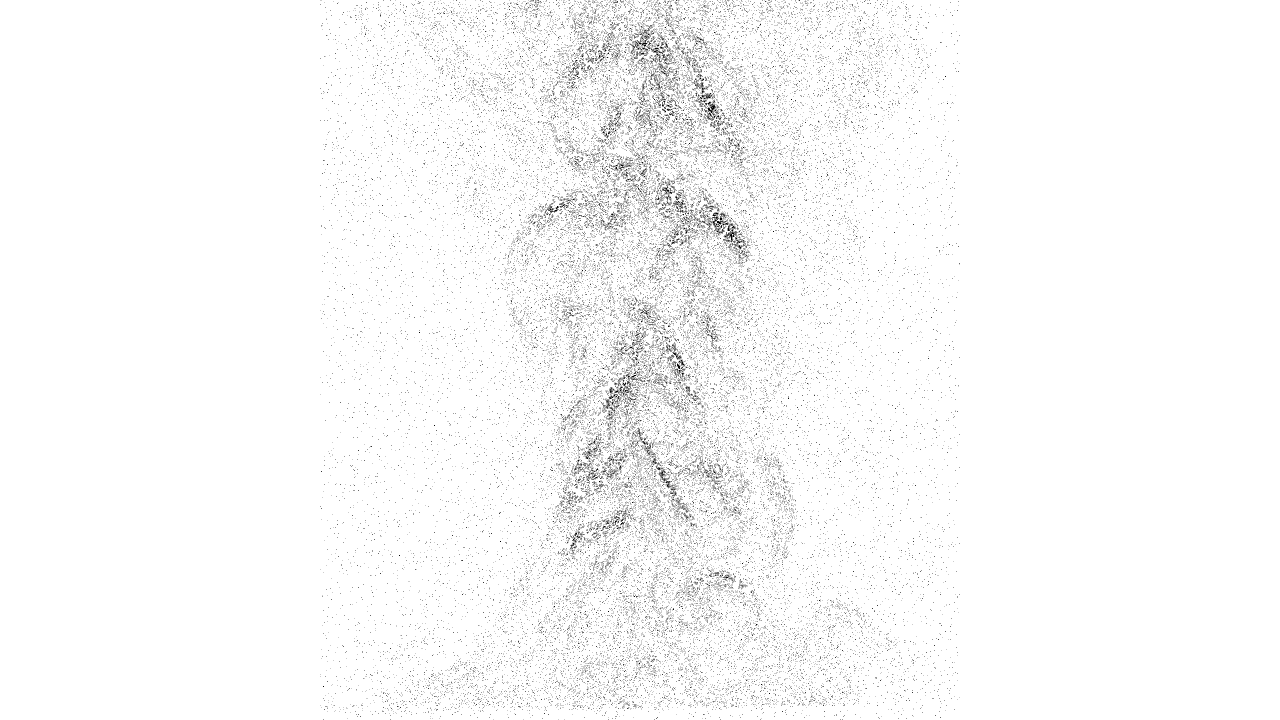}}
		&\includegraphics[clip,trim={12cm 0cm 12cm 0cm},width=\linewidth]{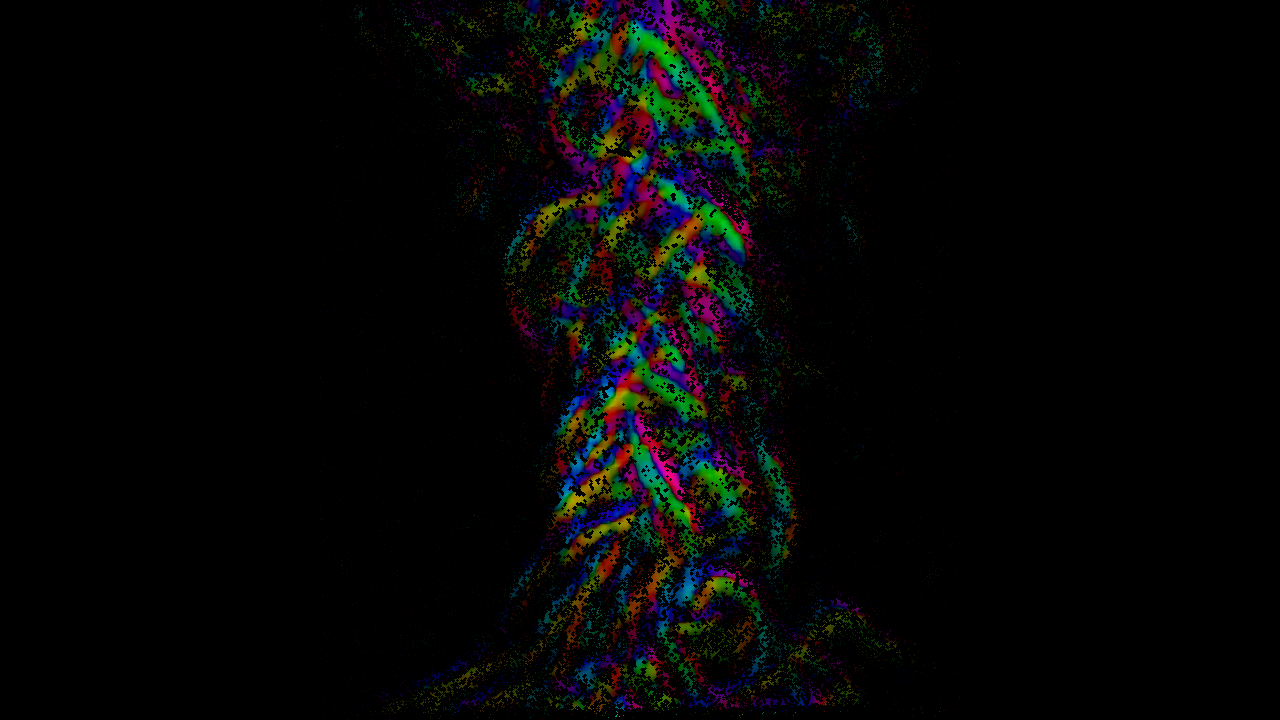}
		&\includegraphics[clip,trim={12cm 0cm 12cm 0cm},width=\linewidth]{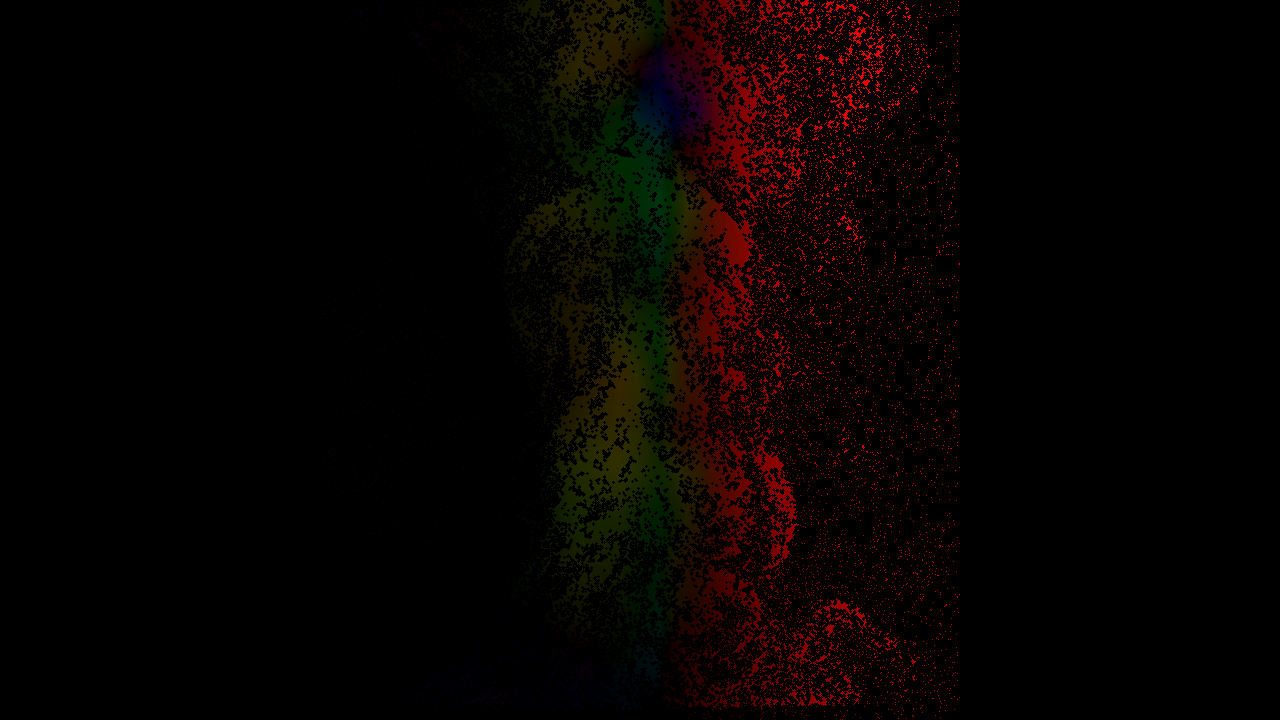}
		&\includegraphics[clip,trim={12cm 0cm 12cm 0cm},width=\linewidth]{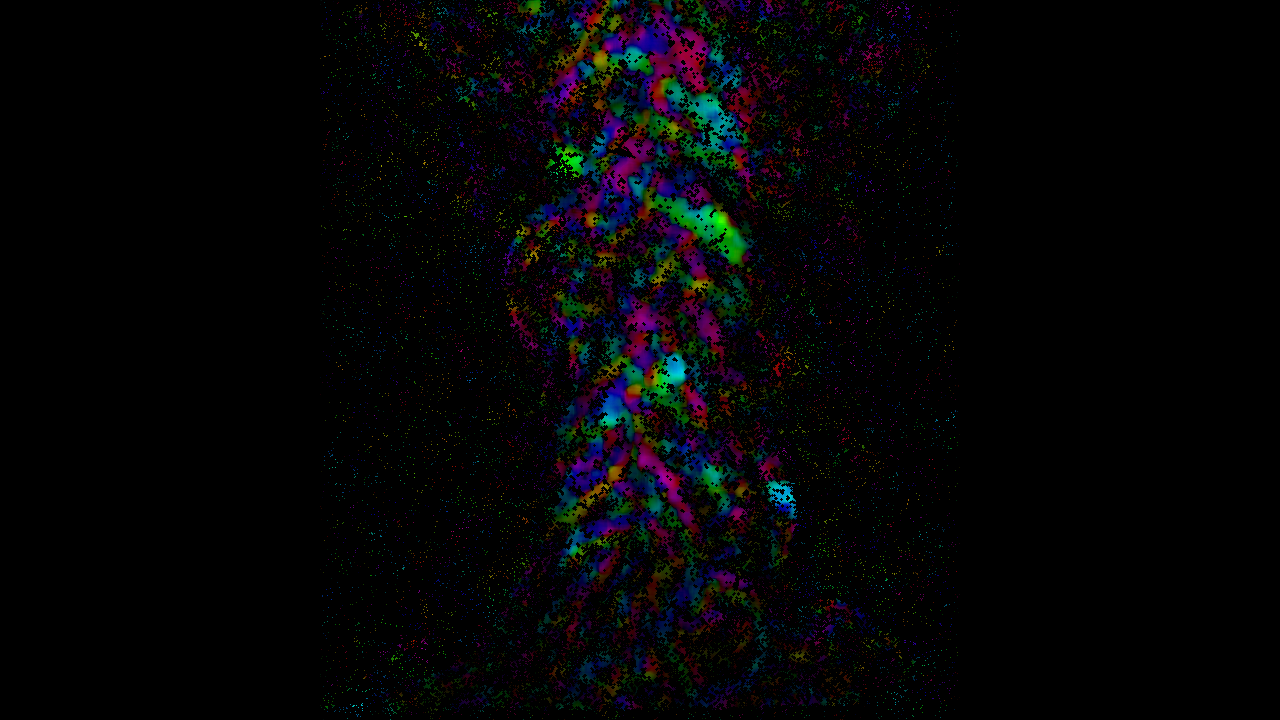}
		&\includegraphics[clip,trim={12cm 0cm 12cm 0cm},width=\linewidth]{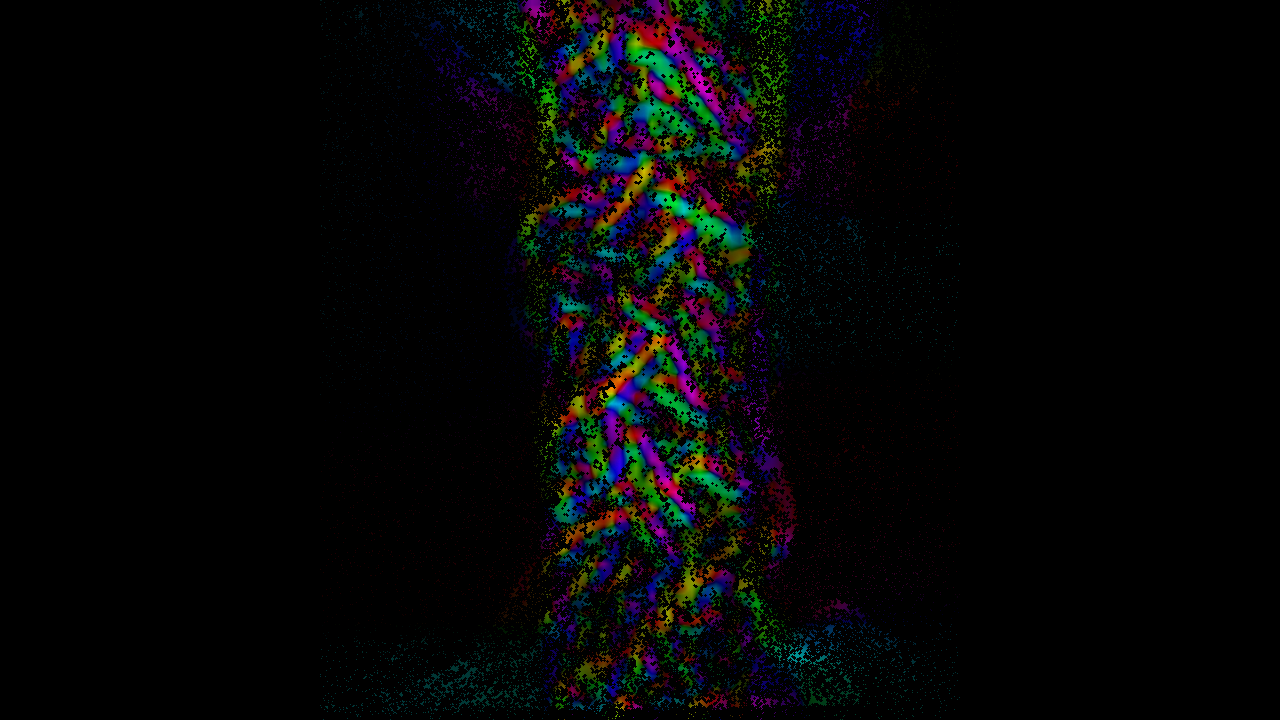}
        \\

        \rotatebox{90}{\makecell{Hair dryer}} %
		&\gframe{\includegraphics[clip,trim={12cm 3cm 12cm 7cm},width=\linewidth]{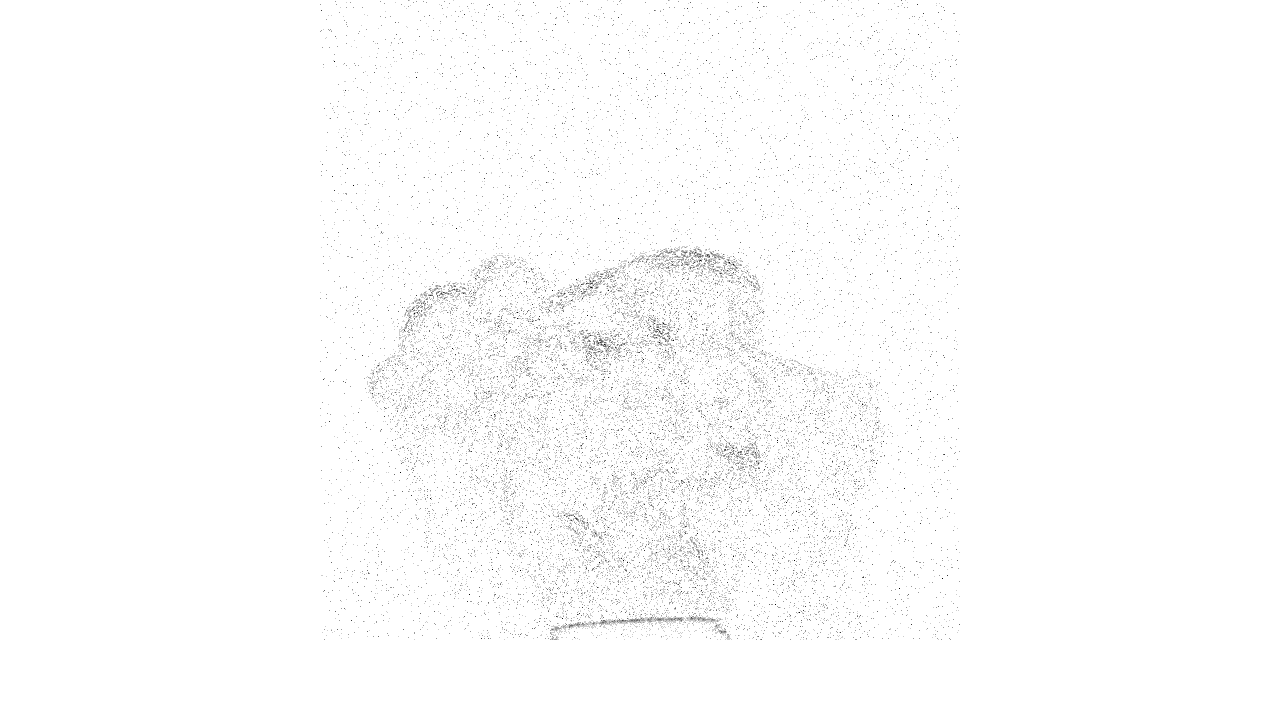}}
		&\includegraphics[clip,trim={12cm 3cm 12cm 7cm},width=\linewidth]{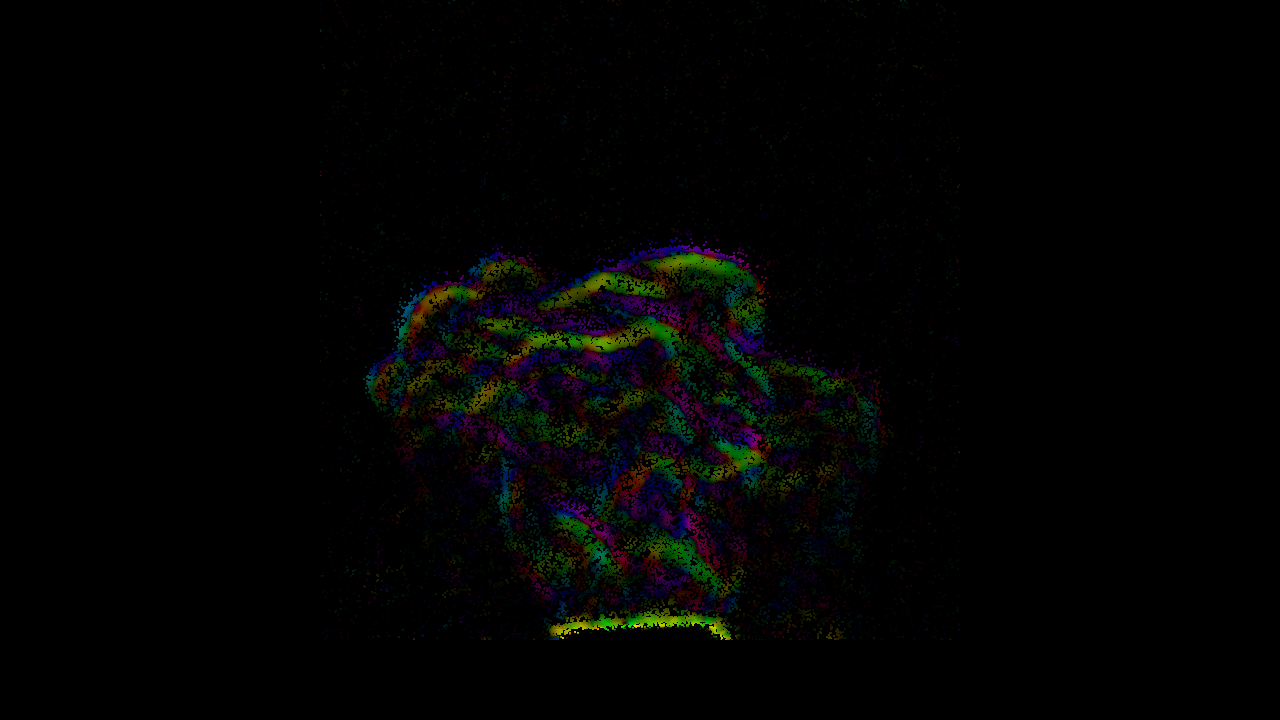}
		&\includegraphics[clip,trim={12cm 3cm 12cm 7cm},width=\linewidth]{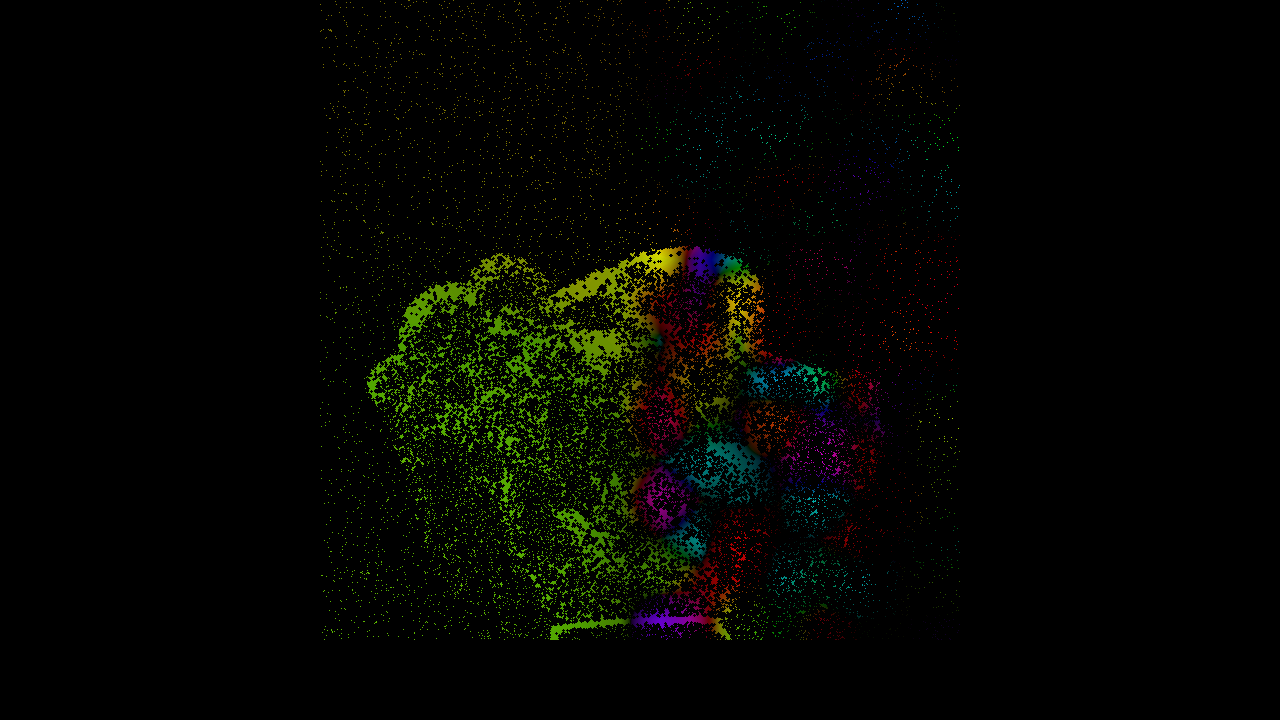}
		&\includegraphics[clip,trim={12cm 3cm 12cm 7cm},width=\linewidth]{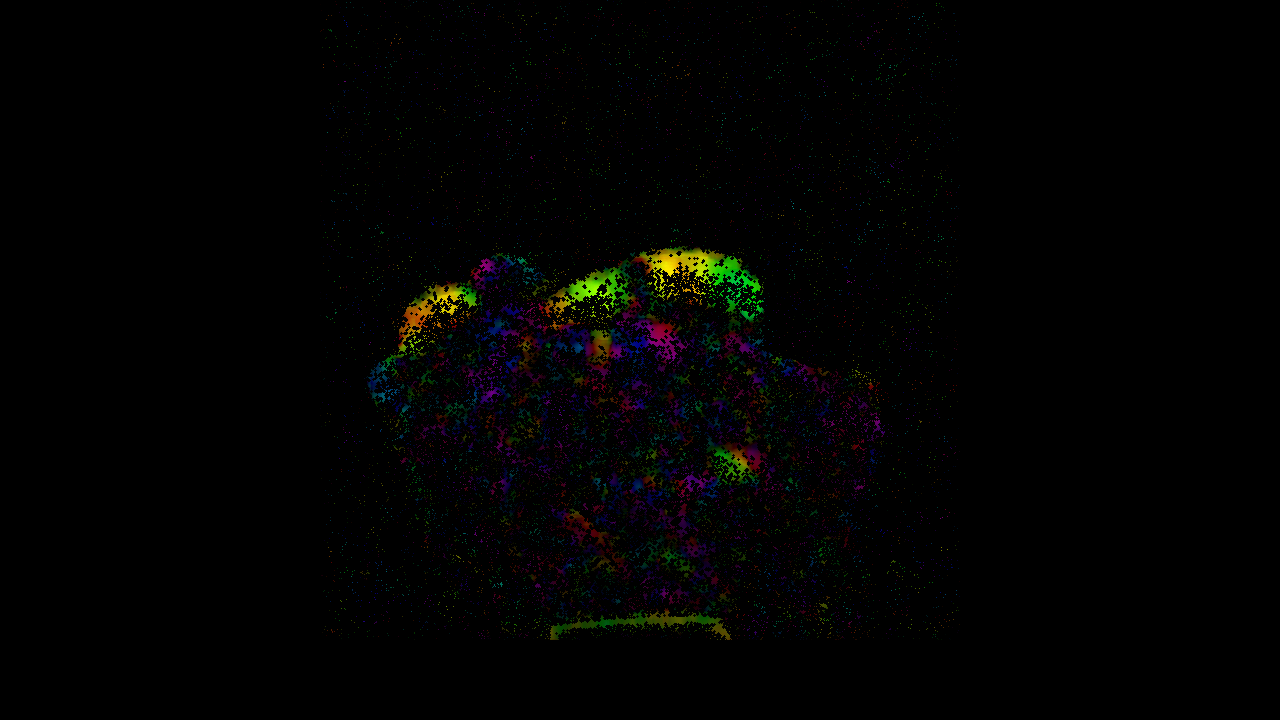}
		&\includegraphics[clip,trim={12cm 3cm 12cm 7cm},width=\linewidth]{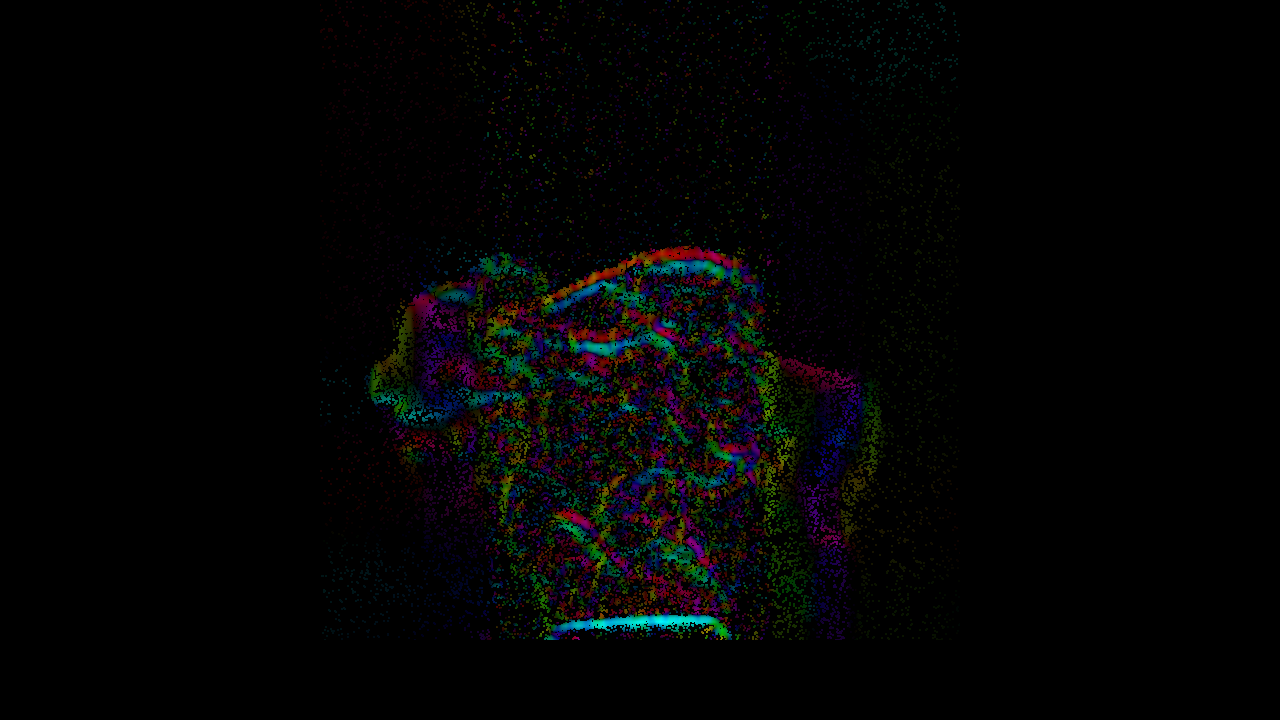}
		\\

        \rotatebox{90}{\makecell{Hair dryer}} %
		&\gframe{\includegraphics[clip,trim={15cm 10cm 20cm 4cm},width=\linewidth]{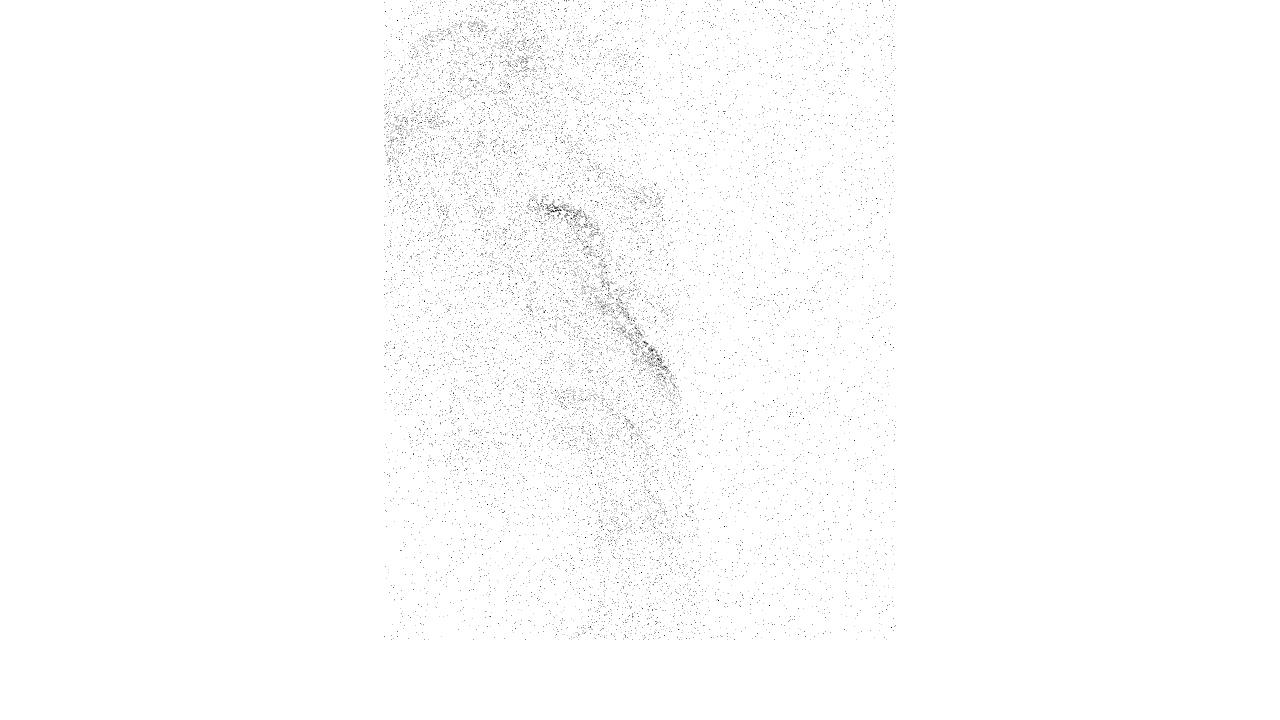}}
		&\includegraphics[clip,trim={15cm 10cm 20cm 4cm},width=\linewidth]{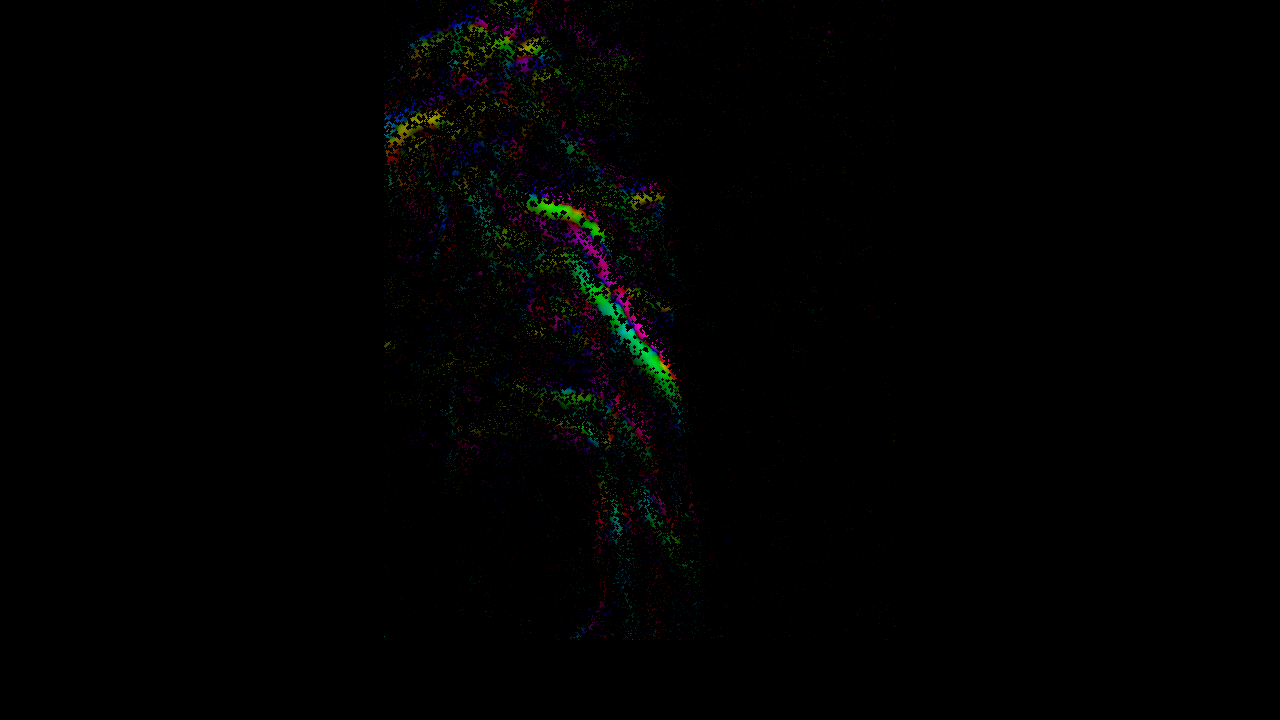}
		&\includegraphics[clip,trim={15cm 10cm 20cm 4cm},width=\linewidth]{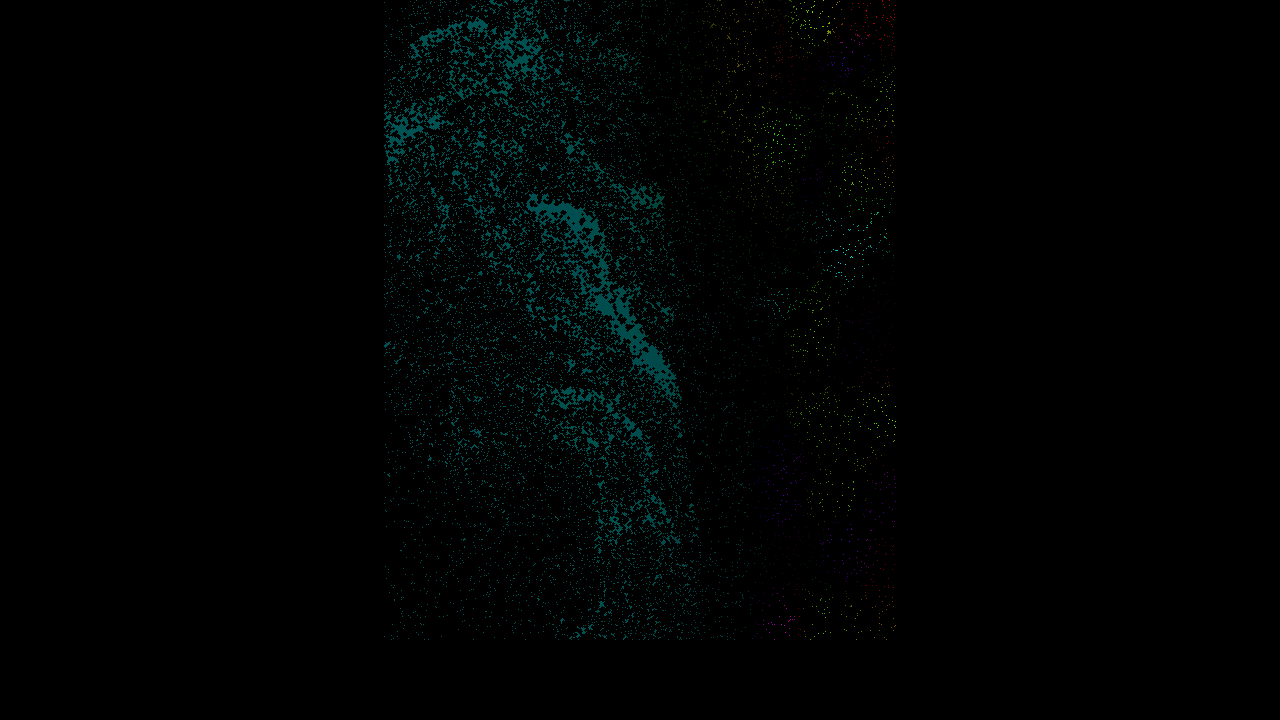}
 		&\includegraphics[clip,trim={15cm 10cm 20cm 4cm},width=\linewidth]{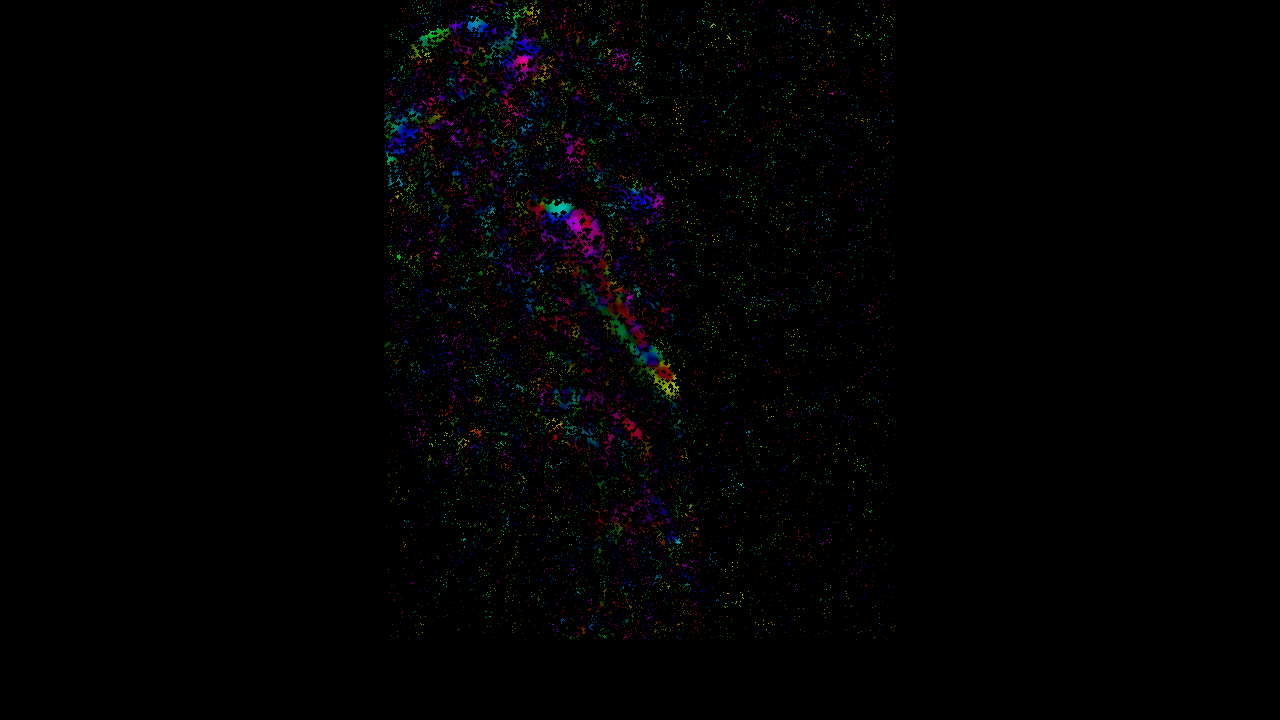}
		&\includegraphics[clip,trim={15cm 10cm 20cm 4cm},width=\linewidth]{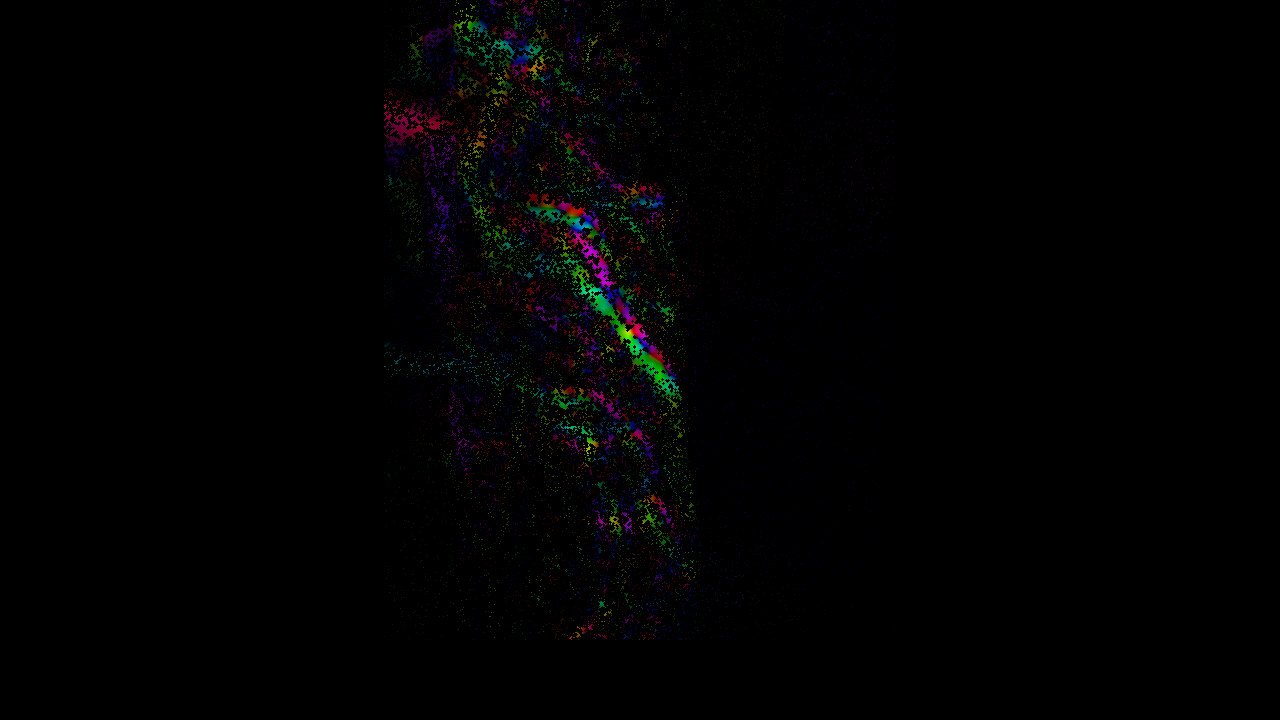}
		\\

        \rotatebox{90}{\makecell{Breathing}}
		&\gframe{\includegraphics[clip,trim={25cm 11cm 8cm 4cm},width=\linewidth]{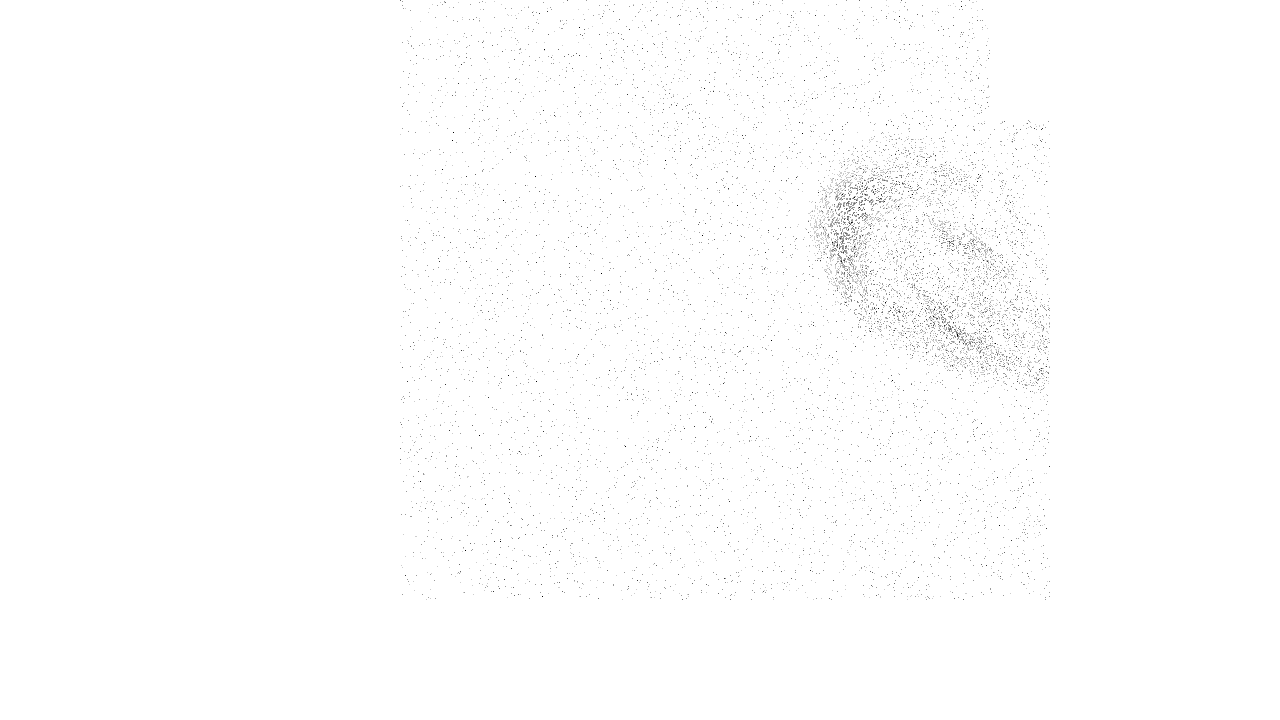}}
		&\includegraphics[clip,trim={25cm 11cm 8cm 4cm},width=\linewidth]{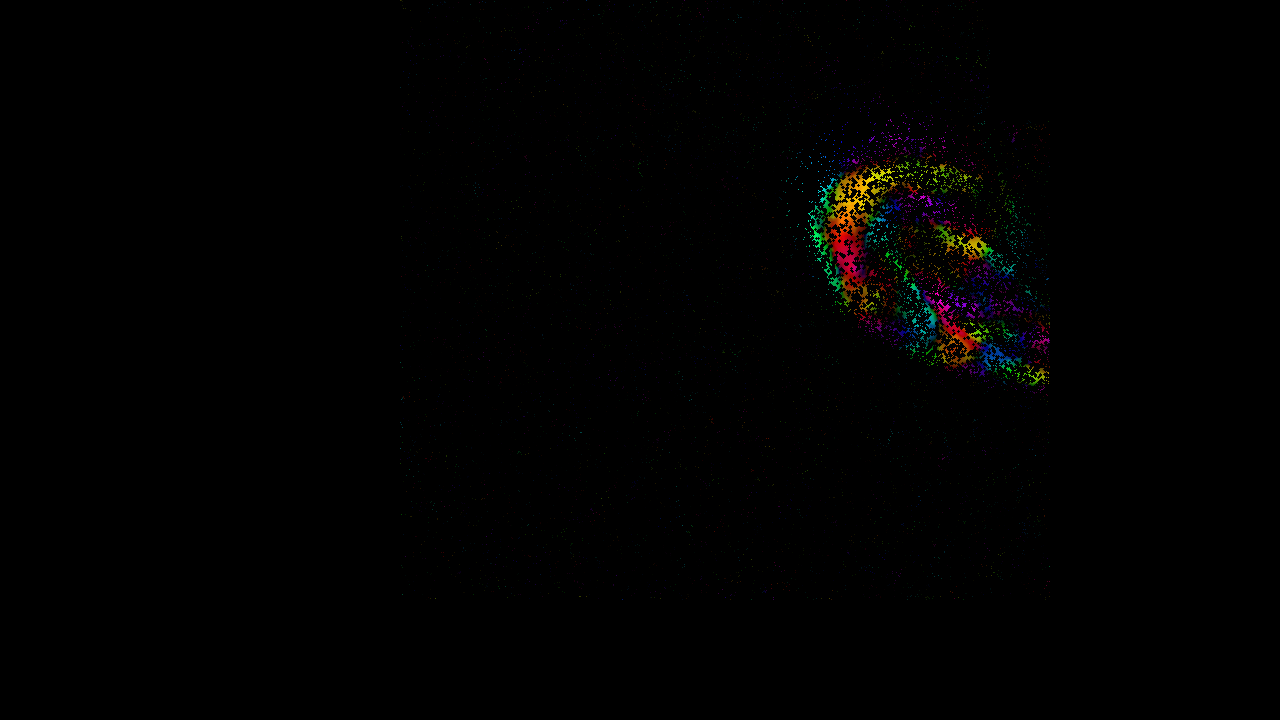}
		&\includegraphics[clip,trim={25cm 11cm 8cm 4cm},width=\linewidth]{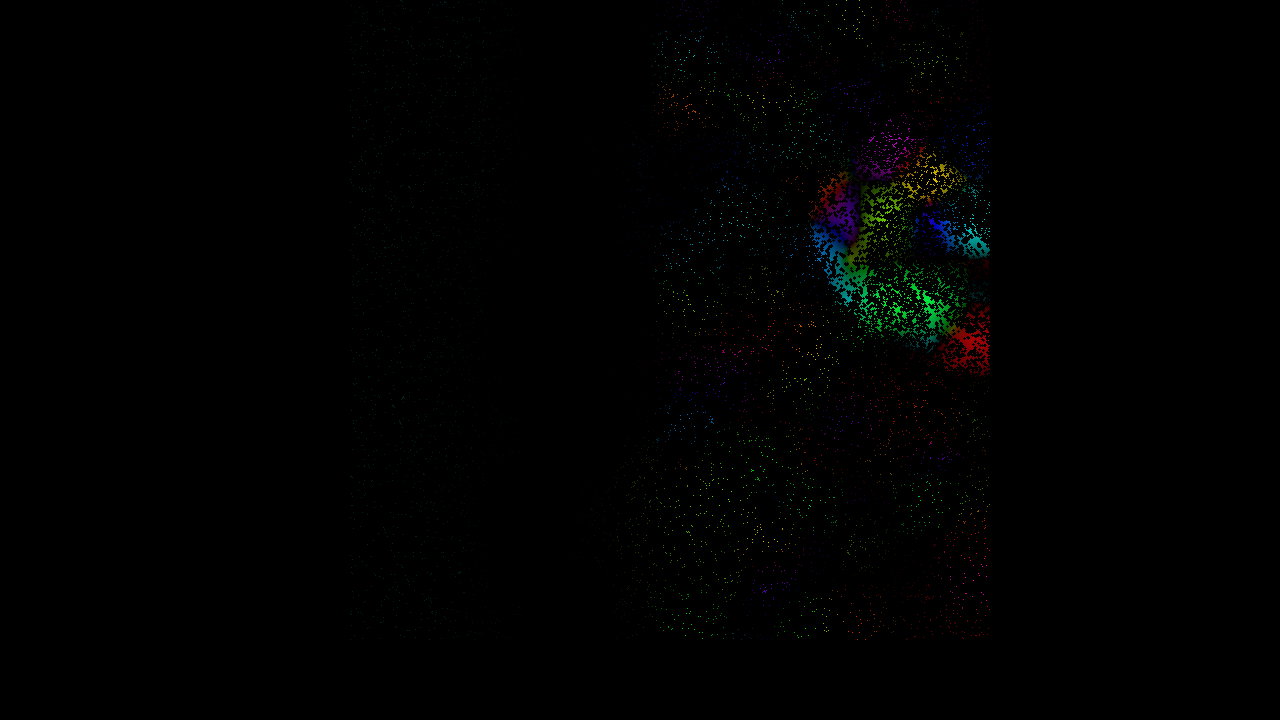}
		&\gframe{\includegraphics[clip,trim={25cm 11cm 8cm 4cm},width=\linewidth]{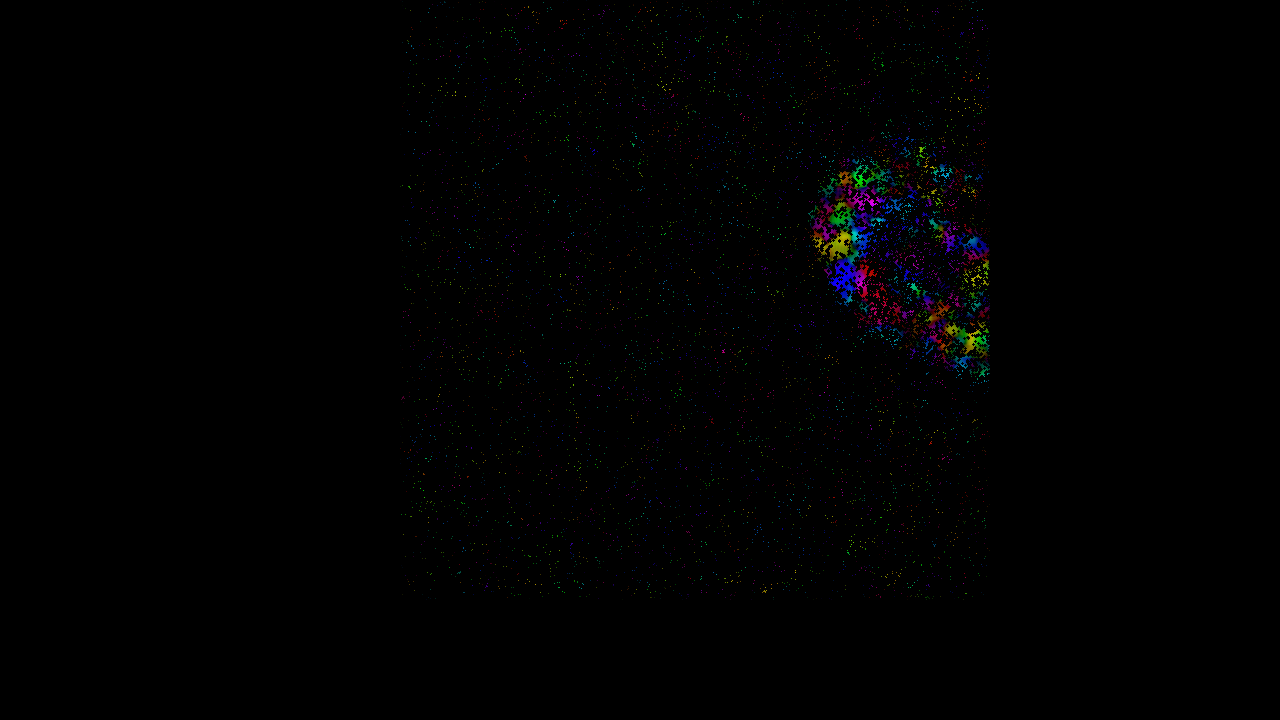}}
		&\includegraphics[clip,trim={25cm 11cm 8cm 4cm},width=\linewidth]{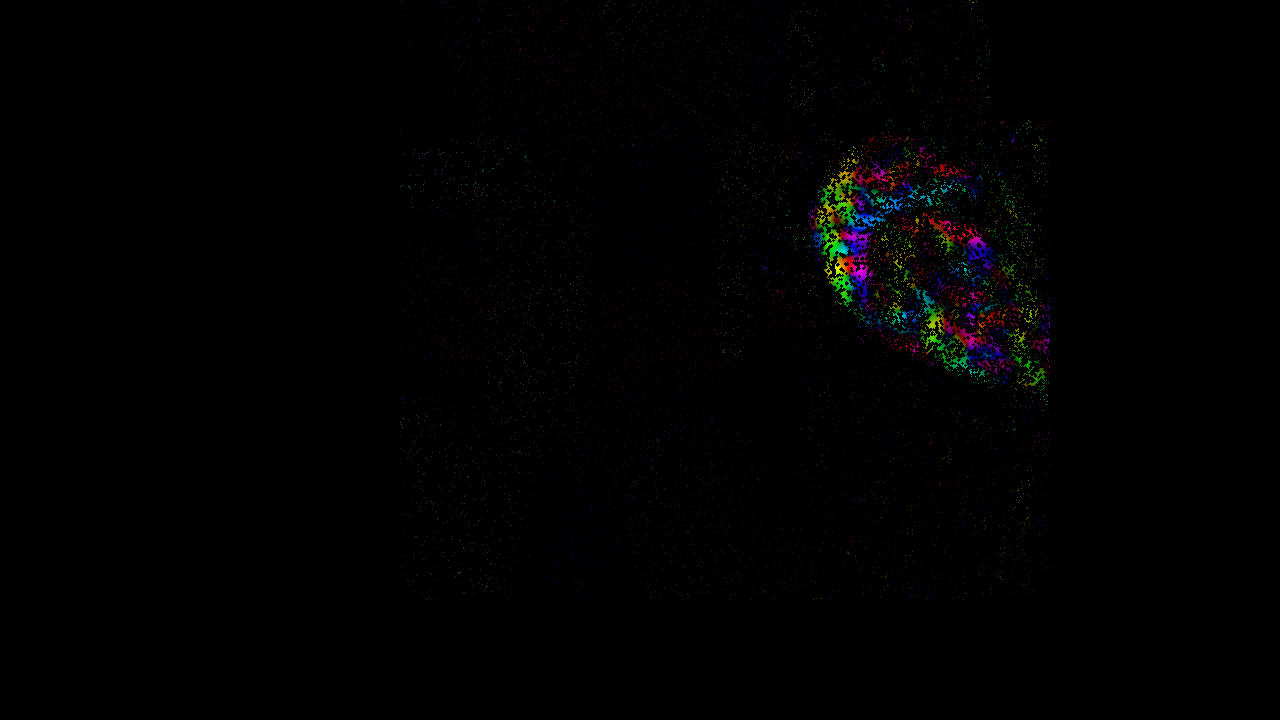}
        \\

		& \textbf{(a)} Input events %
		& \textbf{(b)} GT (Flow from frames)
		& \textbf{(c)} MCM \cite{Shiba22eccv} (events only)
		& \textbf{(d)} Flow from E2VID frames
		& \textbf{(e)} Ours
	\end{tabular}
	}
	\caption{Qualitative comparison between different flow estimation methods. 
	}
	\label{fig:main:compare}
\end{figure*}

%% file: floats/fig_hdr.tex
\def\figmethodwidth{.48\linewidth}
\begin{figure}[t]
\centering
\begin{subfigure}{\figmethodwidth}
  \centering
  {\includegraphics[clip,trim={17cm 0cm 10cm 8cm},width=\linewidth]{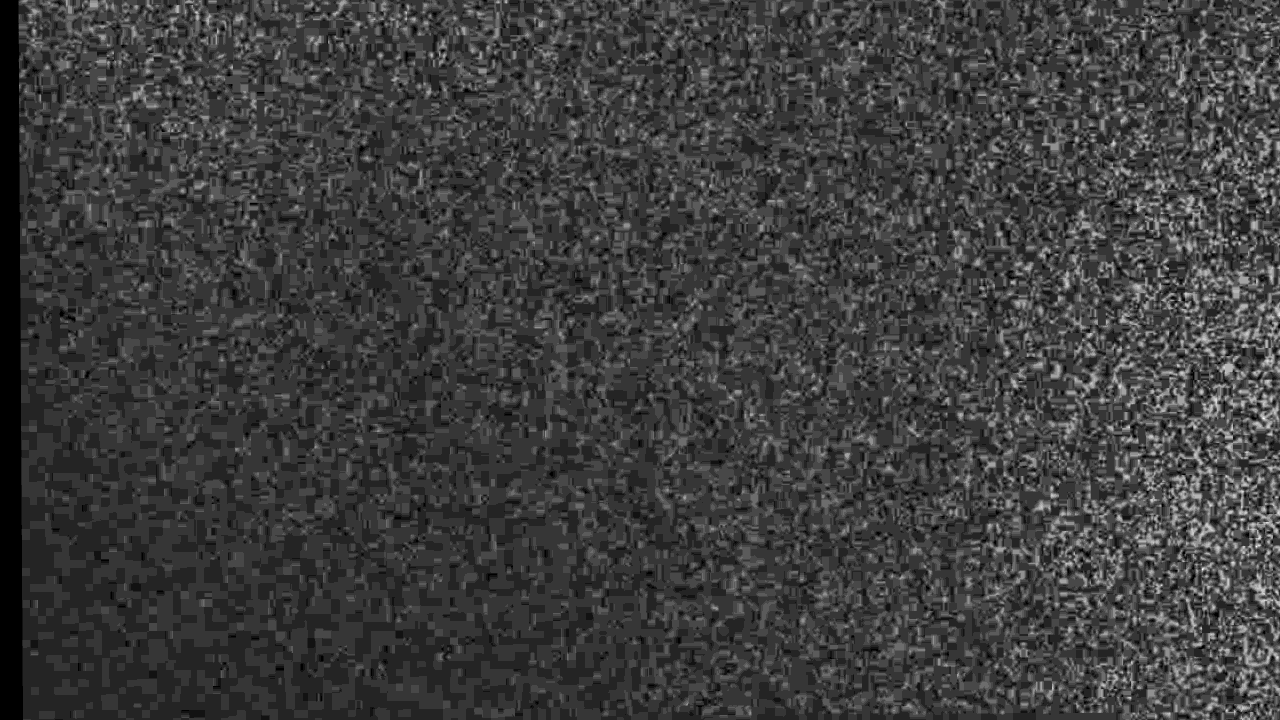}}
\end{subfigure}\;
\begin{subfigure}{\figmethodwidth}
  \centering
  \gframe{\includegraphics[clip,trim={17cm 0cm 10cm 8cm},width=\linewidth]{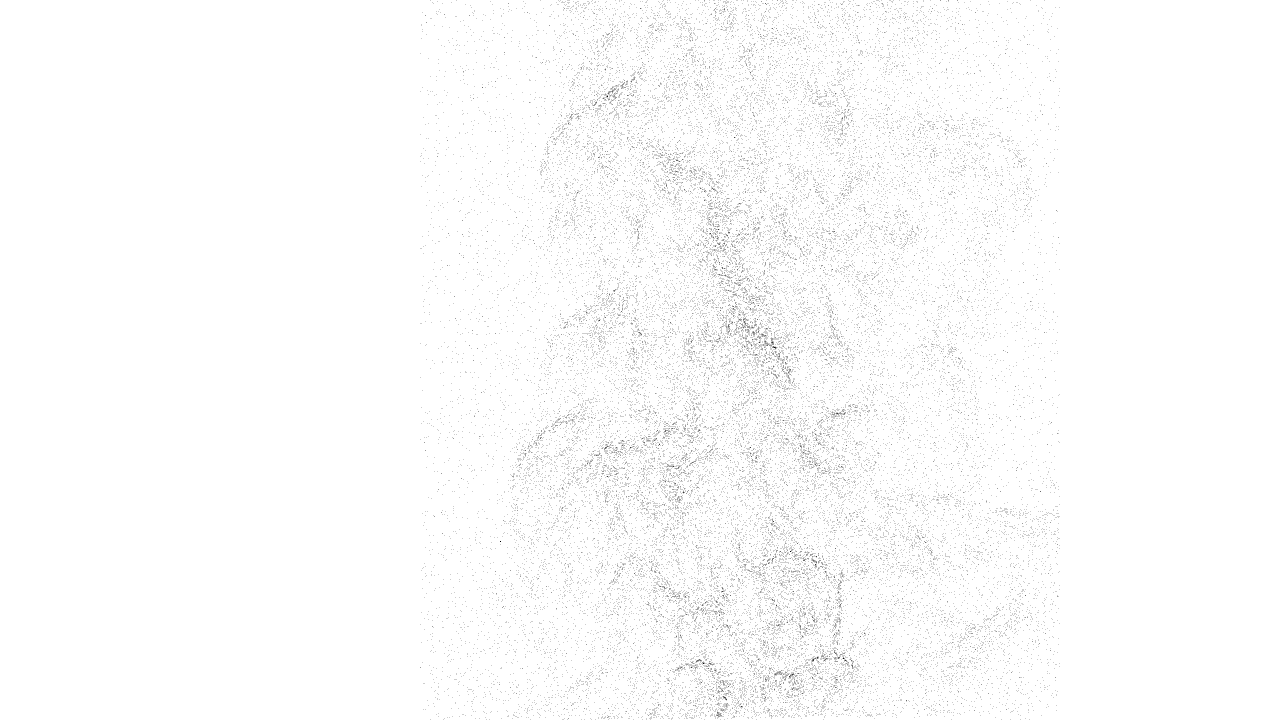}}
\end{subfigure}\\[0.5ex]
\begin{subfigure}{\figmethodwidth}
  \centering
  {\includegraphics[clip,trim={17cm 0cm 10cm 8cm},width=\linewidth]{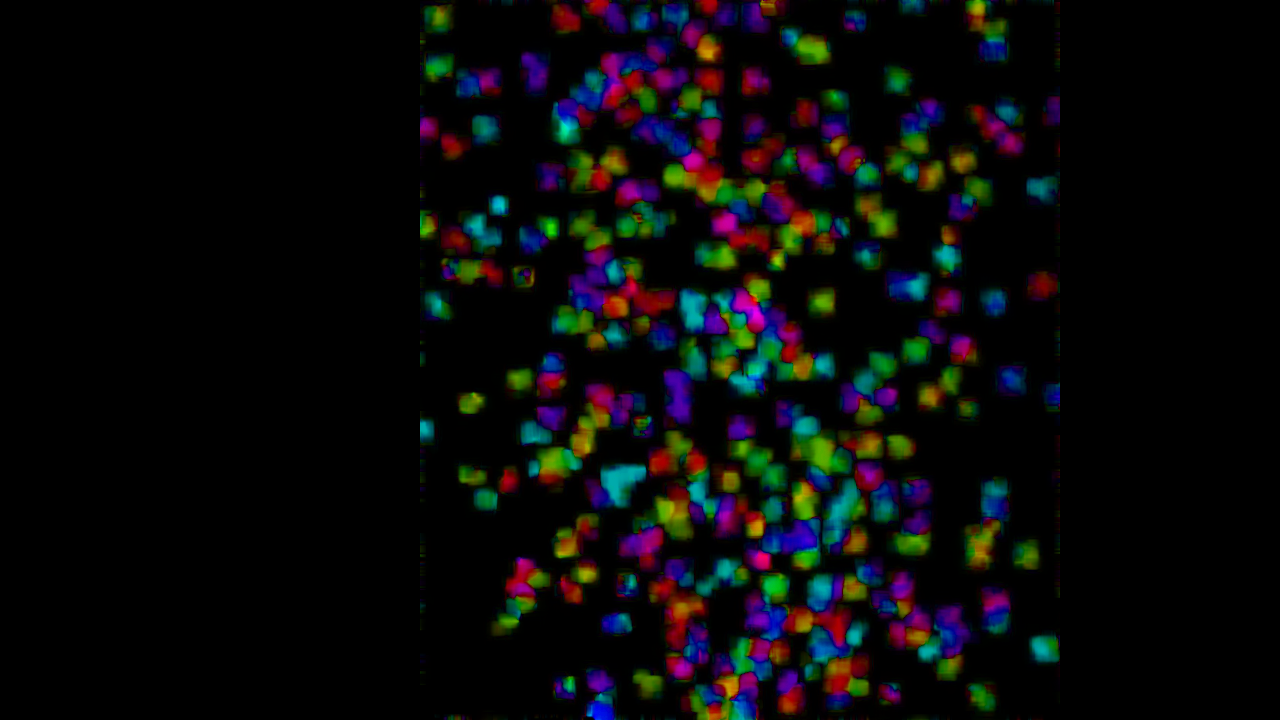}}
  \caption{\centering Frame-based BOS}
\end{subfigure}\;
\begin{subfigure}{\figmethodwidth}
  \centering
  \gframe{\includegraphics[clip,trim={17cm 0cm 10cm 8cm},width=\linewidth]{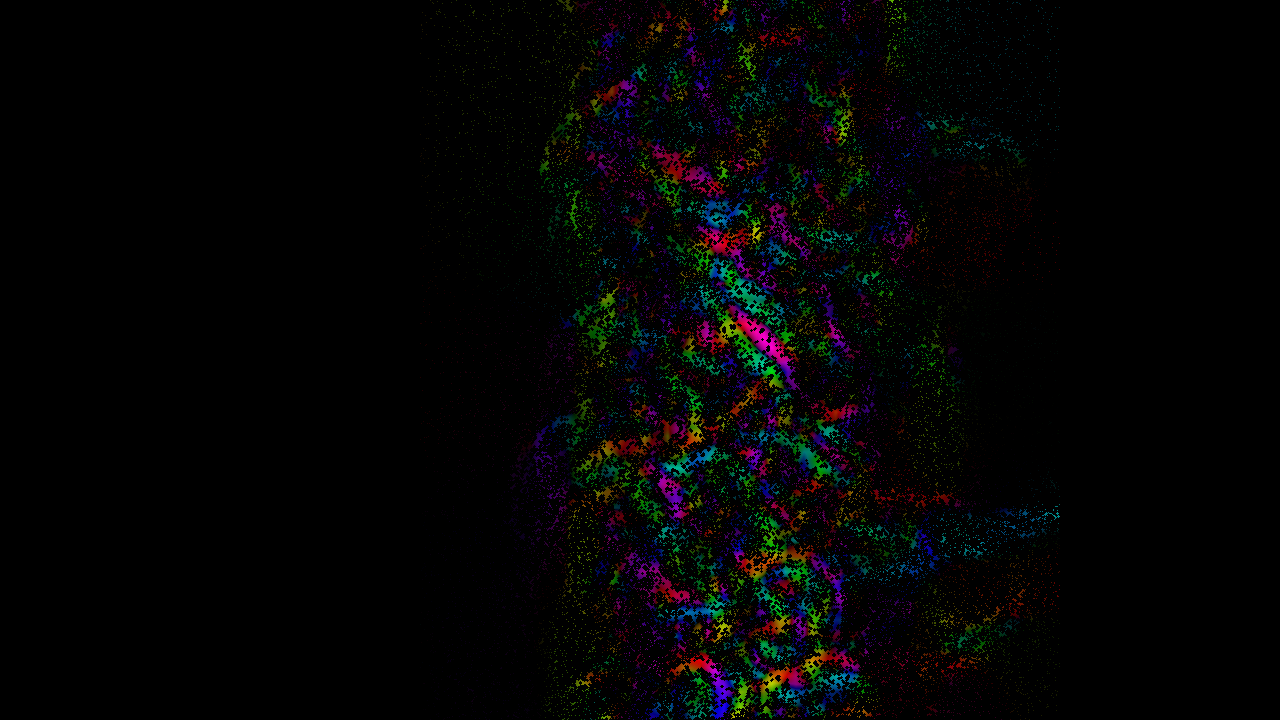}}
  \caption{\centering Ours}
\end{subfigure}
\caption{\emph{Schlieren imaging under poor illumination (225 \si{\lux})}. 
(a) Frame-based methods suffer from the limited dynamic range of the frames, resulting in unrealistic flows with artifacts despite using all grayscale range available for the frames (normalization).
(b) The proposed method produces a realistic flow, similar to the event data, which is visible due to the HDR nature of events.
}
\label{fig:hdr}
\end{figure}

%% file: floats/fig_kymogram.tex
\begin{figure}[t]
\centering
\begin{subfigure}{0.5\linewidth}
  \centering
  {\includegraphics[trim={9.5cm 4.2cm 10cm 5cm},clip,width=\linewidth]{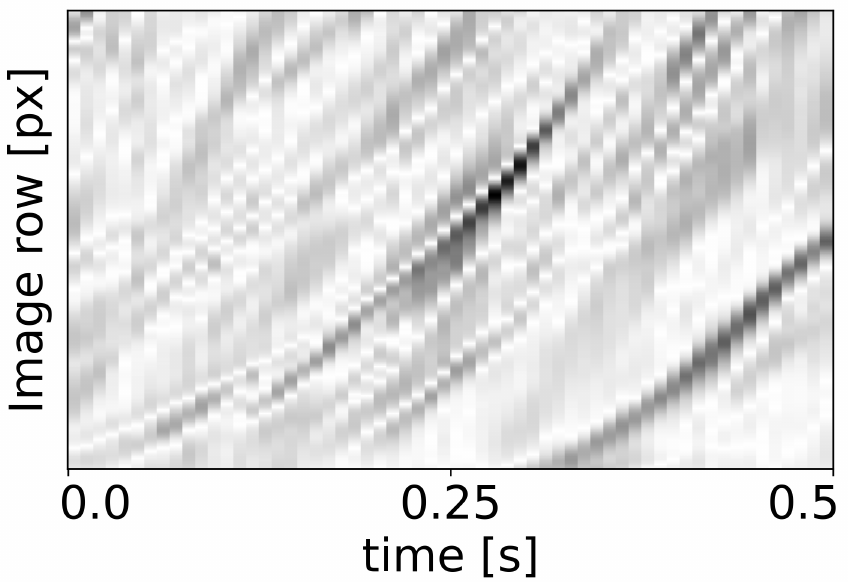}}
  \caption{\centering Frame-based kymogram}
  \label{fig:kymogram:frames}
\end{subfigure}%
\begin{subfigure}{0.5\linewidth}
  \centering
  {\includegraphics[trim={9.5cm 4.2cm 10cm 5cm},clip,width=\linewidth]{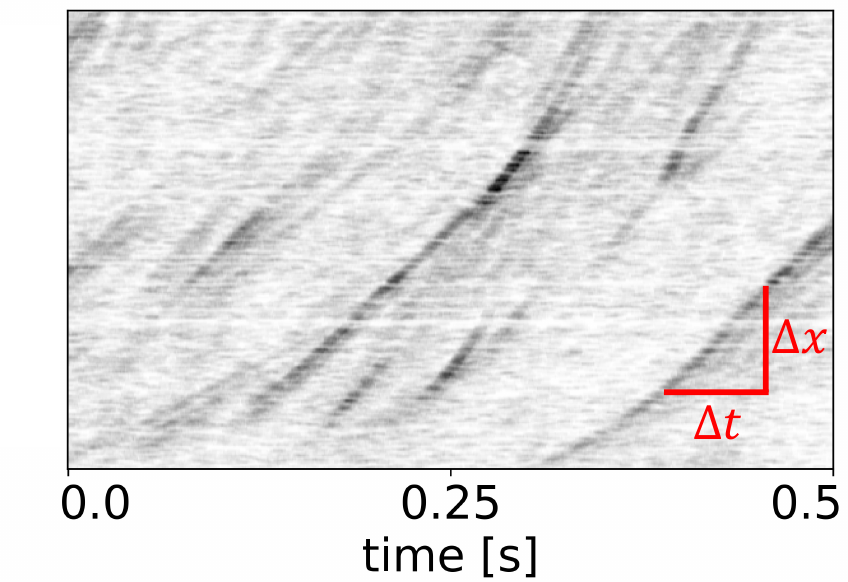}}
  \caption{\centering Event-based kymogram}
  \label{fig:kymogram:events}
\end{subfigure}
\caption{\emph{Kymograms (space-time plots) and velocimetry} for a 0.5 second excerpt of a hotplate sequence.
(a) The frame-based schlieren imaging is limited to the temporal resolution of the camera (120 Hz).
(b) The event-based schlieren can recover higher temporal resolution (e.g., 1200 Hz) thanks to its data characteristics.
}
\label{fig:kymogram}
\end{figure}

%% file: floats/supplfig_declining_light.tex
\sisetup{round-mode=places,round-precision=1,round-integer-to-decimal}
\global\long\def\figWidth{0.215\linewidth}
\begin{figure*}[ht!]
	\centering
    {\scriptsize
    \setlength{\tabcolsep}{2pt}
	\begin{tabular}{
	>{\centering\arraybackslash}m{0.3cm} 
	>{\centering\arraybackslash}m{\figWidth} 
	>{\centering\arraybackslash}m{\figWidth} 
	>{\centering\arraybackslash}m{\figWidth} 
	>{\centering\arraybackslash}m{\figWidth}}

        \rotatebox{90}{\makecell{4000~\si{\lux}}}
		&{\includegraphics[clip,trim={11cm 3.5cm 11cm 4cm},width=\linewidth]{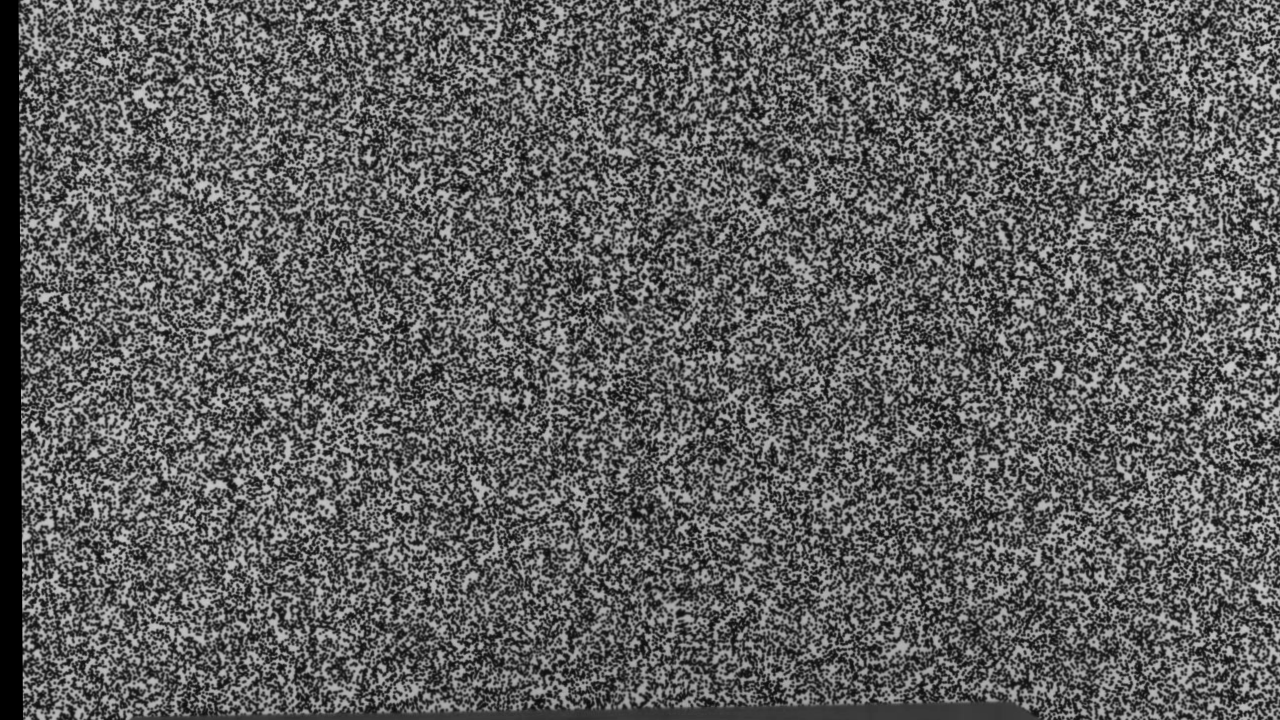}}
		&{\includegraphics[clip,trim={11cm 3.5cm 11cm 4cm},width=\linewidth]{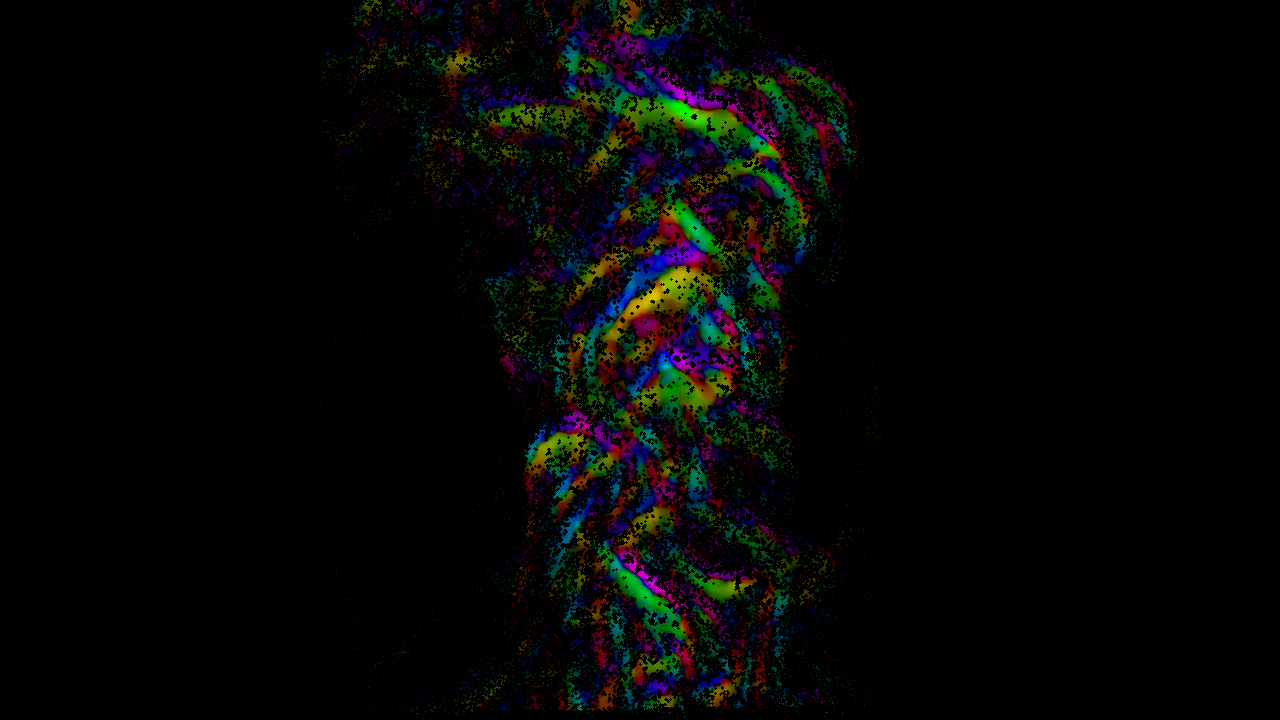}}
		&\gframe{\includegraphics[clip,trim={11cm 3.5cm 11cm 4cm},width=\linewidth]{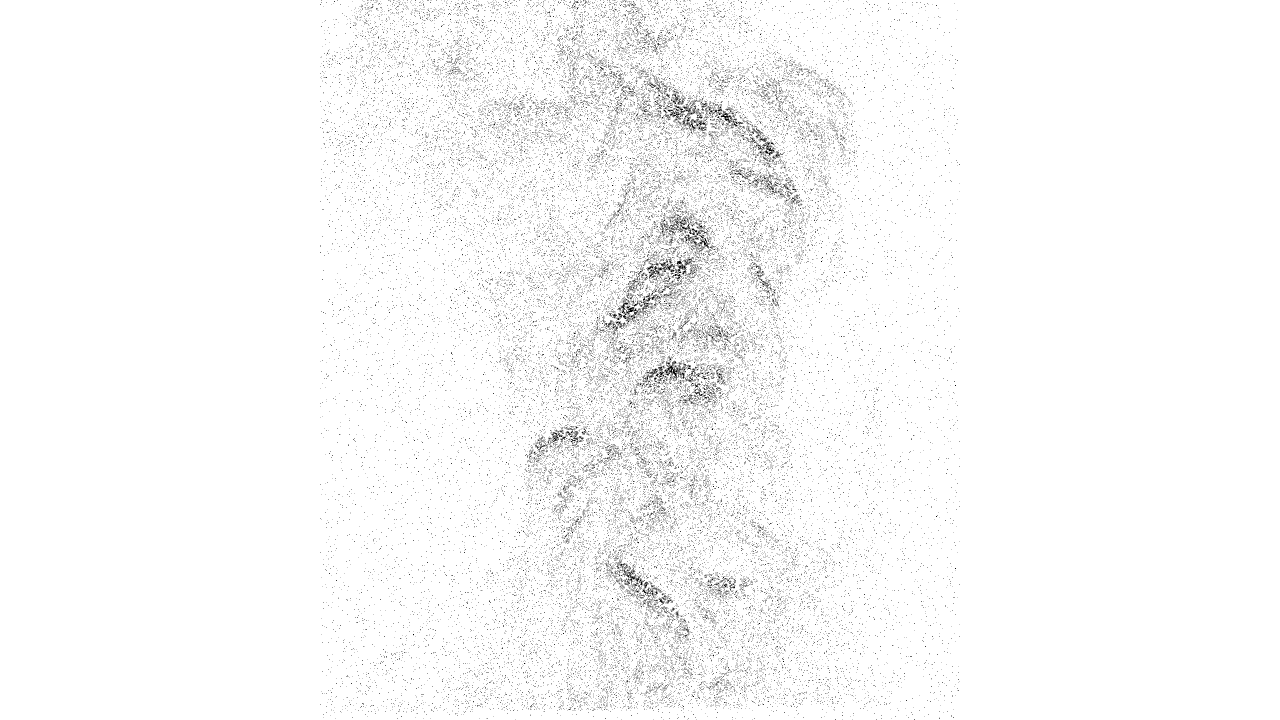}}
		&{\includegraphics[clip,trim={11cm 3.5cm 11cm 4cm},width=\linewidth]{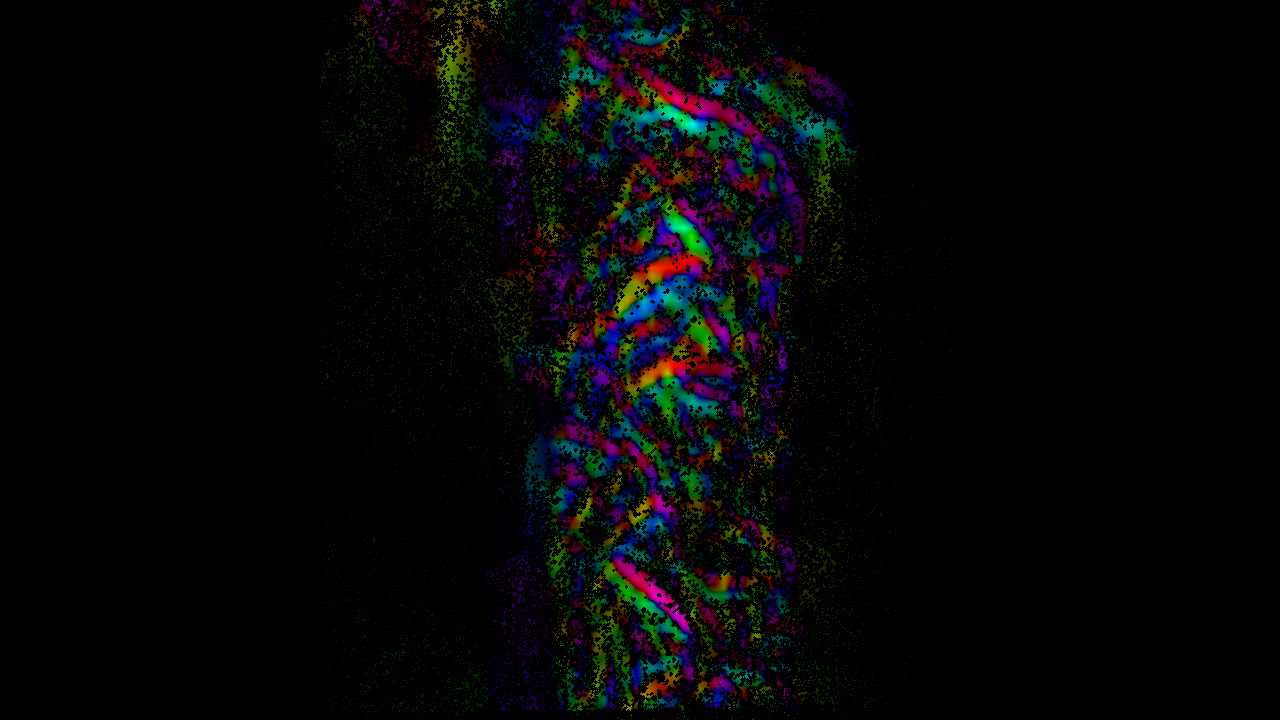}}
        \\

        \rotatebox{90}{\makecell{2000~\si{\lux}}}
		&{\includegraphics[clip,trim={11cm 2.5cm 11cm 5cm},width=\linewidth]{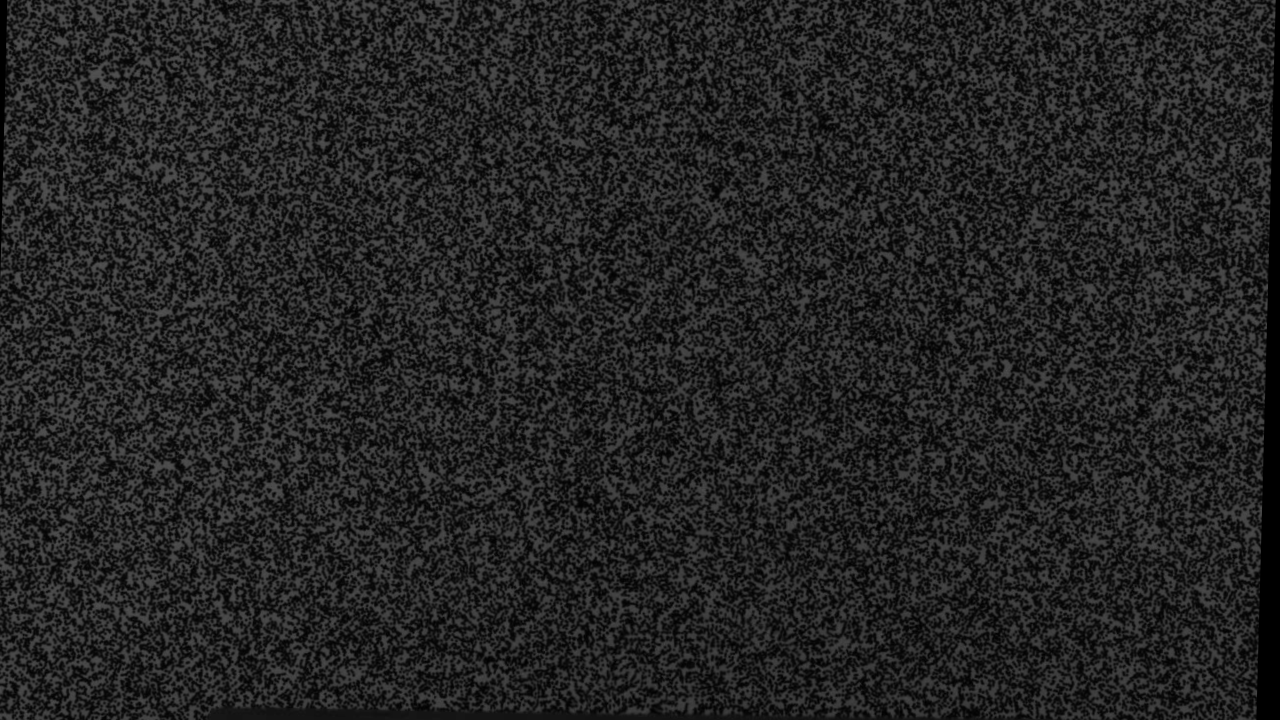}}
		&{\includegraphics[clip,trim={11cm 2.5cm 11cm 5cm},width=\linewidth]{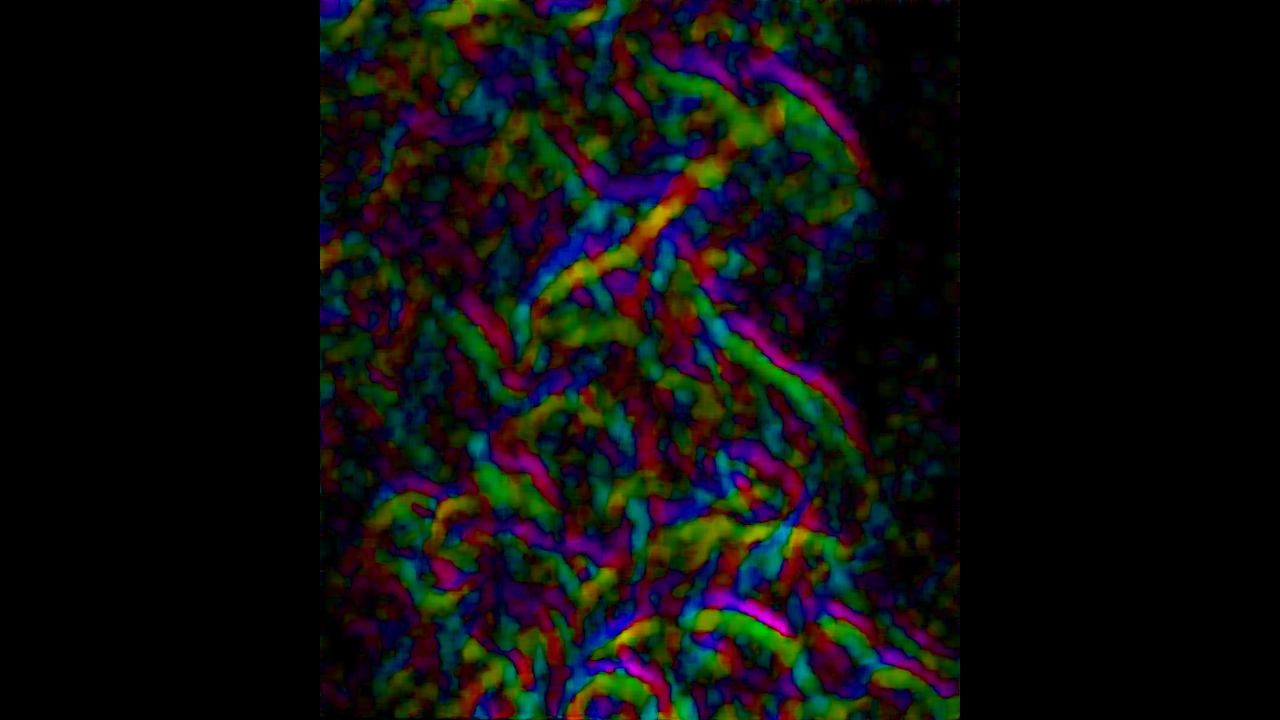}}
		&\gframe{\includegraphics[clip,trim={11cm 2.5cm 11cm 5cm},width=\linewidth]{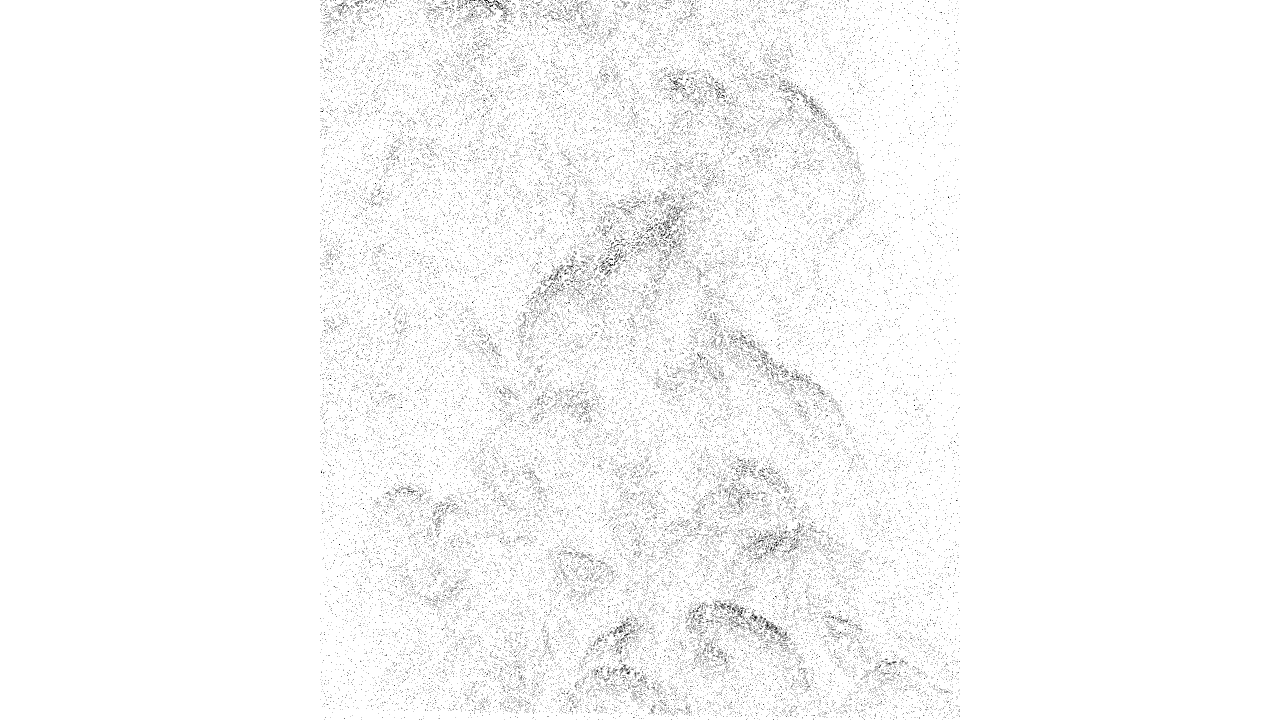}}
		&{\includegraphics[clip,trim={11cm 2.5cm 11cm 5cm},width=\linewidth]{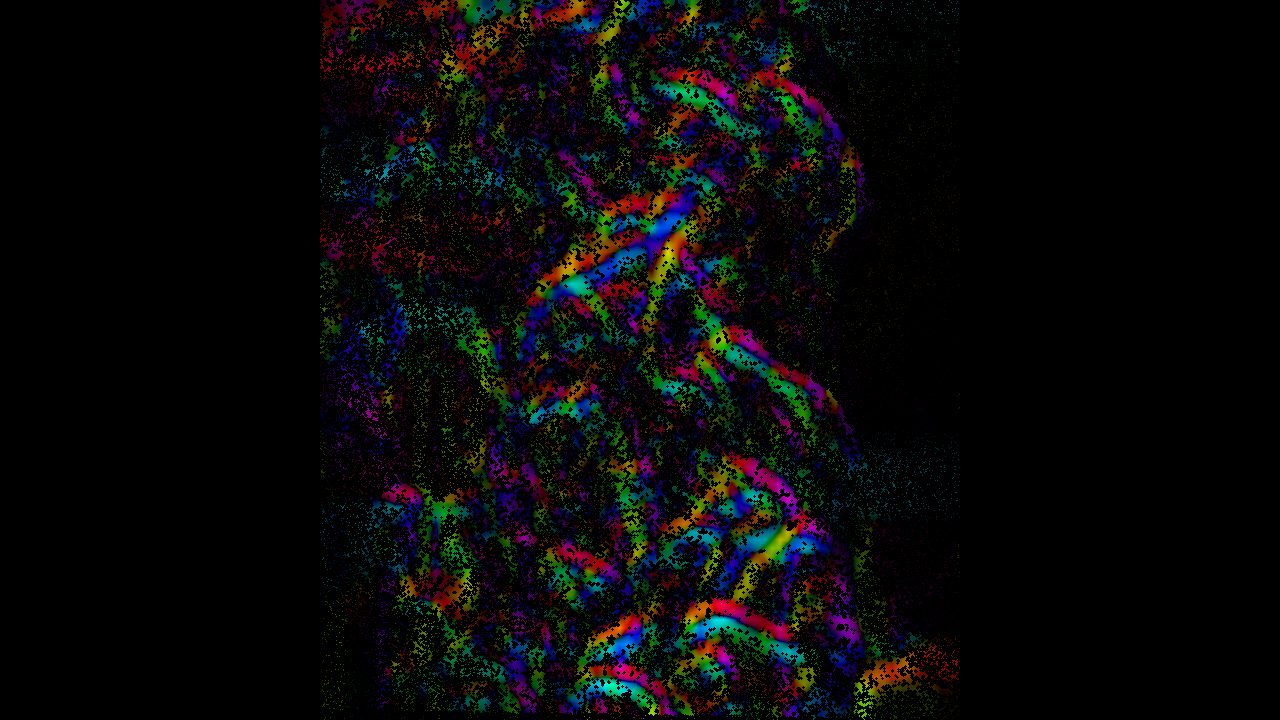}}
        \\

        \rotatebox{90}{\makecell{1000~\si{\lux}}}
		&{\includegraphics[clip,trim={11cm 5cm 11cm 2.5cm},width=\linewidth]{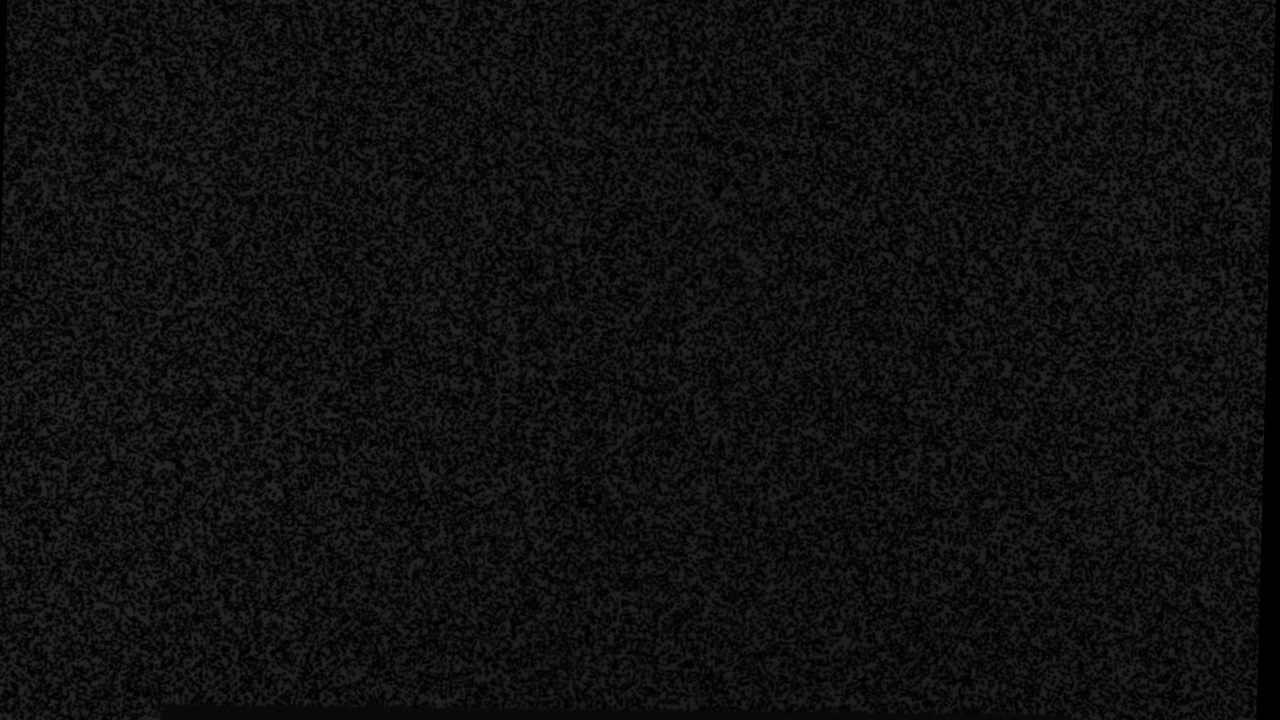}}
		&{\includegraphics[clip,trim={11cm 5cm 11cm 2.5cm},width=\linewidth]{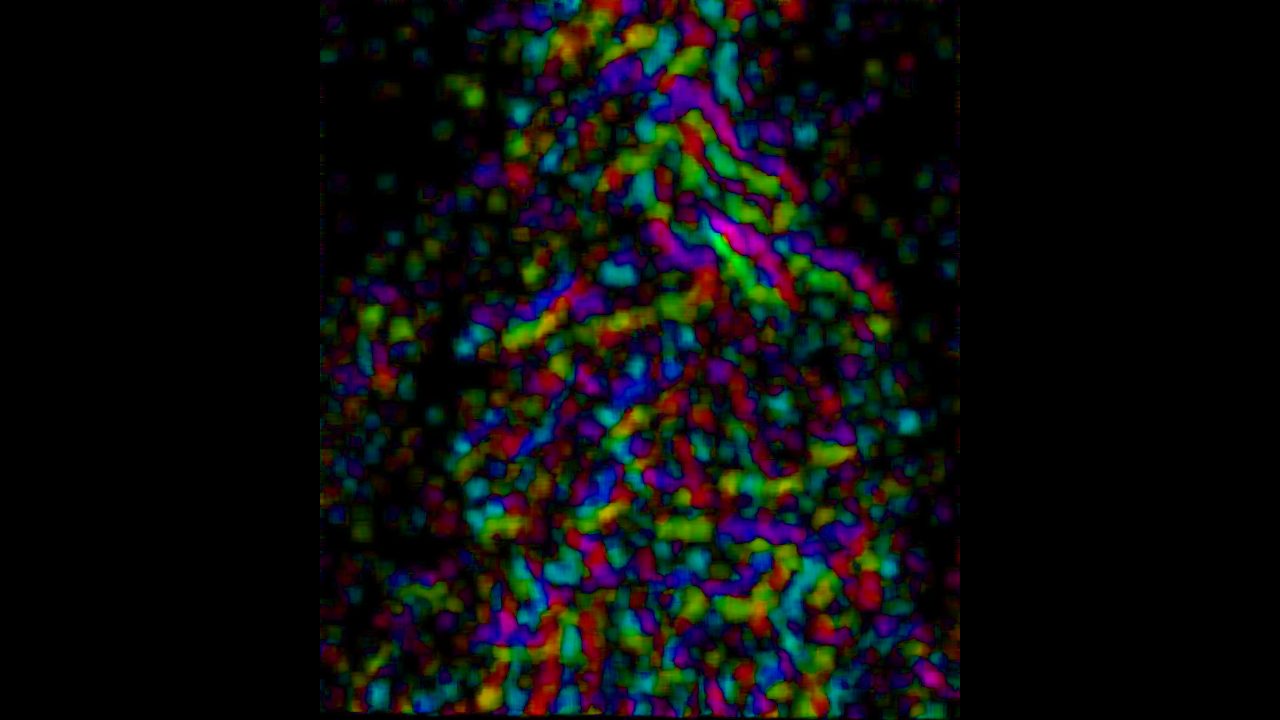}}
		&\gframe{\includegraphics[clip,trim={11cm 5cm 11cm 2.5cm},width=\linewidth]{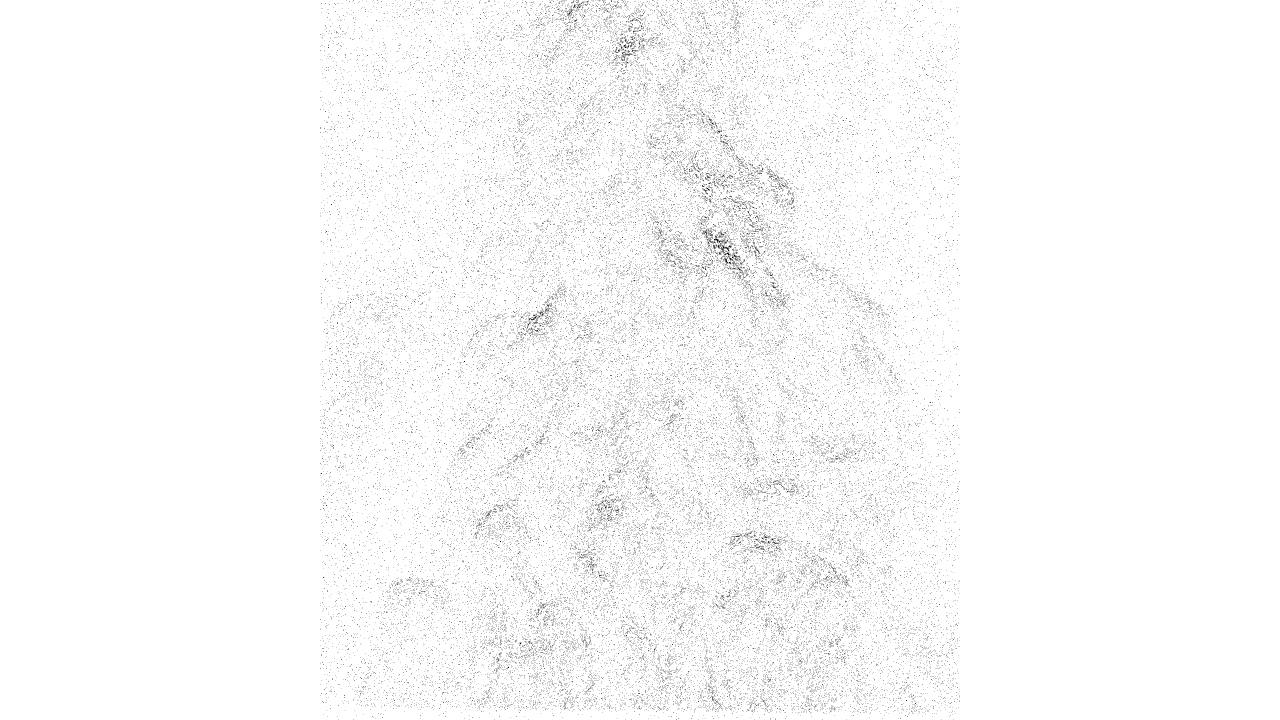}}
		&{\includegraphics[clip,trim={11cm 5cm 11cm 2.5cm},width=\linewidth]{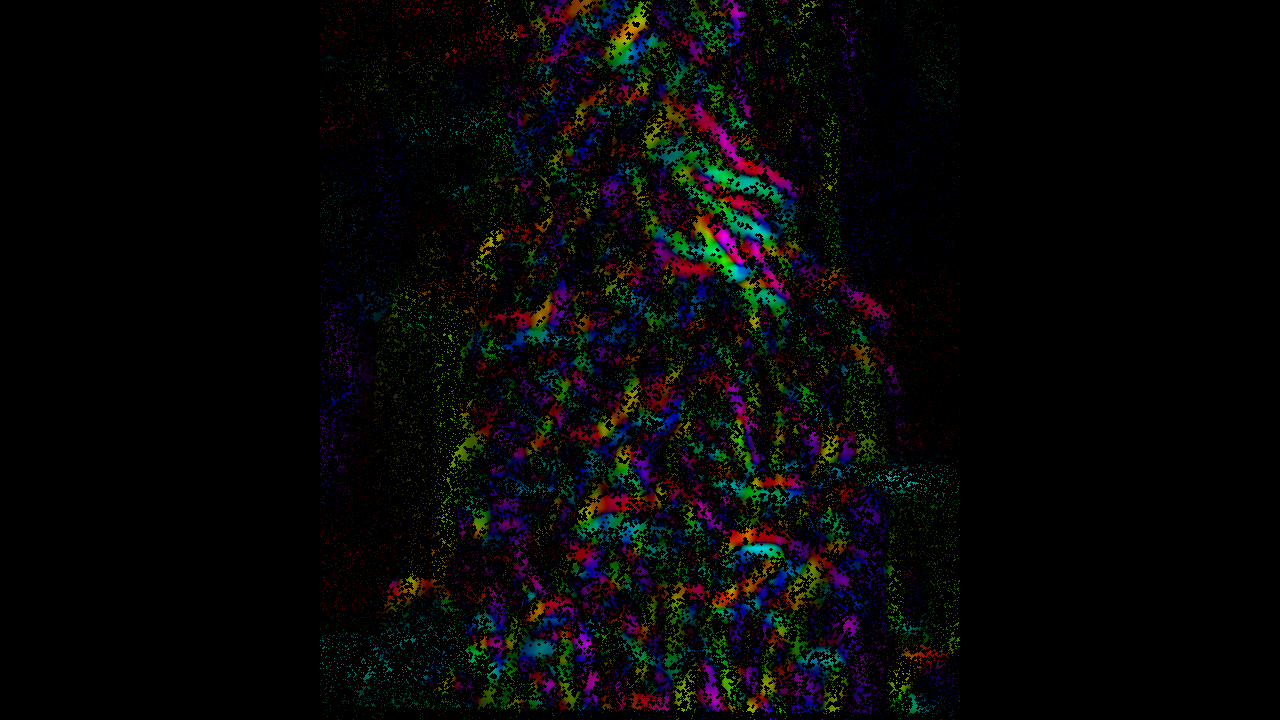}}
        \\

        \rotatebox{90}{\makecell{500~\si{\lux}}}
		&{\includegraphics[clip,trim={10cm 2.5cm 12cm 5cm},width=\linewidth]{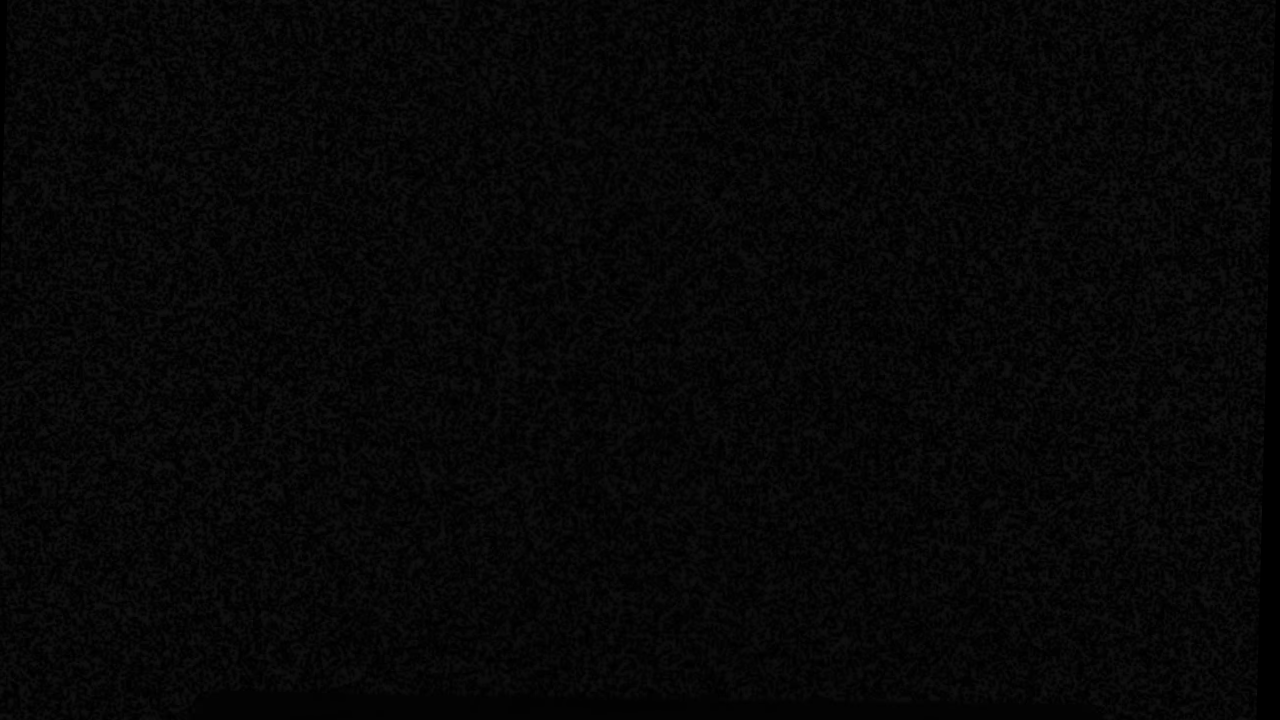}}
		&{\includegraphics[clip,trim={10cm 2.5cm 12cm 5cm},width=\linewidth]{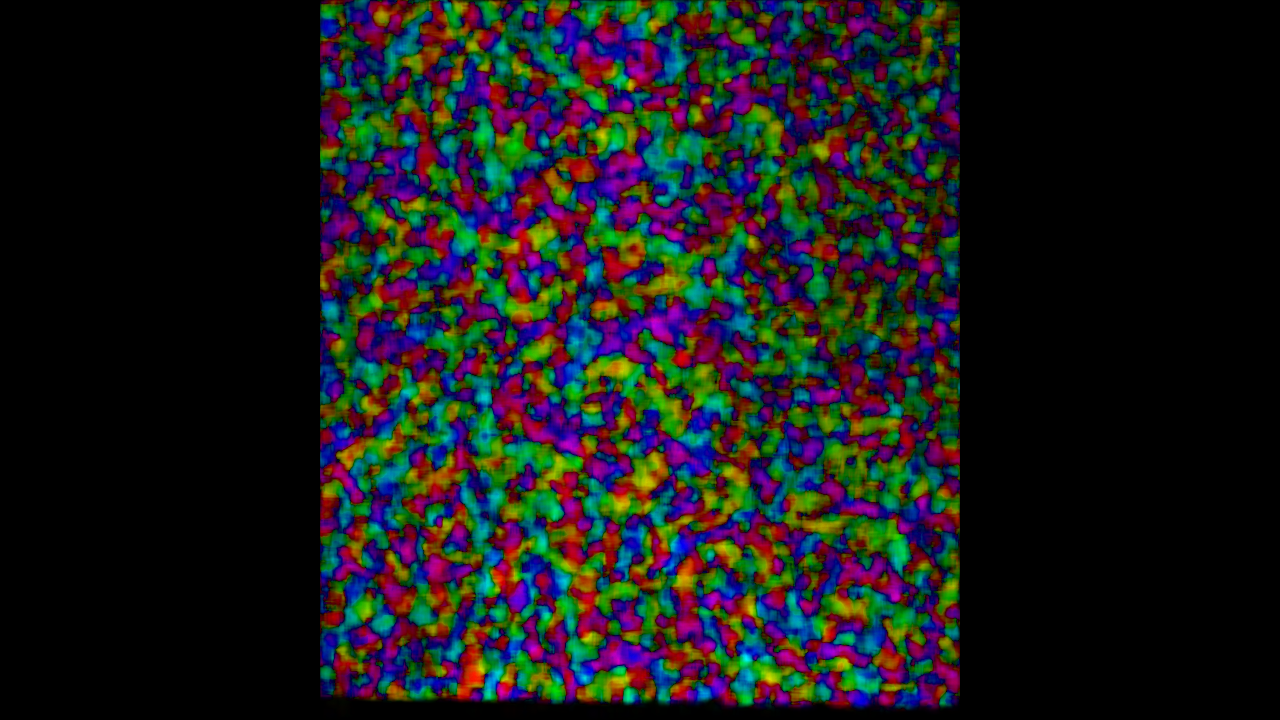}}
		&\gframe{\includegraphics[clip,trim={10cm 2.5cm 12cm 5cm},width=\linewidth]{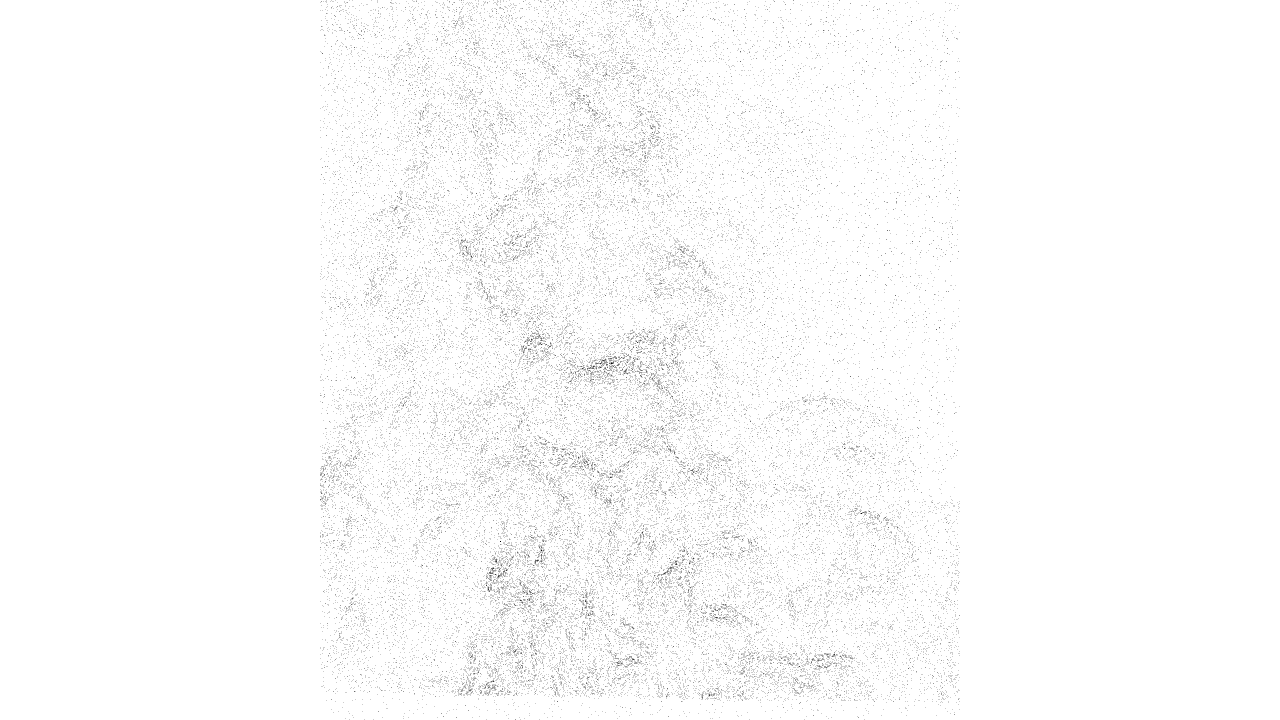}}
		&{\includegraphics[clip,trim={10cm 2.5cm 12cm 5cm},width=\linewidth]{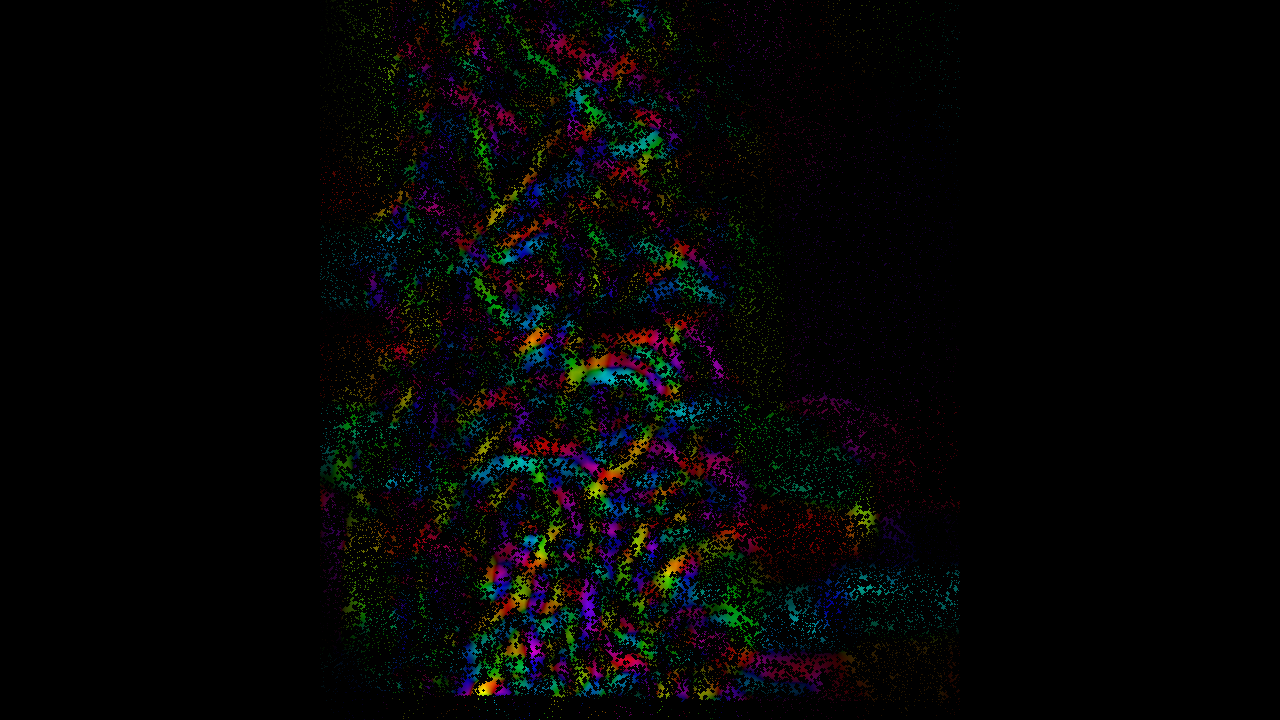}}
        \\

        \rotatebox{90}{\makecell{225~\si{\lux}}}
		&{\includegraphics[clip,trim={14cm 2.5cm 8cm 5cm},width=\linewidth]{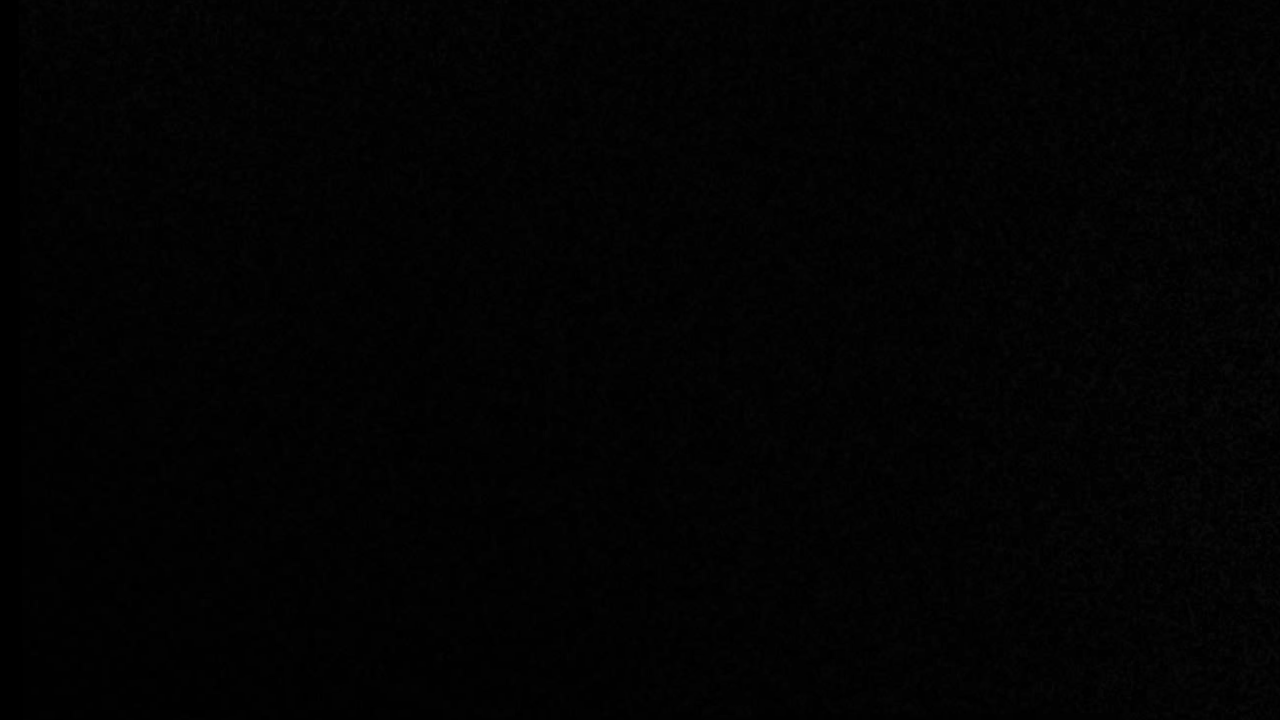}}
		&{\includegraphics[clip,trim={14cm 2.5cm 8cm 5cm},width=\linewidth]{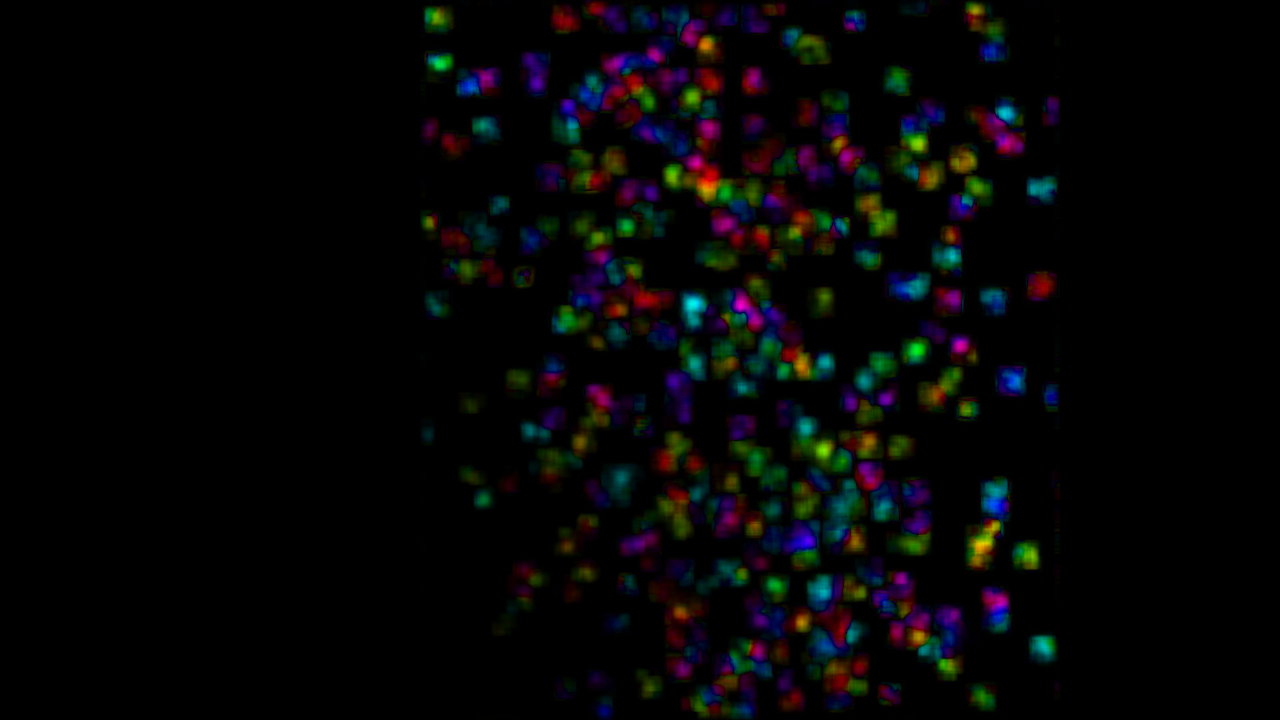}}
		&\gframe{\includegraphics[clip,trim={14cm 2.5cm 8cm 5cm},width=\linewidth]{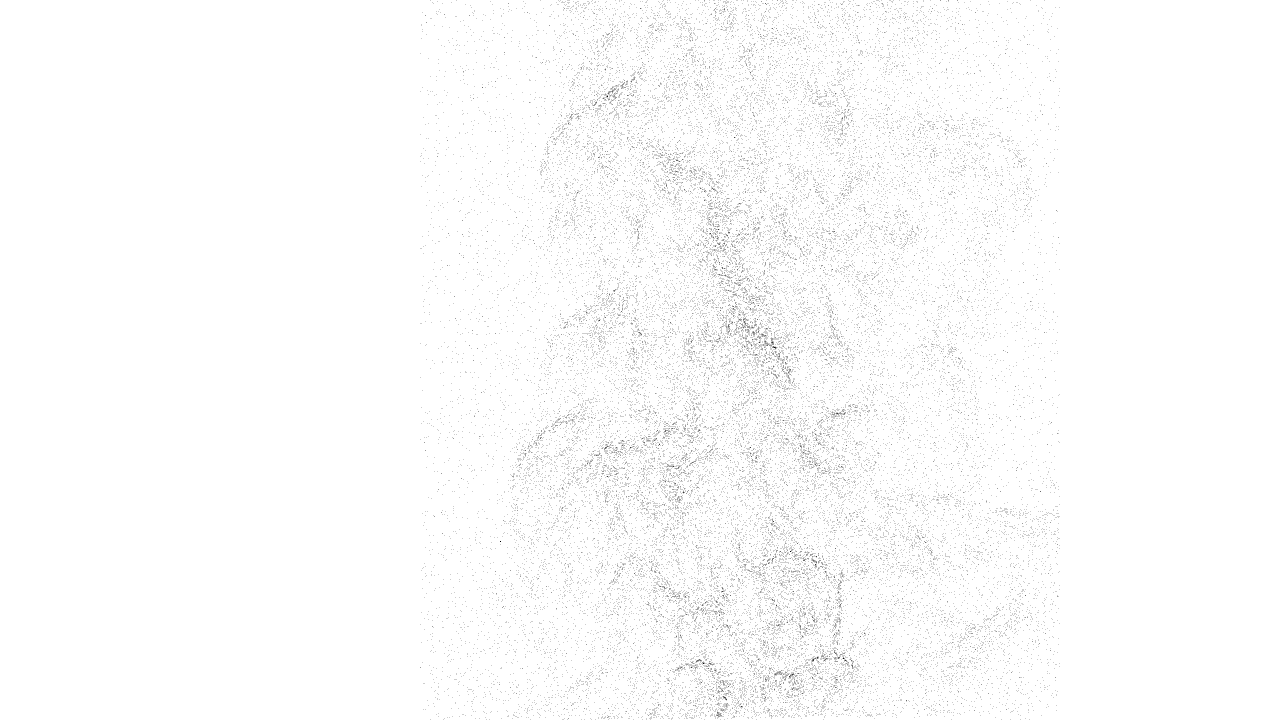}}
		&{\includegraphics[clip,trim={14cm 2.5cm 8cm 5cm},width=\linewidth]{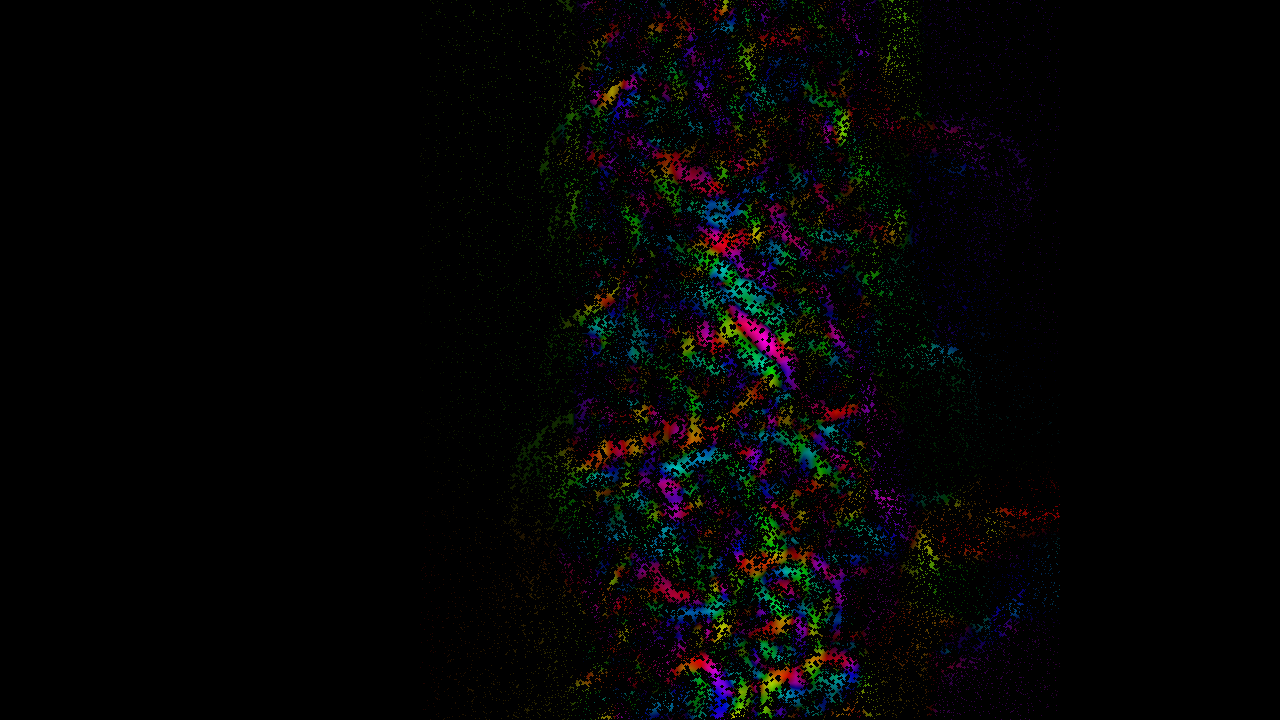}}
        \\

        \rotatebox{90}{\makecell{110~\si{\lux}}}
		&{\includegraphics[clip,trim={11cm 2.5cm 11cm 5cm},width=\linewidth]{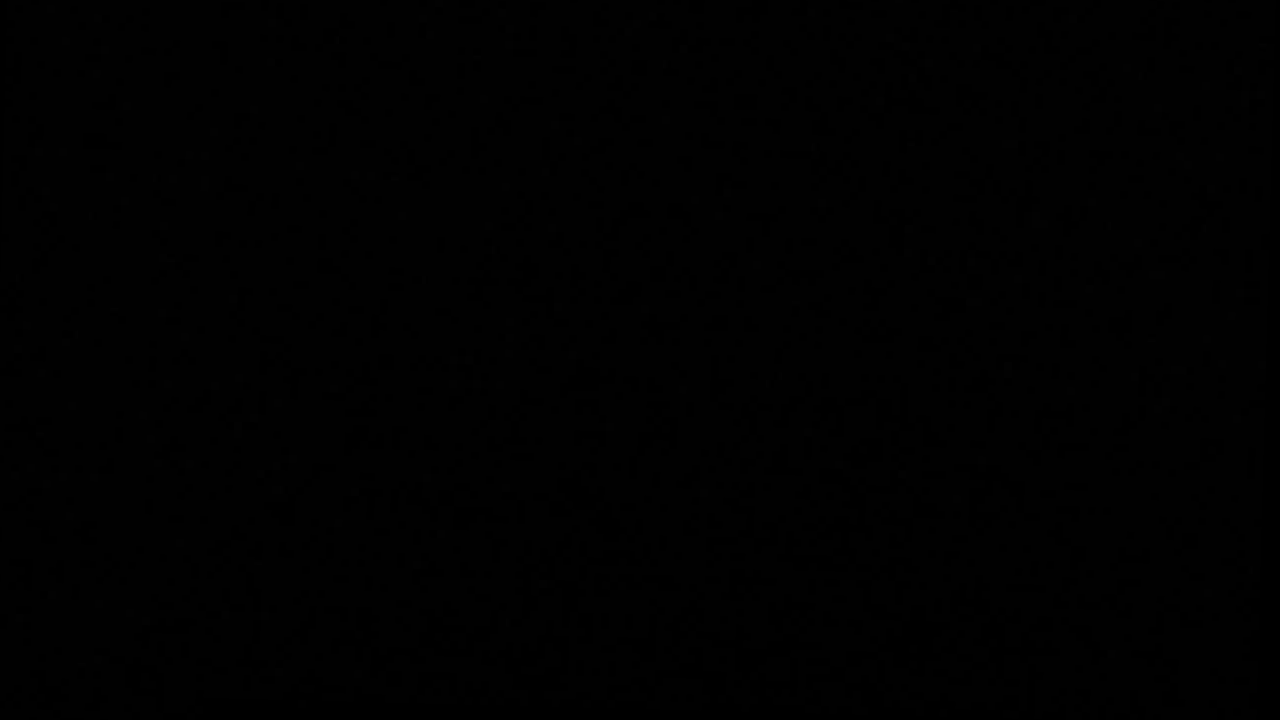}}
		&{\includegraphics[clip,trim={11cm 2.5cm 11cm 5cm},width=\linewidth]{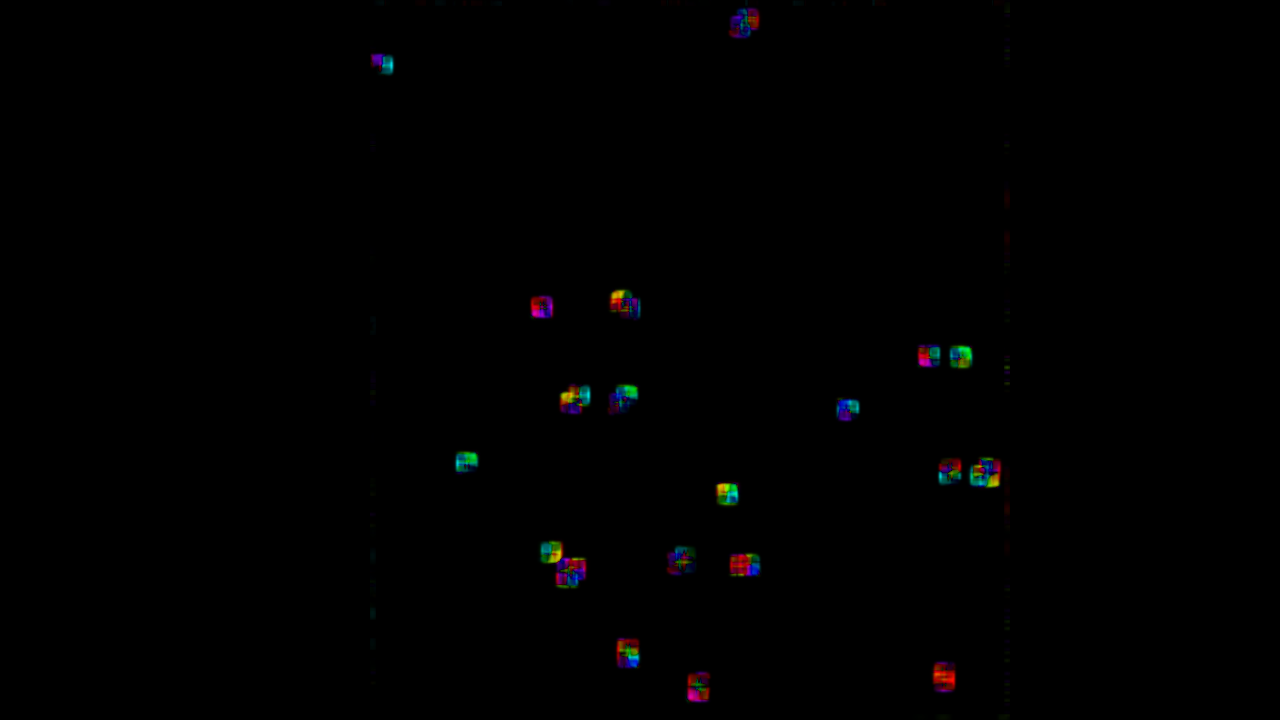}}
		&\gframe{\includegraphics[clip,trim={11cm 2.5cm 11cm 5cm},width=\linewidth]{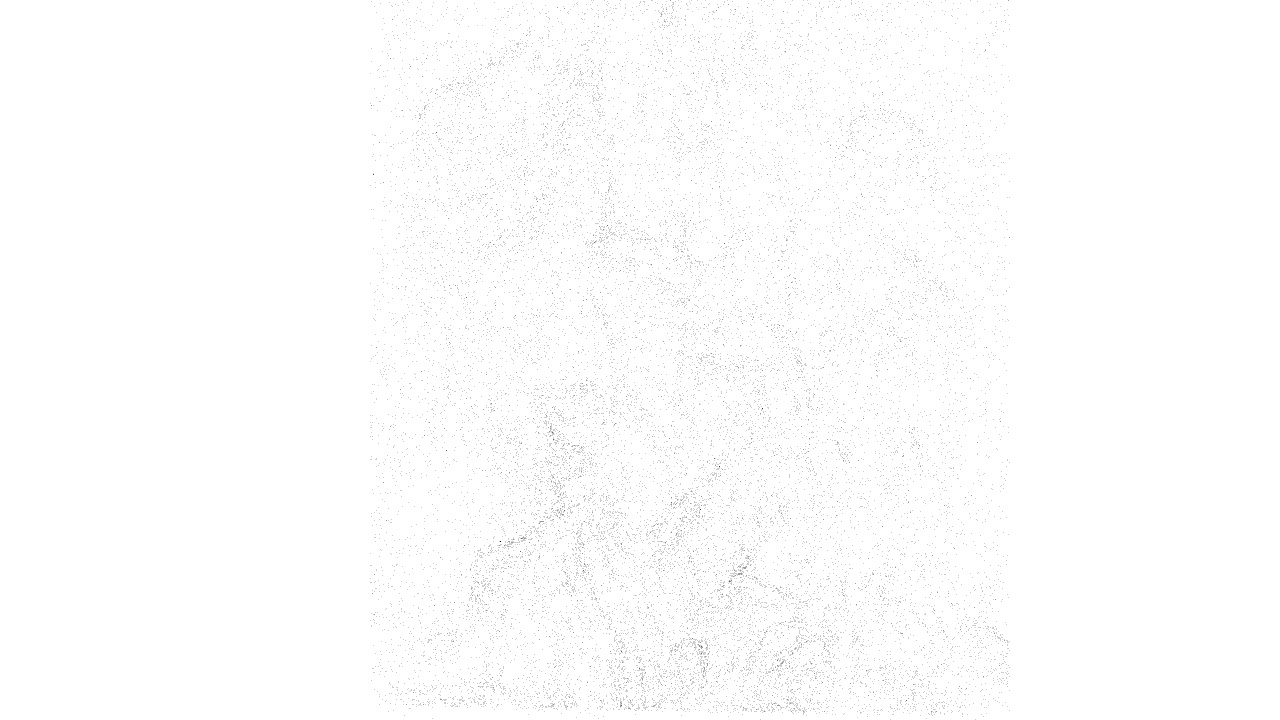}}
		&{\includegraphics[clip,trim={11cm 2.5cm 11cm 5cm},width=\linewidth]{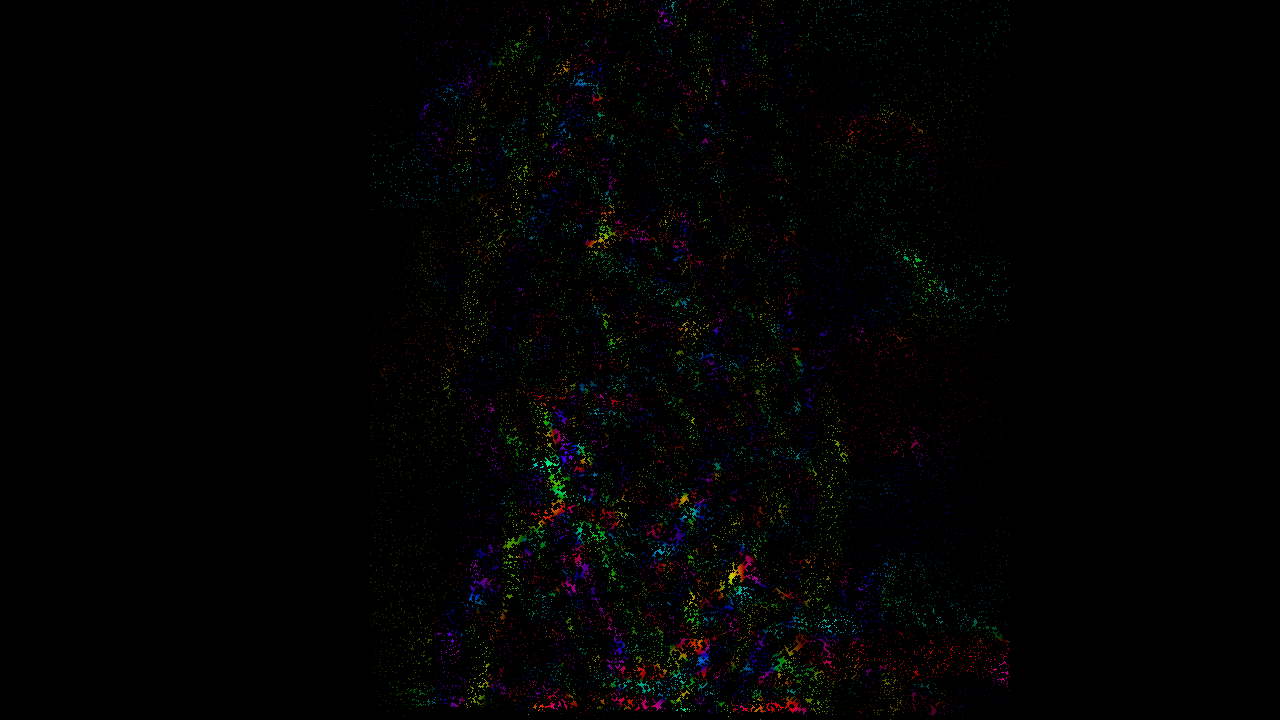}}
        \\

        \rotatebox{90}{\makecell{$<$ 50~\si{\lux}}}
		&{\includegraphics[clip,trim={11cm 2.5cm 11cm 5cm},width=\linewidth]{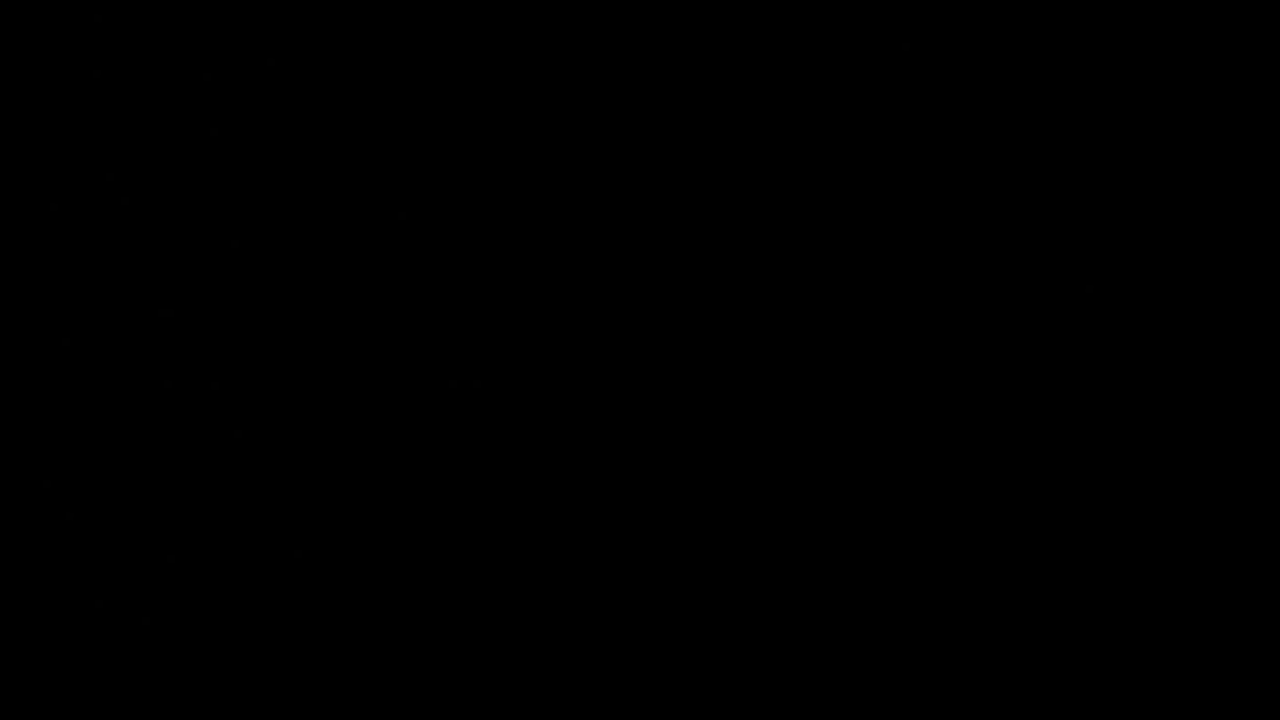}}
		&{\includegraphics[clip,trim={11cm 2.5cm 11cm 5cm},width=\linewidth]{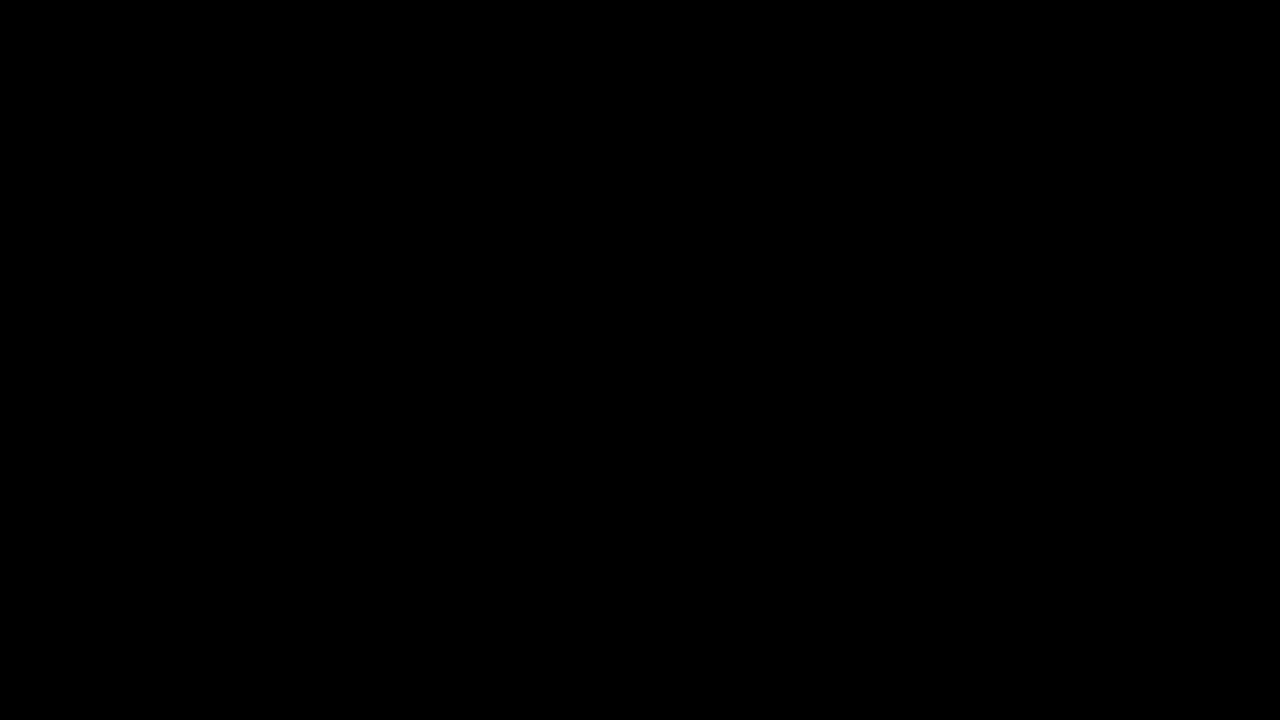}}
		&\gframe{\includegraphics[clip,trim={11cm 2.5cm 11cm 5cm},width=\linewidth]{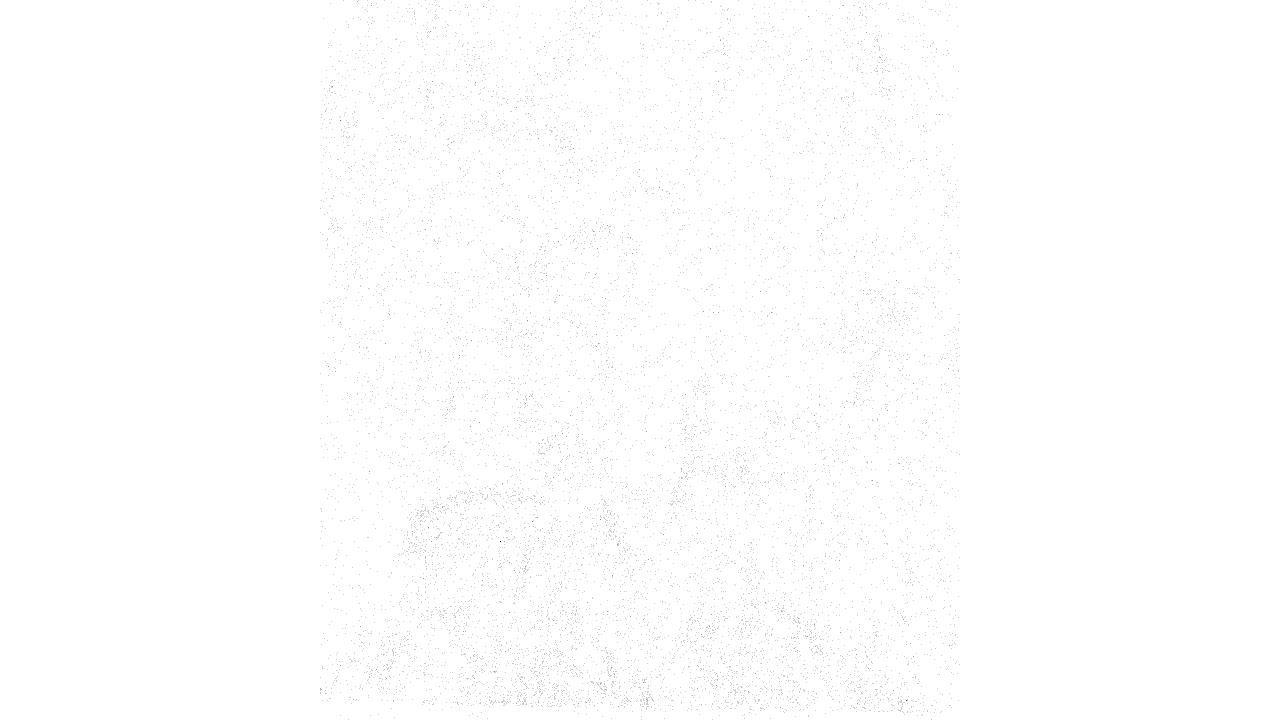}}
		&{\includegraphics[clip,trim={11cm 2.5cm 11cm 5cm},width=\linewidth]{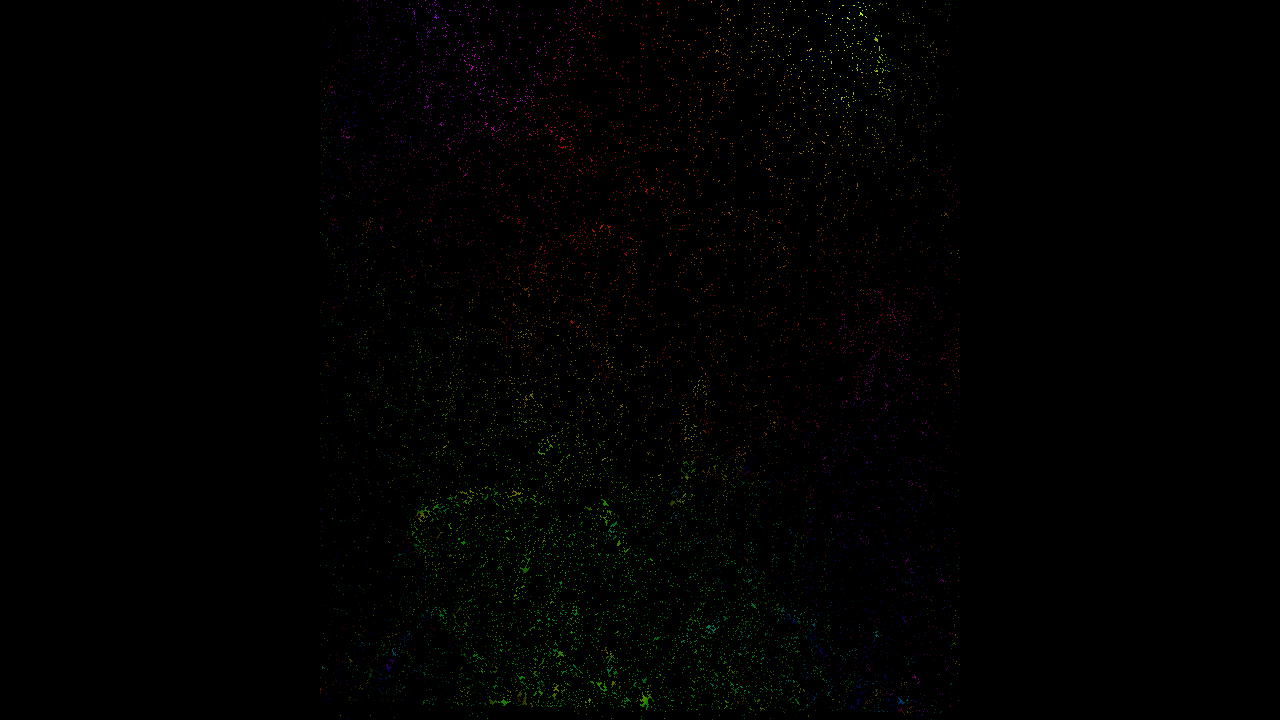}}
        \\

		& \textbf{(a)} Frames
		& \textbf{(b)} Flow from frames
		& \textbf{(c)} Events
		& \textbf{(d)} Ours
	\end{tabular}
	}
	\caption{\emph{Ablation study for different illumination levels.} Flow (b) uses normalized frames of the original ones (a) as input, while our method (d) uses events (c) and the original frames. The frame-based flow deteriorates at around 500~\si{\lux}, while events and the estimated flow capture the schlieren even at illumination levels as low as 225~\si{\lux} and 110~\si{\lux}.}
	\label{fig:suppl:lowLight}
\end{figure*}

%% file: floats/fig_e2vid_framefree.tex
\def\figmethodwidth{.31\linewidth}
\begin{figure}[t]
\centering
\begin{subfigure}{\figmethodwidth}
  \centering
  {\includegraphics[clip,trim={18cm 4cm 15cm 8cm},width=\linewidth]{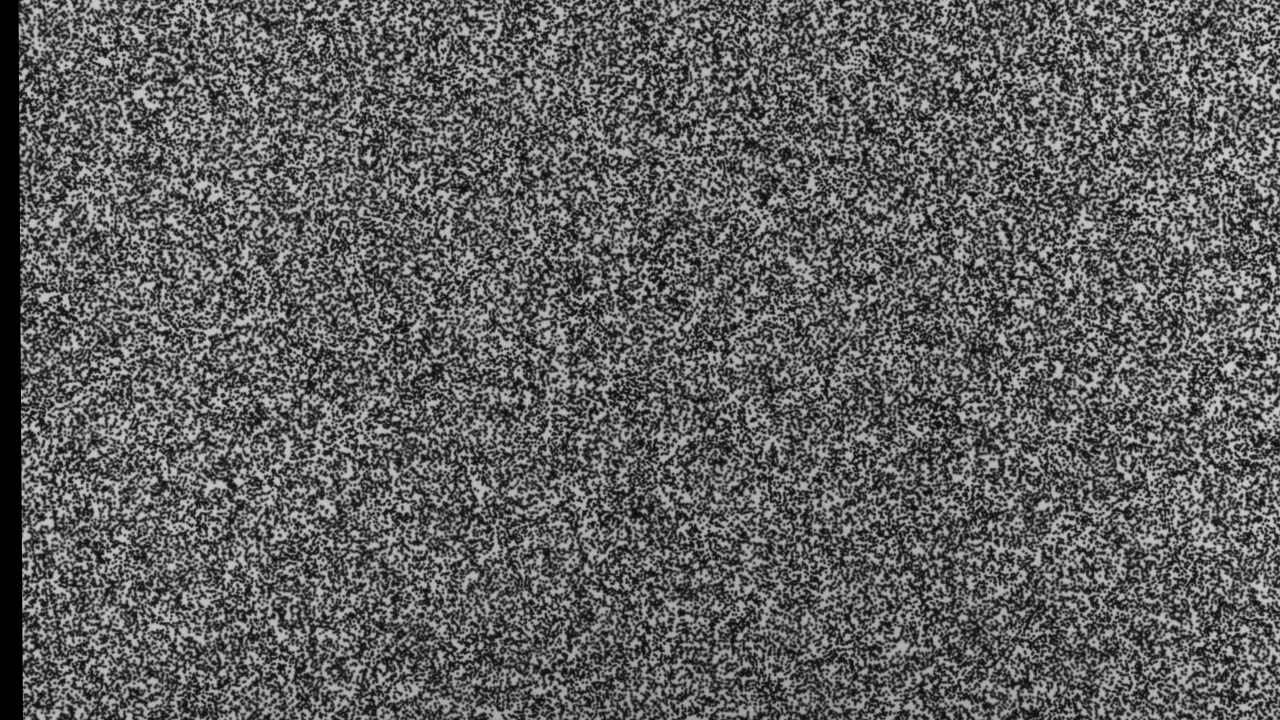}}
\end{subfigure}\;
\begin{subfigure}{\figmethodwidth}
  \centering
  {\includegraphics[clip,trim={18cm 4cm 15cm 8cm},width=\linewidth]{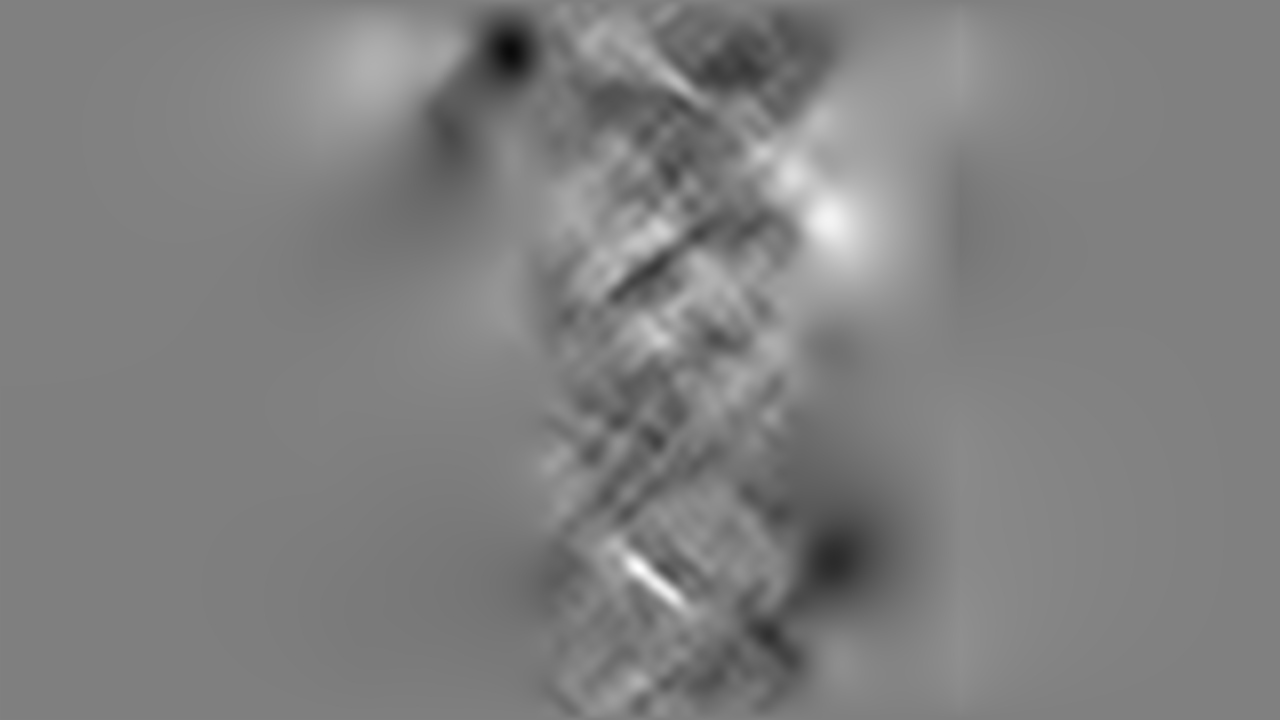}}
\end{subfigure}\;
\begin{subfigure}{\figmethodwidth}
  \centering
  {\includegraphics[clip,trim={18cm 4cm 15cm 8cm},width=\linewidth]{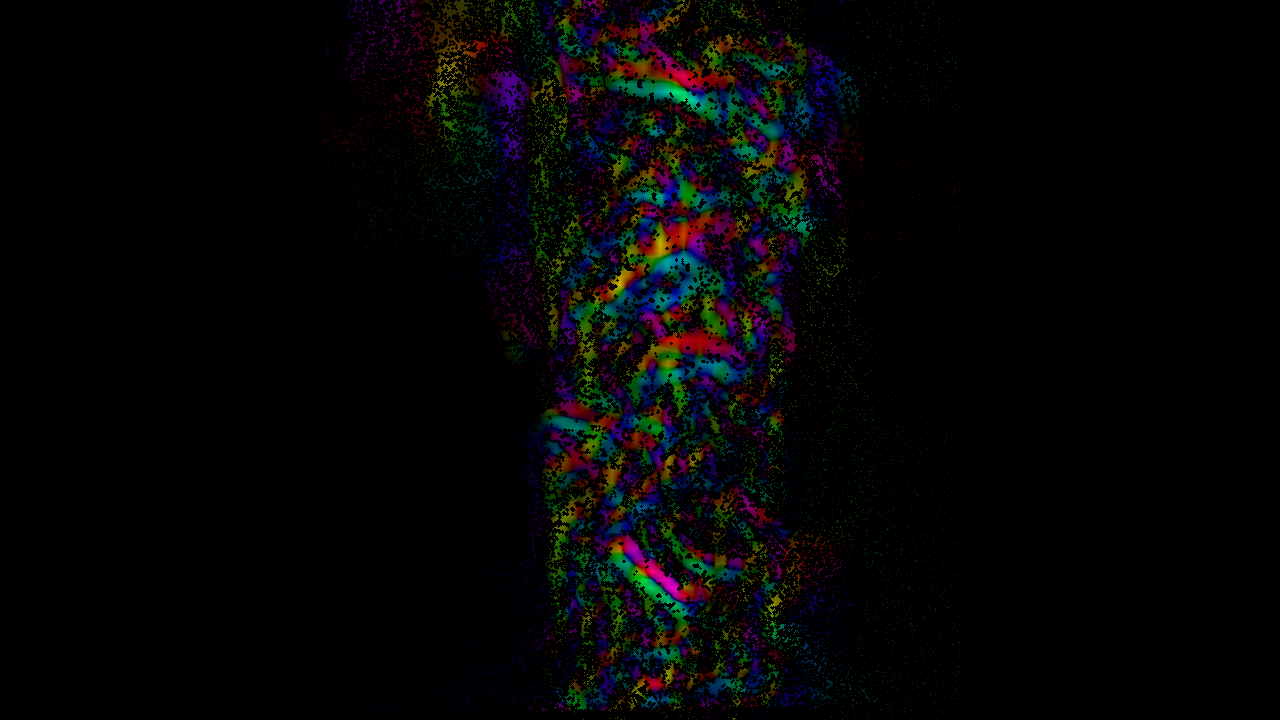}}
\end{subfigure}\\[1ex]
\begin{subfigure}{\figmethodwidth}
  \centering
  {\includegraphics[clip,trim={18cm 4cm 15cm 8cm},width=\linewidth]{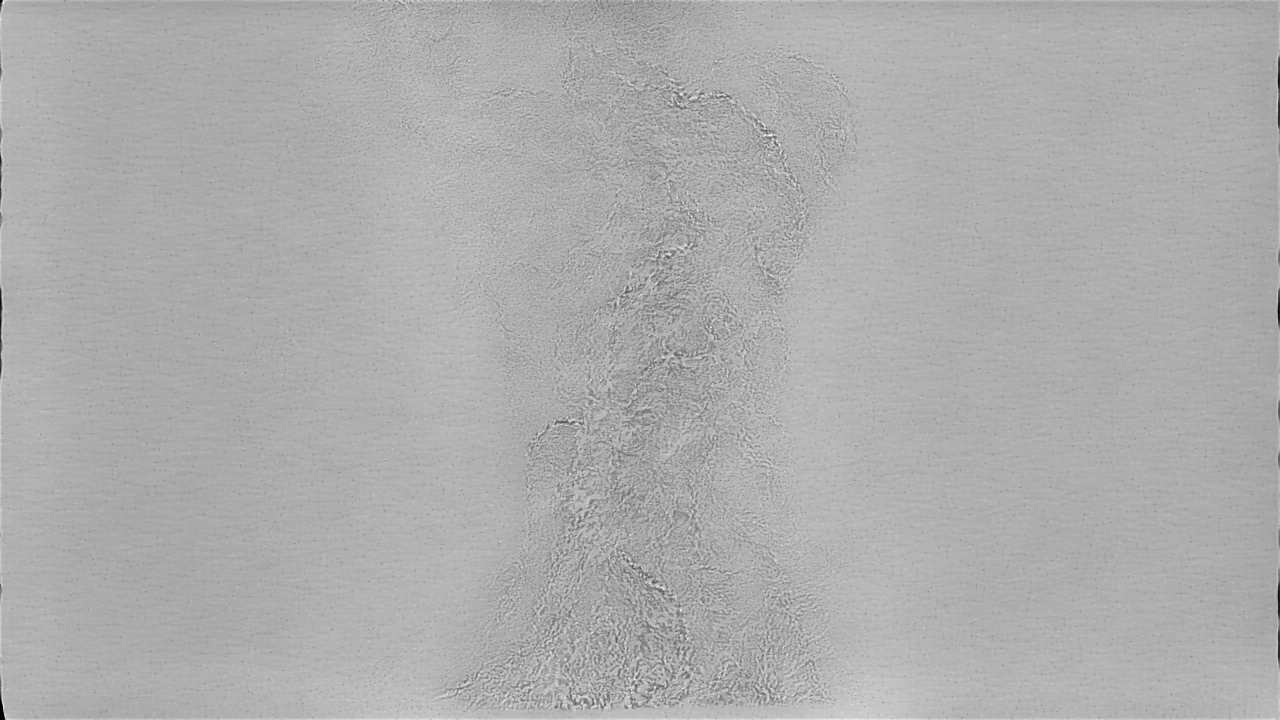}}
  \caption{\centering Input}
\end{subfigure}\;
\begin{subfigure}{\figmethodwidth}
  \centering
  {\includegraphics[clip,trim={18cm 4cm 15cm 8cm},width=\linewidth]{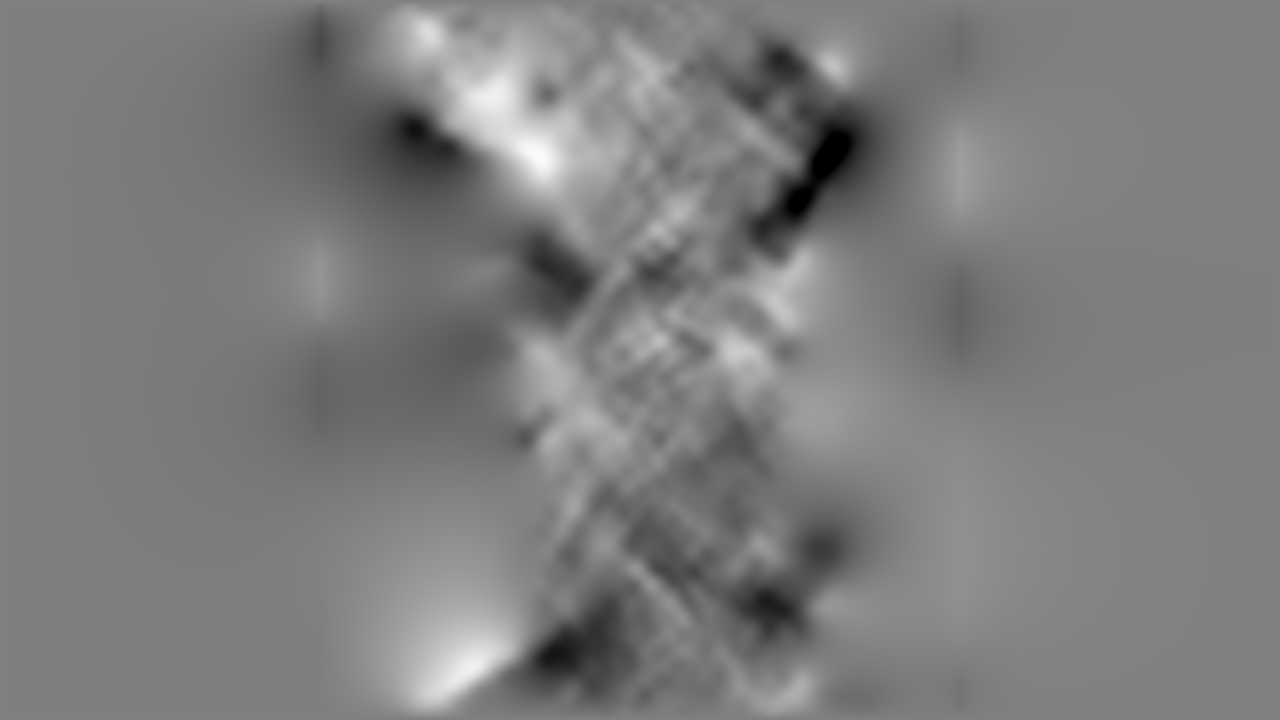}} 
  \caption{\centering Output Poisson}
\end{subfigure}\;
\begin{subfigure}{\figmethodwidth}
  \centering
  {\includegraphics[clip,trim={18cm 4cm 15cm 8cm},width=\linewidth]{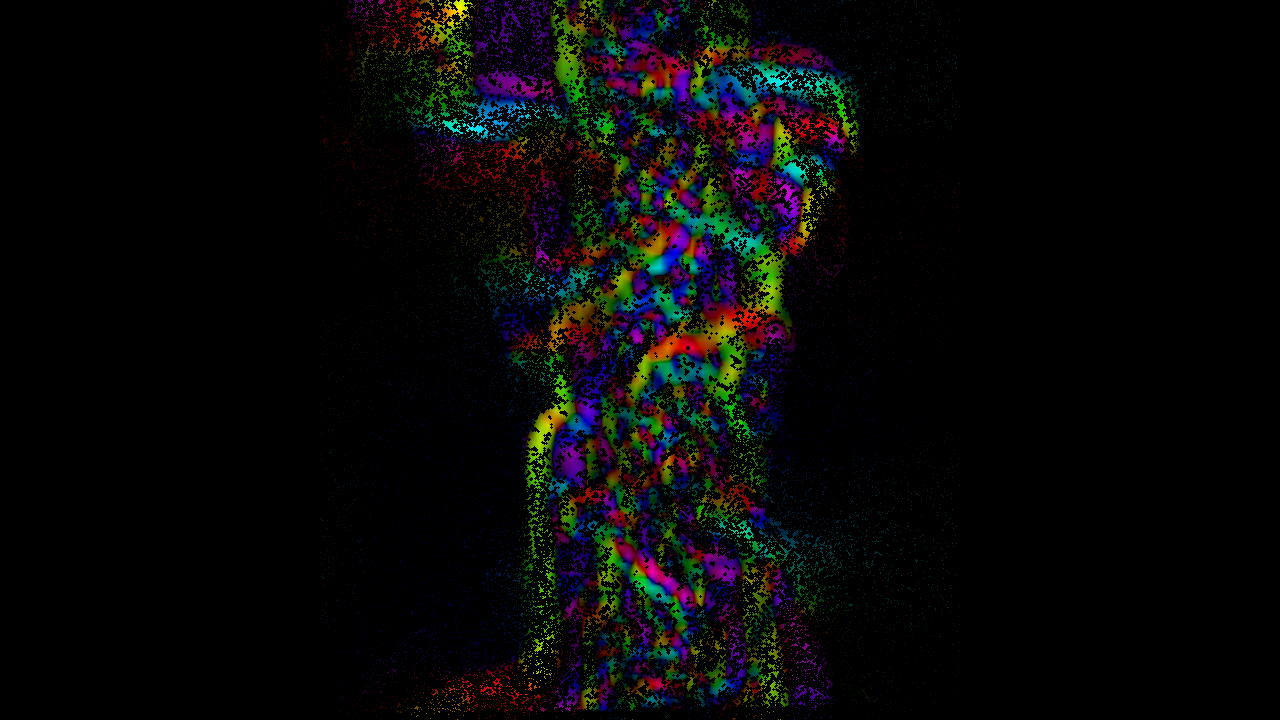}}
  \caption{\centering Output Flow}
\end{subfigure}
\caption{\emph{Towards a frame-free method}. 
The top row shows the originally proposed method with the frame-based camera input. 
The bottom row shows an E2VID-reconstructed image as the alternative input.
In spite of the large quality difference between the two inputs (a), the output Poisson and flow images have some visual similarities (b,c).
}
\label{fig:frameFree}
\end{figure}

%% file: floats/suppltab_ablation.tex
\sisetup{round-mode=places,round-precision=3}
\begin{table*}[!ht]
\centering
\caption{Effect of the regularizers and the translation field.
}
\adjustbox{max width=\textwidth}{%
\setlength{\tabcolsep}{2pt}
\begin{tabular}{l*{10}{S}}

\toprule
 & \multicolumn{3}{c}{Hot plate} %
 & \multicolumn{3}{c}{Crushed Ice} %
 & \multicolumn{3}{c}{Dryer} %
 \\
 \cmidrule(l{1mm}r{1mm}){2-4}
 \cmidrule(l{1mm}r{1mm}){5-7}
 \cmidrule(l{1mm}r{1mm}){8-10}

&\text{AEE $\downarrow$} & \text{\%Out $\downarrow$} & \text{AE $\downarrow$}
&\text{AEE $\downarrow$} & \text{\%Out $\downarrow$} & \text{AE $\downarrow$}
&\text{AEE $\downarrow$} & \text{\%Out $\downarrow$} & \text{AE $\downarrow$}
\\

\midrule
Ours ($\lambda_1 = 0.5$, $\lambda_2 = 0.1$)

& 0.48688 & 12.21458 & 0.42067
& 0.32560 & 5.17659 & 0.30111
& 0.39536 & 10.17393 & 0.33696
\\ 
\midrule
w/o regularizers (i.e., $\lambda_1 =\lambda_2=0$) %
& 3.37093 & 82.03858 & 1.11148
& 2.49937 & 76.32511 & 1.01746
& 1.23317 & 48.62564 & 0.75627
\\ 
w/o translation model (i.e., $\bp=0$) %
& 0.59072 & 18.60890 & 0.48846
& 0.36759 & 7.79103 & 0.31284
& 0.39390 & 10.89641 & 0.32394
\\ 
\midrule
$\lambda_1 = 0.05$, $\lambda_2 = 0.1$
& 0.58576 & 14.46827 & 0.49360
& 0.51751 & 11.29478 & 0.44011
& 0.44934 & 11.10363 & 0.38667
\\ 
$\lambda_1 = 1.0$, $\lambda_2 = 0.1$
& 0.48157 & 10.46225 & 0.41612
& 0.38972 & 3.84948 & 0.34898
& 0.37840 & 7.27373 & 0.32990
\\ 
$\lambda_1 = 0.5$, $\lambda_2 = 0.01$
& 0.50901 & 11.00129 & 0.44009
& 0.43651 & 5.48248 & 0.38577
& 0.39756 & 7.12934 & 0.34375
\\ 
$\lambda_1 = 0.5$, $\lambda_2 = 1.0$
& 0.51738 & 11.59796 & 0.44298
& 0.42886 & 5.60888 & 0.37853
& 0.40948 & 8.11195 & 0.35004
\\

\bottomrule
\end{tabular}
\label{tab:suppl:ablation}
}
\end{table*}

%% file: floats/fig_abl_deblur.tex
\begin{figure}[t]
\centering
\begin{subfigure}{0.48\linewidth}
  \centering
  \gframe{\includegraphics[trim={15cm 6cm 15cm 11cm},clip,width=\linewidth]{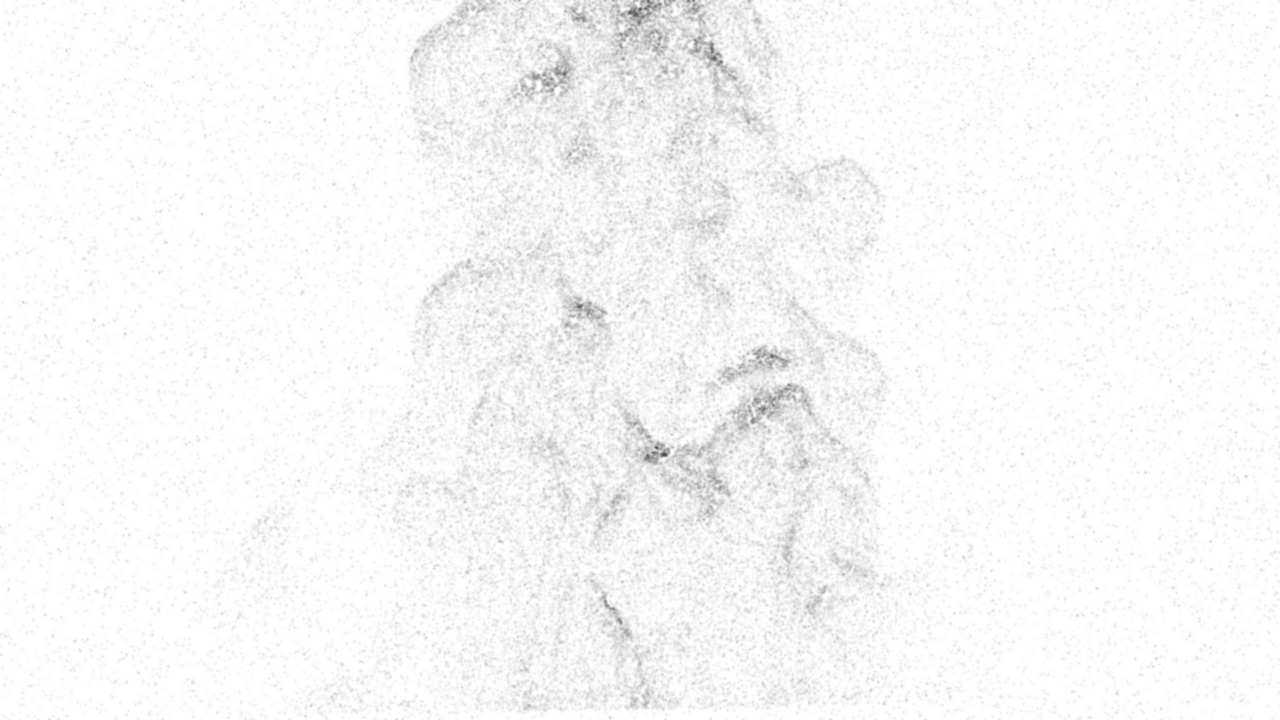}}
  \caption{\centering Original events}
  \label{fig:deblur:orig}
\end{subfigure}\;%
\begin{subfigure}{0.48\linewidth}
  \centering
  \gframe{\includegraphics[trim={15cm 6cm 15cm 11cm},clip,width=\linewidth]{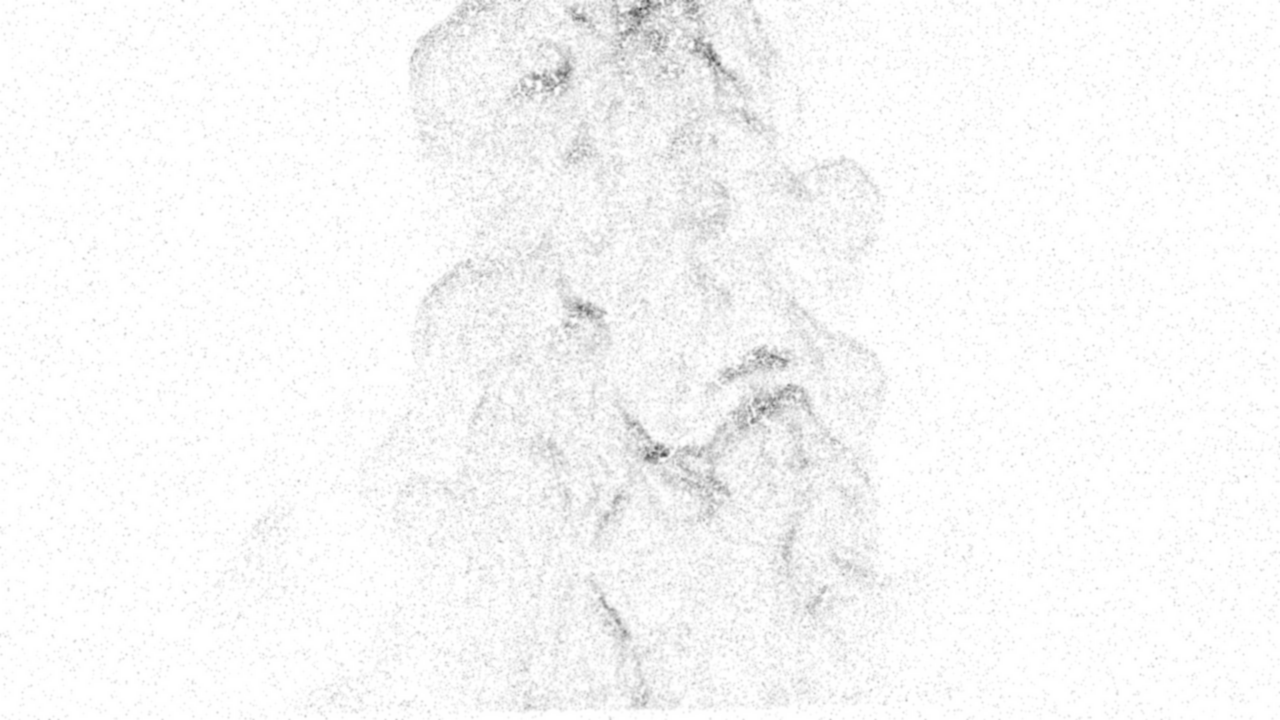}}
  \caption{\centering Warped events}
  \label{fig:deblur:iwe}
\end{subfigure}
\caption{\sblue{\emph{Event warping results using the estimated flow.} Event warping does not improve the sharpness of the data as explained in \cref{sec:experim:deblur}.}
}
\label{fig:deblur}
\end{figure}

%% file: floats/fig_helium_esim_single_col.tex
\def\figmethodwidth{.38\linewidth}
\begin{figure*}
    \centering
\begin{subfigure}{0.174\linewidth}
  \centering
  {\includegraphics[clip,trim={22cm 0cm 0cm 0cm},width=\linewidth]{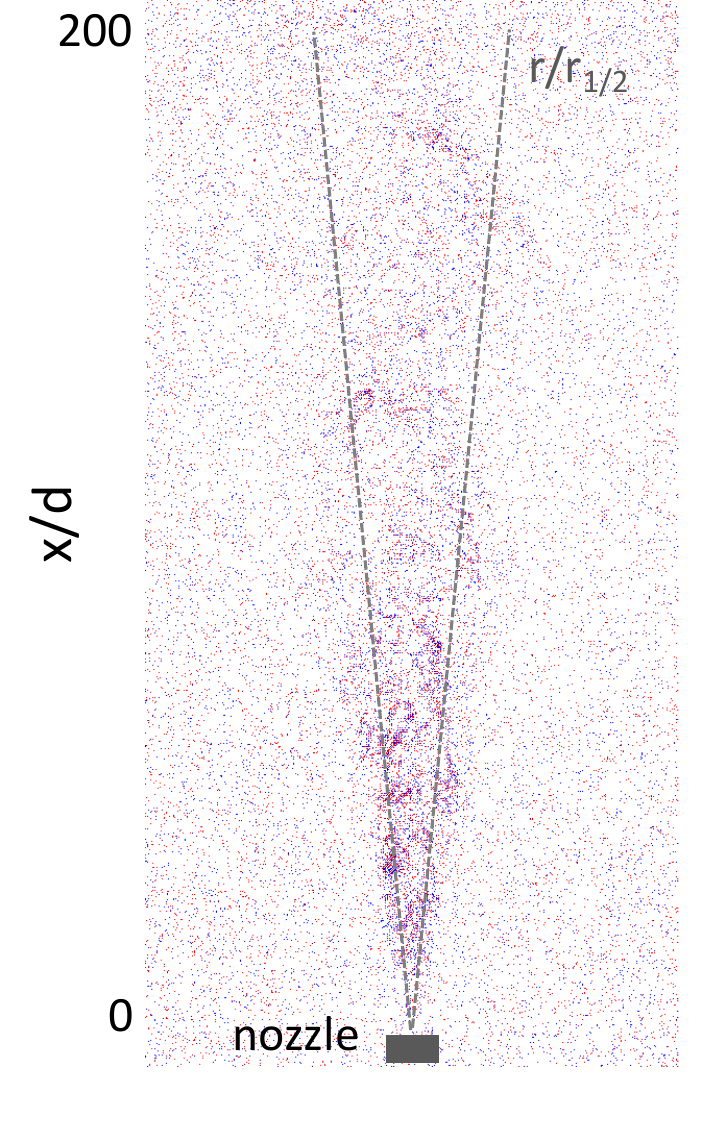}}
   \caption{\centering Events}
   \label{fig:heliumSingleCol:events}
\end{subfigure}\;    
\begin{subfigure}{\figmethodwidth}
  \centering
 {\includegraphics[clip,trim={19cm 8cm 0 0},width=\linewidth]{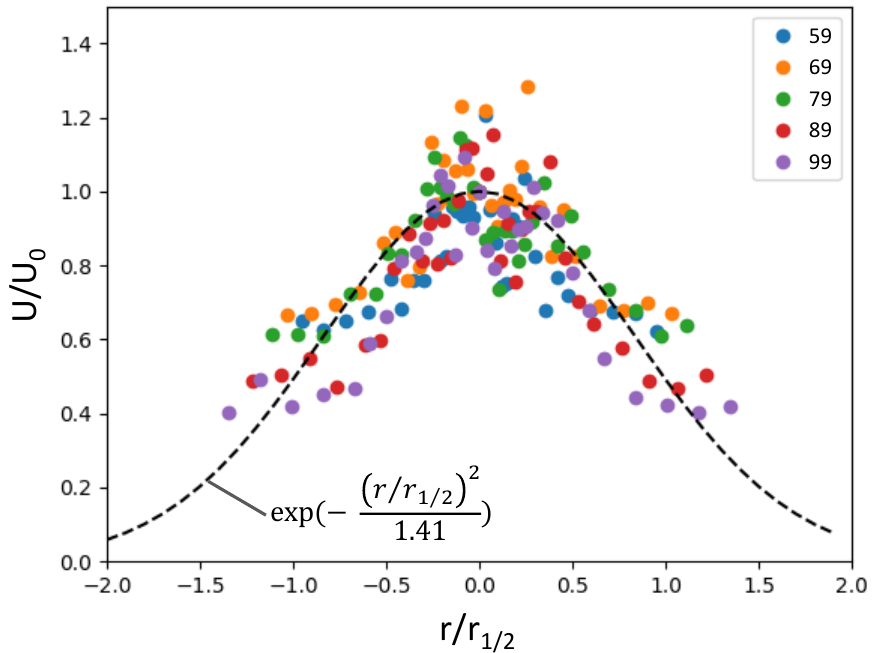}}
  \caption{\centering Similarity analysis}
   \label{fig:heliumSingleCol:similarity}
\end{subfigure}\;
\begin{subfigure}{\figmethodwidth}
  \centering
 {\includegraphics[clip,trim={19cm 8cm 0 0},width=\linewidth]{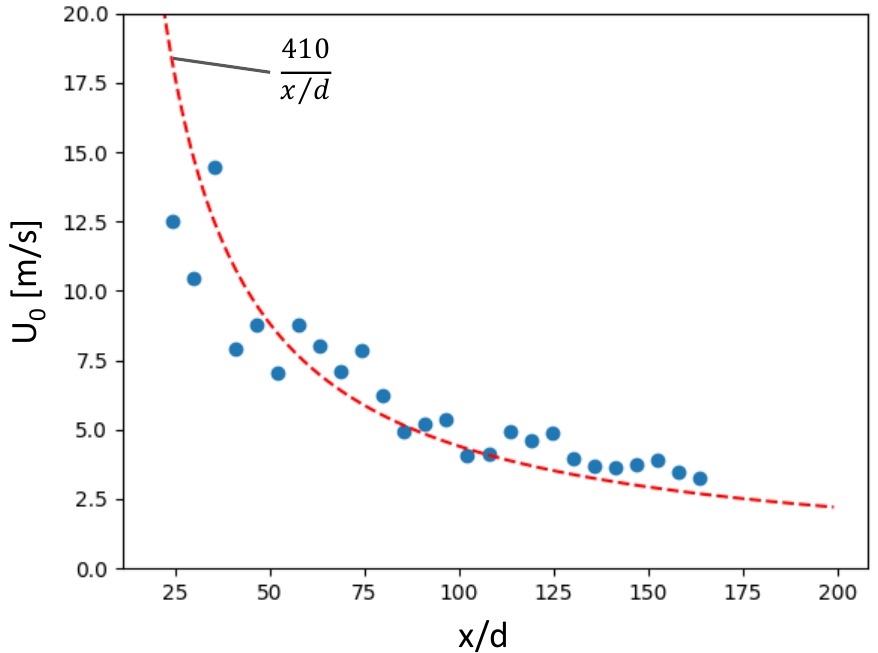}}
  \caption{\centering Velocity analysis}
   \label{fig:heliumSingleCol:velocity}
\end{subfigure}
    \caption{\sblue{\emph{Helium jet experiment} (image data comes from \cite{settles2022schlieren}).
    (a) Event data (simulated from frames) and geometry of the scene.
    (b) Similarity analysis of jet radial profiles for simulated events.  Different colors indicate different distances from the nozzle, $x/d = \{59, 69, 79, 89, 99\}$.
    (c) Velocimetry at the center of the jet nozzle: the velocity estimated from the simulated events follows the $1/x$-type decay.
    Following the same notation as \cite[Fig.~7,8]{settles2022schlieren},
    $x/d$ is the distance normalized by the size of the nozzle (d=1.4mm),
    $r/r_{1/2}$ is the radius normalized by the jet half-width,
    $U_0$ is the estimated velocity profile along the axis of the jet,
    and $U$ is the estimated velocity (parallel to the jet axis) at each pixel.}
    }
    \label{fig:heliumSingleCol}
\end{figure*}

%% file: floats/fig_esim_kymogram.tex
\begin{figure}[t]
 {\includegraphics[clip,trim={10.5cm 3.5cm 0cm 0cm},width=\linewidth]{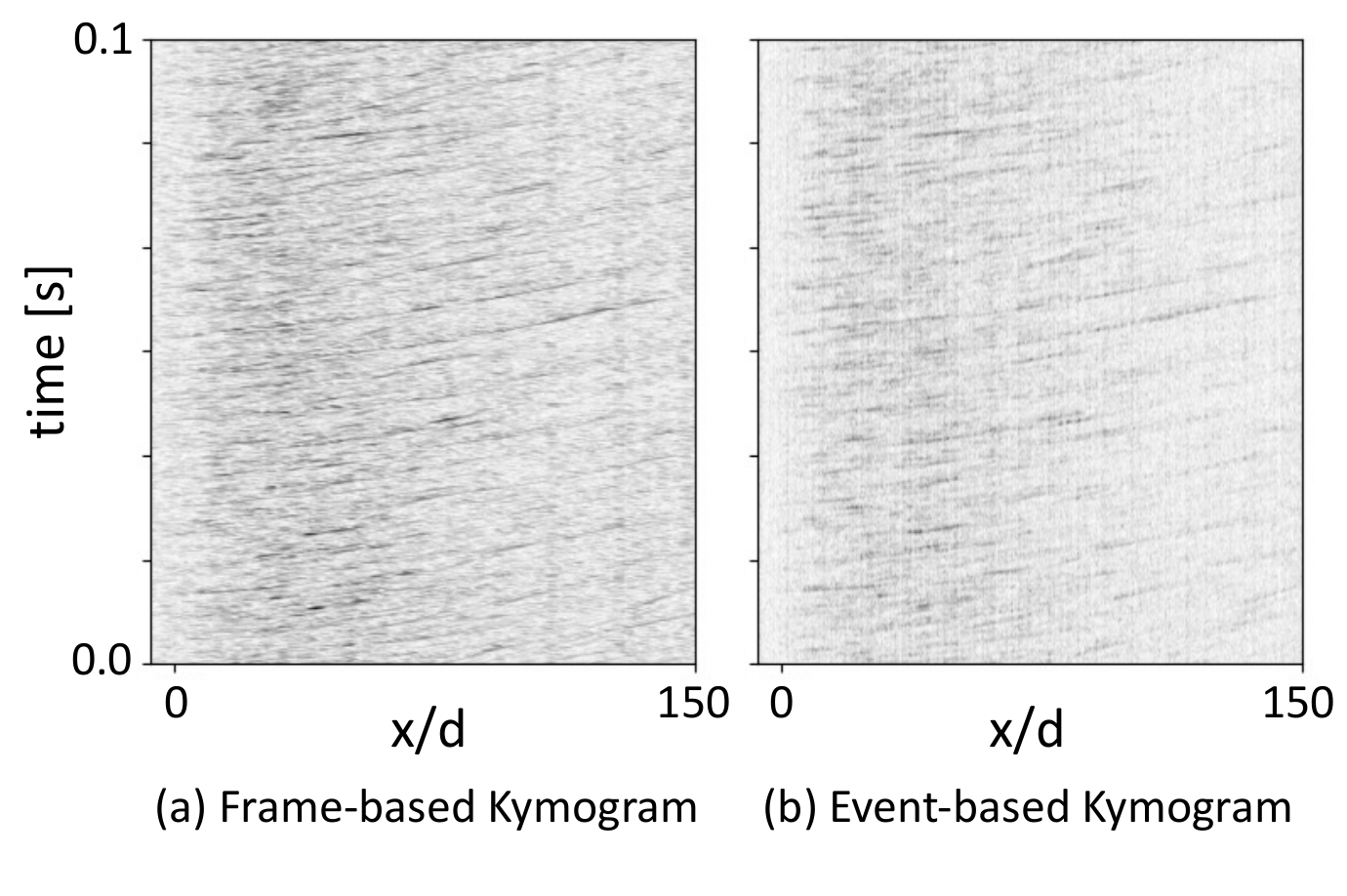}}
\caption{\sblue{\emph{Kymograms} of the self-similar turbulent flow from the helium jet experiment. 
(a) using frame data and (b) using simulated events.
The axes follow the same notation as \cite{settles2022schlieren} for ease of comparison with the figures therein.}}
\label{fig:esimVelo:kymogram}
\end{figure}